\documentclass[12pt]{article}
\usepackage{newtxtext,newtxmath}
\usepackage{appendix}

\usepackage[colorlinks=true, linkcolor=royalblue, citecolor=royalblue, urlcolor=royalblue]{hyperref}

\definecolor{royalblue}{RGB}{65,105,225} 
\usepackage{booktabs,array,tabularx}
\usepackage{xcolor}  
\usepackage{cite}
\usepackage{hyperref}
\usepackage{booktabs}
\definecolor{deepblue}{RGB}{0,51,102}  
\newtheorem{definition}{Definition}
\newtheorem{proposition}{Proposition}
\newtheorem{lemma}{Lemma}
\newtheorem{theorem}{Theorem}
\newtheorem{proof}{Proof}
\newtheorem{corollary}{Corollary}

\renewcommand{\citeform}[1]{\textup{\textcolor{deepblue}{#1}}}
\usepackage{graphicx}
\usepackage{float}
\usepackage[ruled,linesnumbered]{algorithm2e}

\newtheorem{remark}{Remark}
\DontPrintSemicolon
\SetAlgoCaptionSeparator{}
\SetAlFnt{\footnotesize\linespread{0.88}\selectfont}
\SetAlCapFnt{\footnotesize\bfseries}
\SetAlCapNameFnt{\footnotesize}
\usepackage[letterpaper,margin=1in]{geometry}
\renewenvironment{abstract}
	{\quotation}
	{\endquotation}

\date{}

\makeatletter
\renewcommand{\fnum@figure}{\textbf{Figure \thefigure}}
\renewcommand{\fnum@table}{\textbf{Table \thetable}}
\makeatother

\usepackage{scicite}

\usepackage{url}

\def\scititle{%
    {\fontsize{20}{11}\selectfont Finding Kissing Numbers with Game-theoretic Reinforcement Learning}%
}

\title{\bfseries \boldmath \vspace{-4.0em} \scititle}

\author{
    \begin{tabular}{c}
    \fontsize{12}{8}\selectfont Chengdong~Ma$^{1, 2, \ast, \dag}$ \hspace{0.6em} 
    Théo TAO Zhaowei$^{1, 2, \ast}$ \hspace{0.6em} 
    Pengyu~Li$^{1, 2, \ast}$ \hspace{0.6em}
    Minghao~Liu$^{1}$ \\
    \fontsize{12}{8}\selectfont Haojun~Chen$^{1}$ \hspace{0.6em} 
    Zihao~Mao$^{1}$ \hspace{0.6em}
    Yuan~Cheng$^{2, 3}$ \hspace{0.6em} 
    Yuan~Qi$^{2, 3, \dag}$ \hspace{0.6em} 
    Yaodong~Yang$^{1, \dag}$
    \end{tabular} \and
	\small $^{1}$Institute for Artificial Intelligence, Peking University. $^{2}$Shanghai Academy of AI for Science.\and
	\small $^{3}$Artificial Intelligence Innovation and Incubation Institute, Fudan University.\and
	\small $^\ast$Equal contribution. $^\dag$Corresponding authors: chengdong.ma@stu.pku.edu.cn, \\ \small qiyuan@fudan.edu.cn and yaodong.yang@pku.edu.cn
}

\begin{document}

\maketitle

\vspace*{-0.9cm}

\begin{abstract} \fontsize{12.55}{12}\selectfont \bfseries \boldmath

Since Isaac Newton first studied the Kissing Number Problem in 1694, determining the maximal number of non-overlapping spheres around a central sphere has remained a defining challenge in discrete geometry. As the local analogue of Hilbert’s 18th problem, it has profound implications across geometry, number theory and information theory. Although lattices and codes have achieved significant progress, the field is confined to isolated extremal configurations, leaving underlying geometric principles obscured. Here we shift the object to the broader extremal configuration space, thereby opening a new path for the Kissing Number Problem. Accordingly, we recast this problem as a cooperative matrix-completion game, and train a reinforcement learning system, \textit{PackingStar}, to solve it. One player fills cosine entries while the other corrects suboptimal ones, making explosive geometric complexity tractable. Working within extremal configuration spaces, \textit{PackingStar} discovers new interpretable geometric structures that improve 15 strong bounds held for decades in kissing numbers and their generalizations, several of them provably optimal under natural inner products. These findings reveal the first explicit spherical-code realization of the Fischer group $Fi_{22}$, extend the classical Euclidean representation of subgroup structure, and directly inspire subsequent breakthroughs by mathematicians. Overall, the work provides an early example of AI-driven progress on a Hilbert-calibre problem, showing how reinforcement learning advances mathematical discovery by unlocking more expressive objects.


\end{abstract}

\newpage


\section*{Introduction}

The Kissing Number Problem was first posed in 1694 by Isaac Newton and David Gregory \cite{pfender2004kissing}, asking for the maximal number of non-overlapping spheres that can touch a central sphere in three-dimensional space. Its natural extension to \(n\)-dimensional Euclidean space defines the kissing number \(K(n)\), a problem that Paul Erd\H{o}s regarded as the beginning of discrete geometry \cite{brass2005research}. Notably, it is the local analogue of Hilbert's 18th problem on sphere packings \cite{gray2000hilbert,cohn2024improved}, with broad implications for geometry, number theory, group theory and information theory \cite{boyvalenkov2012survey,liberti2017mathematical,liu2023kissing}. For example, the optimal configuration in dimension 24 is realized by the Leech lattice and its Conway-group symmetries \cite{boyvalenkov2012survey}, while kissing configurations also form spherical codes that determine distinguishable signal sets under noise \cite{liu2023kissing}.

It is exceptionally difficult to find kissing numbers in most dimensions, and exact values are known only in dimensions \(1,2,3,4,8\), and \(24\) \cite{schutte1952problem,musin2008kissing,korkine1873formes,leech1967notes,leech1964some,viazovska2017sphere,cohn2017sphere}. Progress on lower bounds has also been slow, with only a few breakthroughs in recent decades through lattices \cite{conway2013sphere}, codes \cite{leech1971sphere}, and group representation theory \cite{ganzhinov2025highly}, improving lower bounds in dimensions such as \(10,13,14,17\), and \(25\) \cite{cohn2024improved,ganzhinov2025highly,kallal2017improved,cohn2011rigidity}. However, these approaches rely heavily on human-designed structures and become increasingly difficult to scale as the combinatorial complexity grows exponentially with dimension. More fundamentally, existing approaches remain confined to individual extremal configurations, leaving the relations among discoveries largely invisible. This obscures the distinction between essential and accidental structure and makes it difficult to identify reusable patterns across dimensions. These limitations point to the need for a new paradigm that moves beyond single configurations, reveals the larger structural landscape in which they reside, and makes this landscape explorable at scale.

\begin{figure}[!htb] 
    \centering
    \includegraphics[width=1\textwidth]{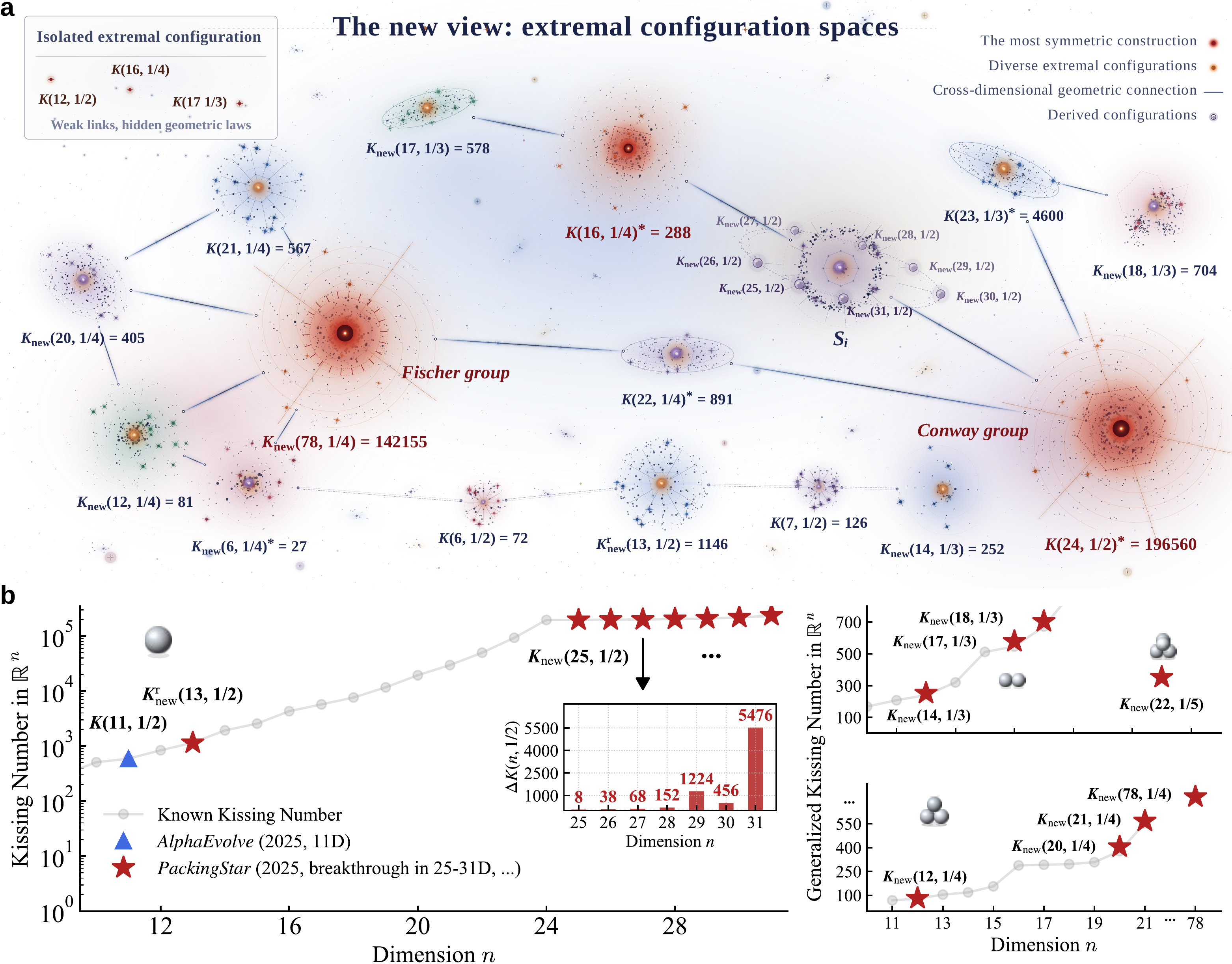} 
    \vspace{-0.8em}
    \caption{\fontsize{11}{11}\selectfont
    \textbf{\textit{PackingStar} reveals extremal configuration spaces and new bounds.}
    \textbf{a}, New view introduced by \textit{PackingStar}. Each galaxy denotes an extremal configuration space, the collection of configurations attaining the same extremal value under a fixed dimension and kissing regime. Stars represent distinct extremal configurations, the bright core marks a highly symmetric representative with generative power, and pathways indicate construction laws linking symmetric cores, derived configurations and related extremal spaces across dimensions.
    \textbf{b}, New records in kissing numbers and generalized kissing numbers. Grey markers show previous best lower bounds, the blue triangle marks the 11-dimensional improvement by \textit{AlphaEvolve} (2025) \cite{novikov2025alphaevolve}, and red stars indicate new values discovered by \textit{PackingStar}. The central inset shows increments \(\Delta K(n,1/2)=K_{\mathrm{new}}(n,1/2)-K_{\mathrm{prev}}(n,1/2)\) for dimensions \(25\)--\(31\), while the right insets summarize new generalized kissing configurations in double-sphere, triple-sphere and quadruple-sphere kissing problems.
    }
    \label{fig1} 
\end{figure}

\begin{figure}[!htb] 
    \centering
    \includegraphics[width=1.00\textwidth]{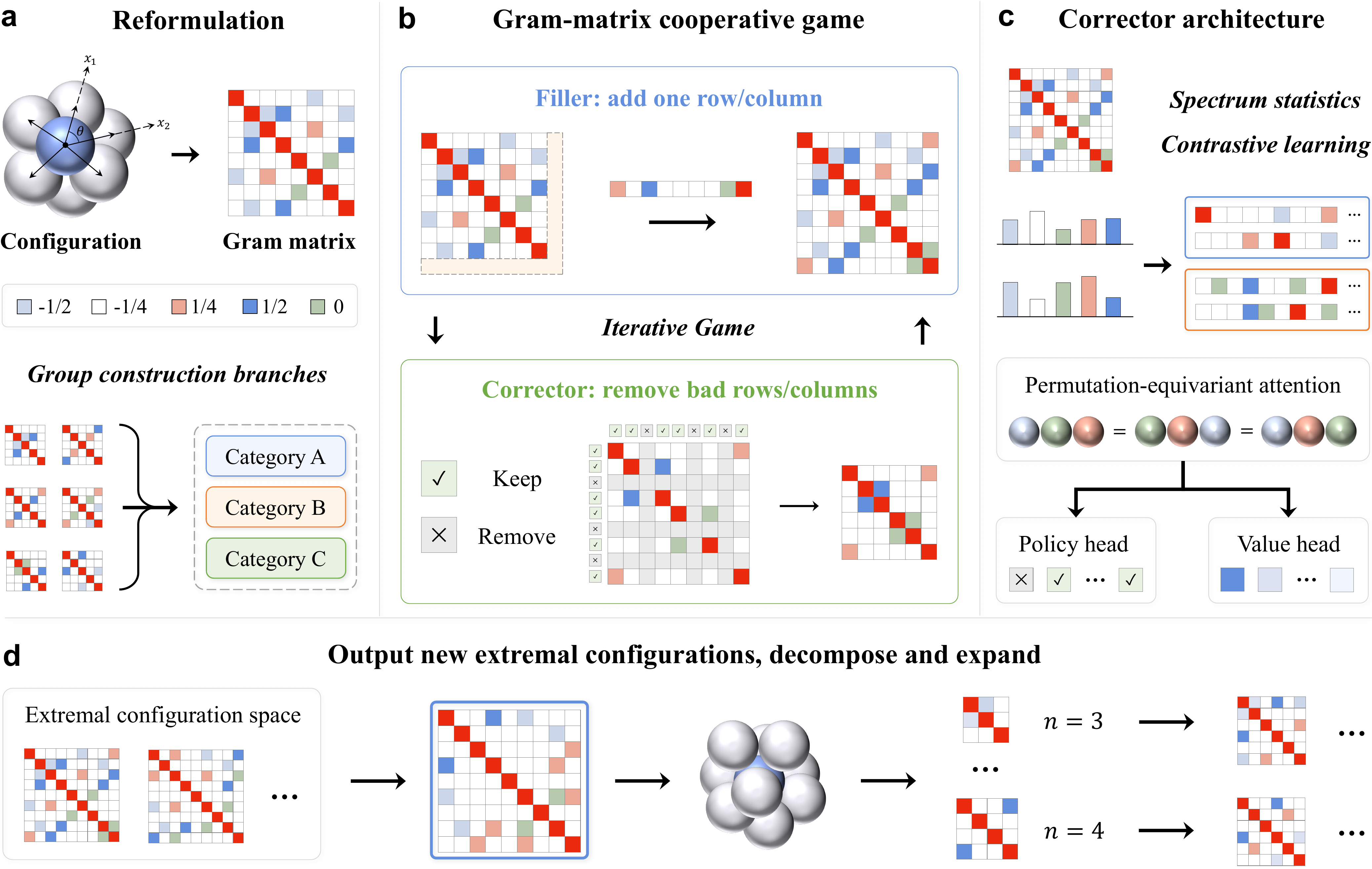} 
    \vspace{-0.8em}
    \caption{\fontsize{11}{11}\selectfont
    \textbf{\textit{PackingStar} for cooperative Gram-matrix exploration.}
    \textbf{a}, A kissing or generalized kissing configuration is represented by its Gram matrix, whose off-diagonal entries encode pairwise cosine values. Algebraic invariants abstract construction branches into structural categories, preserving search diversity while greatly improving exploration efficiency.
    \textbf{b}, Gram-matrix cooperative game. The Filler adds one row and column by sampling candidate cosine entries, while the Corrector removes low-quality rows and columns from the current matrix. Iterative play expands feasible configurations and updates both agents through the shared reward.
    \textbf{c}, Spectrum statistics summarize the row-wise distribution of cosine values and provide contrastive signals that improve representation efficiency and guide reinforcement-learning exploration toward structurally promising matrices. Because configurations are independent of the ordering of spheres, a permutation-equivariant attention module is used to produce order-consistent policy and value estimates for selective correction.
    \textbf{d}, The game outputs sampled extremal configurations, which are decomposed into structural pieces and expanded across dimensions to seed new extremal configuration spaces.
    }
    \label{fig:pipelines} 
\end{figure}

In this paper, we advance the object of study in the Kissing Number Problem from isolated extremal configurations to the richer extremal configuration space, the set of all configurations attaining the same extremal value in a given dimension. The naturalness of this object is underscored by the fact that extremal configurations need not be unique. Previous work on universally optimal spherical codes has shown that the same extremal parameters can admit inequivalent realizations from generalized quadrangles, including the classical elliptic-quadric and Suzuki--Tits constructions governed by different finite simple-group structures such as $U_4(q)$ and $Sz(q)$ \cite{cohn2007universally}. These examples demonstrate that an extremal value can carry a nontrivial geometry of inequivalent realizations, rather than reducing to a single configuration. Thus, the exact extremal value is no longer treated as the endpoint of the problem, but as the signature of a structured family of extremal realizations. This family is the geometric object we call the extremal configuration space.
For intuition, Fig.~\ref{fig1}\textcolor{royalblue}{a} depicts this space through a galaxy metaphor, with individual configurations appearing as stars, the brightest cores marking highly symmetric representatives, and pathways representing construction laws between related spaces. By exploring this space, we uncover which cores have generative power and how their structural pieces reappear across dimensions, relations that an isolated-configuration view cannot reveal.
This shift reframes the Kissing Number Problem from solving isolated instances to understanding a structured world of extremal solutions, where the central questions become why extremal families recur, why certain core configurations generate stronger structures across dimensions, and which substructures truly drive breakthroughs.
We realize this conceptual shift through a fundamentally new computational formulation, \textit{PackingStar}, a multi-agent reinforcement learning system that recasts the Kissing Number Problem as a two-player Gram-matrix completion game (Fig.~\ref{fig:pipelines}\textcolor{royalblue}{b}). Rather than searching directly in coordinate space, \textit{PackingStar} treats the pairwise cosine matrix as the primary object of construction. This formulation is motivated by the observation that large kissing configurations often concentrate their pairwise cosines on a small discrete set, which suggests that extremal geometry can be explored through relational patterns before explicit coordinates are recovered. The resulting Gram-matrix game breaks from existing coordinate-space approaches, avoids high-dimensional numerical instability, and transforms the construction of kissing configurations into a scalable discrete completion problem. The Filler adds entries to grow candidate configurations (Fig.~\ref{fig:pipelines}\textcolor{royalblue}{b}), while the Corrector removes suboptimal entries from a global matrix view, compressing the exponentially large early-stage search space (Fig.~\ref{fig:pipelines}\textcolor{royalblue}{b}). Evolving matrices are not treated merely as terminal outputs. Algebraic invariants abstract construction branches into structural categories (Fig.~\ref{fig:pipelines}\textcolor{royalblue}{a}), while terminal matrices are decomposed into representative substructures that seed subsequent games and induce transitions between extremal configuration spaces (Fig.~\ref{fig:pipelines}\textcolor{royalblue}{d}). These mechanisms turn combinatorial exploration into structural discovery at scale, enabling \textit{PackingStar} to sample extremal configuration spaces, identify recurring blocks, trace relations among extremal solutions and isolate generative laws for stronger constructions.

Working within extremal configuration spaces, \textit{PackingStar} substantially improves 15 long-standing bounds for kissing numbers and their generalizations, including several that had stood for more than 30 years, as summarized in Fig.~\ref{fig1}\textcolor{royalblue}{b}. In 13 dimensions, it surpasses the 1130-sphere rational configurations introduced in 1971 \cite{leech1971sphere}, marking the first breakthrough beyond rational constructions in over half a century. These configurations are not marginal refinements of known constructions, but new geometric constructions outside previously known construction families. Several are moreover provably optimal within their natural inner-product constraints, showing that they are intrinsic extremal objects rather than accidental numerical improvements. \textit{PackingStar} also makes the internal diversity of extremal configuration spaces directly visible. Across dimensions, it discovers more than 8,000 extremal configurations, including over 6,000 distinct constructions at the same extremal value in 14 dimensions alone.  \textit{PackingStar} also reveals strong internal organization in extremal configuration space. Only a small number of exceptional configurations, typically those with maximal symmetry groups, possess genuine generative power across dimensions. In the triple-sphere kissing setting, for example, a highly symmetric 12-dimensional 81-sphere configuration seeds breakthroughs in 20 and 21 dimensions, yielding configurations with 405 and 567 spheres, respectively. These configurations are remarkable not only for their numerical strength, but also for the construction laws they reveal. The same line of discoveries led to the first explicit spherical-code realization of the Fischer group $Fi_{22}$ in 78 dimensions, showing that the 12-, 20- and 21-dimensional breakthroughs are low-dimensional sections of a single higher-dimensional algebraic structure (Fig.~\ref{fig1}\textcolor{royalblue}{a}). This motivates what we call the \emph{Genealogy} of a spherical code: the organized family of distinguished subcodes of an exceptional parent configuration, together with the descent and branching relations between them. Such genealogies provide many benchmarks of spherical-code optimization by establishing new lower bounds. This places the discovery in the lineage of Conway's Leech-lattice work and the ATLAS tradition, viewing spherical codes as Euclidean representations that make subgroup structure visible.

Crucially, these findings have already inspired further advances by mathematicians. Using the construction laws and structural blocks uncovered by \textit{PackingStar}, an improved 18-dimensional spherical code with maximal cosine \(5/14\) and 867 spheres was constructed, surpassing the previous best known construction \cite{cohn2024_spherical_codes}. This demonstrates the transferability and reusability of the discovered laws beyond the original search setting. More broadly, our results suggest that AI can advance mathematics not only by extending known objects, but also by making richer mathematical objects accessible to high-throughput experimental study. \textit{PackingStar} thus opens a new mode of mathematical discovery centered on generating and explaining structure phenomena. All results have been independently validated and are included in standard databases for \href{https://cohn.mit.edu/kissing-numbers/#7}{kissing numbers} and \href{https://spherical-codes.org/}{spherical codes}. In summary, this provides an early example of human-AI collaborative mathematical discovery on a problem of Hilbert-problem calibre: reinforcement learning turns previously inaccessible spaces into objects of systematic exploration, while human abstraction converts the resulting structural patterns into new mathematical concepts and reusable construction laws.

\subsection*{Exploring extremal configuration spaces by Gram-matrix completion}
We now introduce the core framework of \textit{PackingStar}, whose central object is the extremal configuration space. We first formulate kissing and generalized kissing configurations as spherical codes, give their Gram-matrix representation, and then define the extremal configuration spaces that the game samples, classifies and explores. Formally, a kissing or generalized kissing configuration is represented by a finite set of unit vectors
\(X=\{x_1,\ldots,x_m\}\subset\mathbb{S}^{n-1}\), where
$
\mathbb{S}^{n-1}=\{x\in\mathbb{R}^n:\|x\|=1\}.
$
For a maximal inner product \(\alpha\), the generalized kissing number is
\begin{equation}
K(n,\alpha)=
\max\left\{
|X|:
X\subset\mathbb{S}^{n-1},\
\langle x_i,x_j\rangle\le \alpha\ \text{for all }i\ne j
\right\}.
\end{equation}
The classical kissing number is \(K(n)=K(n,1/2)\). Geometrically, \(\alpha=1/3\), \(\alpha=1/4\) and \(\alpha=1/5\) correspond to double-sphere, triple-sphere and quadruple-sphere kissing configurations, respectively.
For a configuration \(X\), its Gram matrix is
\begin{equation}
\mathbf{G}(X)=(G_{ij})_{i,j=1}^m,\qquad 
G_{ij}=\langle x_i,x_j\rangle .
\end{equation}
Thus \(\mathbf{G}(X)\) is symmetric and positive semidefinite, has diagonal entries equal to \(1\), rank at most \(n\), and off-diagonal entries bounded above by \(\alpha\). Conversely, any matrix satisfying these conditions realizes a spherical configuration in \(\mathbb{R}^n\). This makes the construction of kissing configurations equivalent to completing Gram matrices under positivity, rank and inner-product constraints (Fig.~\ref{fig:pipelines}\textcolor{royalblue}{a}).

\begin{definition}[Extremal configuration space]
Let $\mathcal{C}\subset[-1,1)$ be a finite cosine set. Define the class of configurations supported on $\mathcal{C}$ by
\begin{equation}
\Omega(n,\mathcal{C})
=
\{X\subset\mathbb{S}^{n-1}:\langle x_i,x_j\rangle\in\mathcal{C}
\text{ for all distinct }x_i,x_j\in X\}.
\end{equation}
A configuration $X\in\Omega(n,\mathcal{C})$ is called extremal with respect to $\mathcal{C}$ if there is no $y\in\mathbb{S}^{n-1}\setminus X$ such that $X\cup\{y\}\in\Omega(n,\mathcal{C})$. For an attained extremal size $N_*$, the corresponding extremal configuration space is defined as
\begin{equation}
\mathfrak{E}(n,\mathcal{C},N_*)
=
\{X\in\Omega(n,\mathcal{C}): X\text{ is extremal with respect to }\mathcal{C},\ |X|=N_*\}/\!\sim ,
\end{equation}
where \(\sim\) identifies configurations that differ only by an orthogonal transformation of \(\mathbb{R}^n\) and a relabelling of points; equivalently, in the Gram-matrix representation, by simultaneous permutations of rows and columns.
\end{definition}
For each fixed $(n,\mathcal{C},N_*)$, the terminal configurations generated by \textit{PackingStar} collectively form a sampled extremal configuration space,
\begin{equation}
\widehat{\mathfrak{E}}(n,\mathcal{C},N_*)
=
\{[X]: 
X \text{ is extremal with respect to } \mathcal{C},\
|X|=N_*\}
\subseteq
\mathfrak{E}(n,\mathcal{C},N_*).
\end{equation}
where $X$ is generated by \textit{PackingStar}. Under the Gram map $X\mapsto \mathbf{G}(X)$, this sampled space is represented by a collection of Gram matrices up to simultaneous row-column permutations. In this way, \textit{PackingStar} turns the abstract extremal configuration space into a computational object that can be sampled, classified and explored. 

The sampling process is formulated as a cooperative matrix-completion game. 
At each state, the Filler adds one row and column by proposing cosine entries between a new sphere center and the existing centers. 
The Corrector then removes low-quality rows and columns from a global matrix view, allowing the game to recover from suboptimal choices and continue expansion. 
Both agents share a team reward given by the final matrix size, so cooperative learning directly favors larger configurations. Two mechanisms connect this game to extremal configuration spaces rather than merely to individual matrices. 
First, algebraic invariants abstract construction branches into structural categories, allowing the search to compare and prioritize families of related matrices instead of isolated states. Second, terminal matrices are decomposed into representative substructures that seed subsequent games. In this way, a terminal configuration in one sampled extremal configuration space can generate structural pieces that initiate exploration of another. This turns matrix completion into a mechanism for sampling extremal configuration spaces, tracing relations among them and extracting reusable construction laws.

\section*{Results}

\begin{table}[!htb]
    \centering
    \footnotesize
    \setlength{\belowcaptionskip}{8pt}
    \caption{\fontsize{11}{11}\selectfont
    \textbf{New lower bounds $K_{\mathrm{new}}(n,\alpha)$ discovered by \textit{PackingStar}, compared with the previous best known values $K_{\mathrm{prev}}(n,\alpha)$.}
    The final column reports the new construction form for $\alpha=1/2$ and the new cosine set for $\alpha<1/2$. Bold entries mark improved components or bounds discovered by \textit{PackingStar}. In the new construction form, bold items indicate the improved configuration forms discovered by \textit{PackingStar} compared with previous forms \cite{cohn2011rigidity}. in the cosine-set rows, bold cosine sets indicate newly used cosine sets that differ from those in previous constructions. A superscript ``\(\star\)'' denotes a rational construction, ``\(^{\dagger}\)'' indicates a construction obtained by human analysis of data generated by \textit{PackingStar},. and ``--'' indicates that no directly comparable previous realization is listed.}
    \renewcommand{\arraystretch}{1.15}
    \setlength{\tabcolsep}{14pt}
    \begin{tabular}{ccccc}
        \hline
        $n$ &
        $\alpha$ &
        \textit{\textbf{New construction form / cosine set}} &
        $\boldsymbol{K_{\mathrm{new}}(n,\alpha)}$ &
        $K_{\mathrm{prev}}(n,\alpha)$ \\
        \hline

        25 & $1/2$ &
        $K(24,1/2)+|S_1|$ &
        \textbf{197056} &
        197048 \cite{kallal2017improved} \\

        26 & $1/2$ &
        $\boldsymbol{K(2,1/2)}+K(24,1/2)+2|S_1|+2|S_2|$ &
        \textbf{198550} &
        198512 \cite{kallal2017improved} \\

        27 & $1/2$ &
        $\boldsymbol{K(3,1/2)}+K(24,1/2)+2|S_1|+2|S_2|+\sum_{i=3}^{5}|S_i|$ &
        \textbf{200044} &
        199976 \cite{kallal2017improved} \\

        28 & $1/2$ &
        $\boldsymbol{K(4,1/2)}+K(24,1/2)+2\sum_{i=1}^{8}|S_i|$ &
        \textbf{204520} &
        204368 \cite{kallal2017improved} \\

        29 & $1/2$ &
        $\boldsymbol{K(5,1/2)}+K(24,1/2)+\boldsymbol{2\sum_{i=1}^{12}|S_i|+\sum_{i=13}^{14}|S_i|}$ &
        \textbf{209496} &
        208272 \cite{kallal2017improved} \\

        30 & $1/2$ &
        $\boldsymbol{K(6,1/2)}+K(24,1/2)+2\sum_{i=1}^{24}|S_i|$ &
        \textbf{220440} &
        219984 \cite{kallal2017improved} \\

        31 & $1/2$ &
        $\boldsymbol{K(7,1/2)}+K(24,1/2)+\boldsymbol{2\sum_{i=1}^{42}|S_i|}$ &
        \textbf{238350} &
        232874 \cite{kallal2017improved} \\

        13 & $1/2$ &
        $\{-1,\ -1/2,\ -1/4,\ 0,\ 1/4,\ 1/2\}$ &
        \hspace{0.5em}$\mathbf{1146}$\hspace{0.1em}$^{\star}$ &
        \hspace{0.5em}1130\hspace{0.1em}$ ^{\star}$ \cite{leech1971sphere} \\

        14 & $1/3$ &
        $\{-1,\ 0,\ \pm 1/3\}$ &
        \textbf{252} &
        240 \cite{ganzhinov2024spherical} \\

        17 & $1/3$ &
        $\boldsymbol{\{-1,\ -7/9,\ -5/9,\ \pm 1/9,\ \pm 1/3\}}$ &
        \textbf{578} &
        546 \cite{ericson2001codes} \\

        18 & $1/3$ &
        $\boldsymbol{\{-1,\ 0,\ \pm 1/3\}}$ &
        \textbf{704} &
        672 \cite{ericson2001codes} \\

        12 & $1/4$ &
        $\boldsymbol{\{-1/2,\ -1/8,\ 1/4\}}$ &
        \textbf{81} &
        79 \cite{cohn2024_spherical_codes} \\

        20 & $1/4$ &
        $\boldsymbol{\{-1/2,\ -1/8,\ 1/4\}}$ &
        \textbf{405} &
        378 \cite{ericson2001codes} \\

        21 & $1/4$ &
        $\boldsymbol{\{-1/2,\ -1/8,\ 1/4\}}$ &
        \textbf{567} &
        554 \cite{1995Spherical} \\

        22 & $1/5$ &
        $\boldsymbol{\{-1,\ -1/5,\ 1/5\}}$ &
        \textbf{352} &
        338 \cite{ericson2001codes} \\

        78 & $1/4$ &
        $\{-1/2,\ -5/16,\ -1/8,\ 1/16,\ 1/4\}$ &
        \textbf{142155} $^{\dagger}$&
        -- \\

        \hline
    \end{tabular}
    \label{tab:new_bounds}
\end{table}

We use \textit{PackingStar} to explore extremal configuration spaces across multiple dimensions and multi-sphere kissing settings. Rather than producing isolated configurations, the system samples structured families of extremal solutions, extracts reusable blocks and reveals geometric transitions between related spaces. We report four classes of results. First, exploration of the \(S_i\) extremal configuration space improves kissing-number lower bounds in dimensions \(25\)--\(31\). Second, generalized kissing configurations reveal reusable construction laws across double-sphere, triple-sphere and quadruple-sphere kissing problems. Third, several triple-sphere configurations are proved optimal under their natural prescribed inner-product set. Finally, rational extremal configurations in dimension \(13\) break the long-standing 1130-sphere rational barrier. These results show that \textit{PackingStar} can both recover classical lattice-based structures and uncover new geometric laws that expand the known landscape of kissing configurations. The resulting bounds, cosine sets and construction forms are summarized in Table~\ref{tab:new_bounds}. The main configurations are available at
\href{https://github.com/CDM1619/PackingStar}{https://github.com/CDM1619/PackingStar}. More analysis can be found in the \textbf{Supplementary Information}.

\subsection*{New bounds from extremal configuration spaces in dimensions 25--31}

We first evaluate \textit{PackingStar} in dimensions \(25\le n\le 31\), where previous lower bounds were built from the Leech lattice kissing configuration together with carefully chosen subsets \(S_i\) of its minimal vectors \cite{kallal2017improved,cohn2011rigidity}. This setting provides a natural test of whether \textit{PackingStar} can turn a classical construction framework into a space of new structural possibilities.

The first task is to explore the extremal configuration space formed by structurally distinct \(S_i\) with the same extremal value. Each \(S_i\) is a subset of shortest Leech-lattice vectors whose pairwise cosine values are bounded by \(1/4\). In the \textit{PackingStar} framework, this corresponds to the finite cosine sets \(\mathcal{C}_1=\mathcal{C}_2=\{-1,0,\pm 1/4\}\), together with the structural constraint that \(\mathcal{C}_*\) lies inside the Leech-lattice minimal-vector set. Previous constructions reached subsets of size \(|S_i|=488\) \cite{kallal2017improved}. Rather than treating this value as a terminal object, \textit{PackingStar} samples the extremal configuration space at this size and compares inequivalent realizations of \(S_i\).

\begin{figure}[!htb] 
    \centering
    \includegraphics[width=\textwidth]{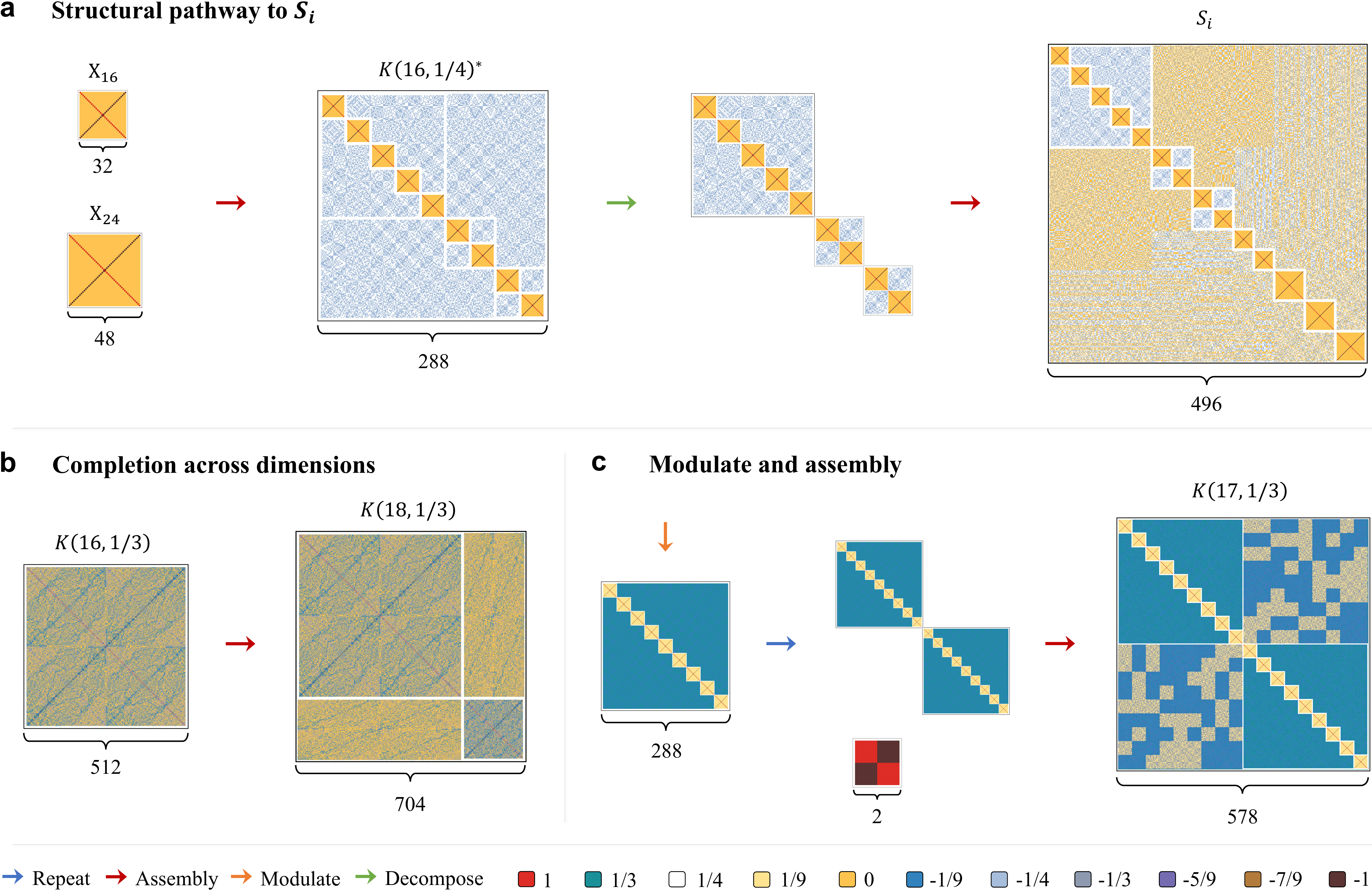} 
    \setlength{\abovecaptionskip}{-10pt}
    \caption{\fontsize{11}{11}\selectfont
    \textbf{\textit{PackingStar} uncovers reusable structural paths between extremal configuration spaces.}
    Each matrix is the Gram-matrix representation of a kissing or generalized kissing configuration, with colors encoding pairwise cosine values.
    \textbf{a}, Structural pathway to $S_i$. \textit{PackingStar} uses the symmetric-frame blocks $X_{16}$ and $X_{24}$ together with an optimal $K(16,1/4)^*$ configuration, then decomposes and reassembles the resulting blocks to construct the 496-sphere Leech-lattice subset $S_i$.
    \textbf{b}, Completion across dimensions extends the 16-dimensional double-sphere kissing configuration $K(16,1/3)=512$ to $K(18,1/3)=704$.
    \textbf{c}, Modulation, repetition and assembly transform a 16-dimensional triple-sphere kissing configuration of size 288 into the 17-dimensional double-sphere kissing configuration $K(17,1/3)=578$.
    Together, these examples show how reusable blocks transfer across dimensions and cosine regimes, revealing structural links between extremal configuration spaces. Blue, red, orange and green arrows denote repeat, assembly, modulate and decompose operations, respectively. A superscript $*$ marks a configuration proved optimal under the corresponding constraint.}
    \label{fig:3} 
\end{figure}

This search reveals a highly symmetric 488-sphere representative whose internal organization is not apparent from the previous construction. In particular, it contains a distinguished substructure belonging to the optimal \(K(16,1/4)\) configuration (Fig.~\ref{fig:3}\textcolor{royalblue}{a}), suggesting that the 488-sphere layer already carries a reusable geometric block. Decomposing this representative block and reinitializing the search from it leads to larger subsets with \(|S_i|=496\). The 496-sphere solutions sampled by \textit{PackingStar} define a new extremal configuration space for \(S_i\). Although its inequivalent realizations differ in detail, they share a stable structural skeleton: 28 symmetric-frame structures \(X_8\) in 8-dimensional subspaces, each containing 16 spheres, together with one 24-dimensional symmetric-frame structure \(X_{24}\), the Conway--Curtis cross \cite{conway2013sphere}, containing 48 spheres. This shared backbone identifies the basic geometric framework of the 496-sphere extremal configuration space. Thus the improvement from 488 to 496 is not an isolated numerical gain; it arises from navigating the extremal configuration space at size 488, identifying a generative core, and using that core to move to a stronger configuration space.

The second task is to discover new assembly forms that use these subsets to build full kissing configurations. Here \textit{PackingStar} does not simply instantiate the previous Leech-lattice template. It searches over alternative ways of combining the Leech-lattice component, the improved \(S_i\), and lower-dimensional kissing configurations. This leads to new construction forms that deviate from the classical pattern \cite{kallal2017improved,cohn2011rigidity}. In dimensions \(26\)--\(31\), the system consistently augments the Leech-lattice component by adding lower-dimensional parts \(K(m,1/2)\), where \(m=n-24\). In dimension \(31\), \textit{PackingStar} uncovers a new assembly pattern by partitioning a 7-dimensional kissing configuration into 42 disjoint unit-radius equilateral triangles, resulting in an \(84\)-fold weighted \(S_i\) and surpassing the previous \(75\)-fold weighted \(S_i\). Similarly, in dimension \(29\), \textit{PackingStar} embeds 12 disjoint unit-radius equilateral triangles into the 5-dimensional kissing configuration to generate a \(26\)-fold weighted \(S_i\), exceeding the prior \(24\)-fold weighted \(S_i\).

These two advances, enlarging the \(S_i\) extremal configuration space and discovering new assembly forms, jointly yield strictly larger kissing configurations in all dimensions from \(25\) to \(31\). The resulting lower bounds \(K_{\mathrm{new}}(n,1/2)\) and their construction forms are summarized in Table~\ref{tab:new_bounds}.

\subsection*{Construction laws for generalized kissing configurations}

We next use generalized kissing problems to test whether the structural laws discovered by \textit{PackingStar} transfer beyond the ordinary kissing number setting. 
The cases \(\alpha=1/3\), \(\alpha=1/4\) and \(\alpha=1/5\) correspond geometrically to double-sphere, triple-sphere and quadruple-sphere kissing problems, respectively, where one asks how many additional unit spheres can be simultaneously tangent to two, three or four mutually tangent spheres. 
These problems provide a controlled family of local packing settings in which extremal configuration spaces can be compared across dimensions and kissing regimes.

As summarized in Table~\ref{tab:new_bounds}, \textit{PackingStar} improves several known values of \(K(n,\alpha)\). 
In the triple-sphere kissing problem, it finds
$K_{\mathrm{new}}(12,1/4)=81$,
$K_{\mathrm{new}}(20,1/4)=405$,
$K_{\mathrm{new}}(21,1/4)=567$. In the double-sphere kissing number problem, it obtains
$K_{\mathrm{new}}(14,1/3)=252$,
$K_{\mathrm{new}}(17,1/3)=578$,
$K_{\mathrm{new}}(18,1/3)=704$,
and in the quadruple-sphere kissing problem it gives a new 22-dimensional configuration with \(352\) spheres. 
These numerical improvements are only the visible layer of the result. 
The more important outcome is that the configurations organize into reusable construction laws, which become visible only by sampling the corresponding extremal configuration spaces.

\begin{figure}[!htb] 
    \centering
    \includegraphics[width=\textwidth]{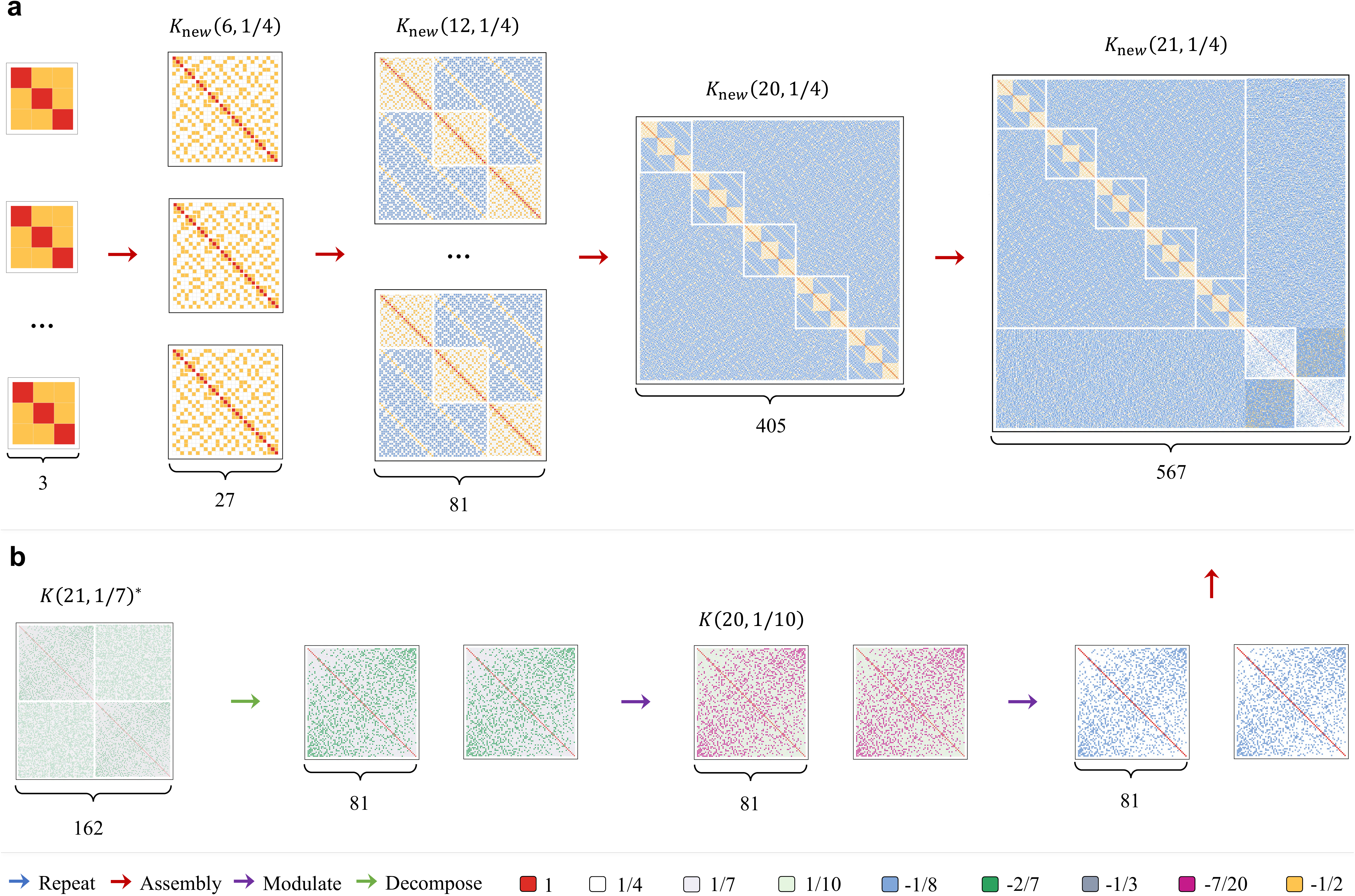} 
    \setlength{\abovecaptionskip}{-10pt}
    \caption{\fontsize{11}{11}\selectfont
    \textbf{\textit{PackingStar} reveals a structural route to stronger triple-sphere kissing configurations.}
    Each matrix is the Gram-matrix representation of a kissing or generalized kissing configuration, with colors encoding pairwise cosine values.
    \textbf{a}, Repeated assembly of small triple-sphere kissing blocks produces several 6-dimensional 27-sphere blocks, which are further assembled into inequivalent 12-dimensional 81-sphere configurations. The highly symmetric 81-sphere structures then serve as generative cores for the 20-dimensional 405-sphere configuration and the 21-dimensional 567-sphere configuration.
    \textbf{b}, An optimal $K(21,1/7)^*$ configuration is decomposed and modulated through a $K(20,1/10)$ construction, yielding reusable 81-sphere blocks that support the  21-dimensional 567-sphere configuration assembly pathway.
    Together, these paths show how \textit{PackingStar} identifies reusable blocks inside extremal configuration spaces and uses them to generate stronger configurations across dimensions and cosine regimes. Blue, red, purple and green arrows denote repeat, assembly, modulate and decompose operations, respectively. A superscript $*$ marks a configuration proved optimal under the corresponding constraint.}
    \label{fig:4} 
\end{figure}

The triple-sphere kissing family gives the clearest example. 
The 12-dimensional 81-sphere configuration is not unique. 
\textit{PackingStar} samples several inequivalent realizations in the 81-sphere extremal configuration space and identifies a distinguished representative with the largest symmetry group. 
This most symmetric realization is the Kronecker product of the Schläfli spherical code with the unit equilateral triangle, and has a transitive symmetry group of order \(311040\). 
It acts as a generative core rather than merely as one extremal solution. 
Its block structure shows that the 12-dimensional 81-sphere configuration arises from three copies of a 6-dimensional 27-sphere block. 
Repeating and completing these blocks produces the 20-dimensional 405-sphere configuration, as shown in Fig.~\ref{fig:4}\textcolor{royalblue}{a}. 
Thus, \textit{PackingStar} reveals a construction law in which larger triple-sphere kissing configurations emerge from repeated block assembly around a highly symmetric representative of an extremal configuration space.

The same principle reappears at the next level. 
The 20-dimensional 405-sphere configuration is also not unique, and different realizations in its extremal configuration space have different generative capacity. 
In our search, the branch with the largest symmetry group is the one that extends further, leading to the 21-dimensional 567-sphere configuration in Fig.~\ref{fig:4}\textcolor{royalblue}{a}. 
This shows that the triple-sphere kissing improvements are not isolated numerical gains. 
They form a construction path through extremal configuration spaces, where highly symmetric representatives serve as cores from which stronger configurations are generated.

This path also reveals a deeper algebraic origin. 
The triple-sphere kissing configurations in 12, 20 and 21 dimensions can be viewed as low-dimensional sections of a single 78-dimensional spherical code with \(142155\) spheres and maximal inner product \(1/4\), realizing the Fischer group \(Fi_{22}\). 
This code is obtained from the primitive action of \(Fi_{22}\) on \(142155\) points. 
The associated permutation representation contains a 78-dimensional irreducible constituent, whose normalized projection gives \(142155\) unit vectors in \(\mathbb R^{78}\) with all off-diagonal inner products at most \(1/4\). 
The same framework also contains the 22-dimensional 891-sphere configuration \(K(22,1/4)=891\), which appears as another section of this 78-dimensional structure. 
From the Leech-lattice side, the 22-dimensional 891-sphere configuration is obtained as a second tangent link of the Leech-lattice kissing configuration. 
Equivalently, it arises from the codimension-two affine section determined by two mutually tangent minimal vectors, followed by projection and normalization. 
The discovered triple-sphere kissing route can therefore be summarized as
\begin{equation}
6\text{D-}27
\longrightarrow
12\text{D-}81
\longrightarrow
20\text{D-}405
\longrightarrow
21\text{D-}567
\longrightarrow
22\text{D-}891
\longrightarrow
78\text{D-}142155 .
\end{equation}
The early steps are governed by repeated block assembly and completion, while the later steps reveal a common higher-dimensional algebraic source. 
Thus, \textit{PackingStar} establishes an explicit geometric bridge between the Fischer-group geometry of \(Fi_{22}\) and the Conway-group symmetry of the Leech lattice. 
This relation cannot be inferred from any single configuration, but emerges from the organization of the full extremal configuration space.

The remaining generalized kissing records show the same space-level mechanism in different forms. 
In the double-sphere kissing setting, the 16-dimensional 512-sphere configuration arises as a section of the 23-dimensional 4600-sphere code \(K(23,1/3)\). 
The 18-dimensional 704-sphere configuration is even more revealing. 
It is not unique, and \textit{PackingStar} samples several inequivalent realizations in its extremal configuration space. 
Among them, only the realization with the largest symmetry group appears as an 18-dimensional section of the same 23-dimensional 4600-sphere code. 
This 23-dimensional code is obtained from an affine hyperplane section of the Leech-lattice kissing configuration, followed by projection and normalization. 
Thus, the \(16\)D-512 and the symmetric \(18\)D-704 configurations are not isolated double-sphere kissing constructions, but lower-dimensional sections of a common Leech-lattice-derived extremal space.

A related section mechanism appears in the quadruple-sphere kissing setting, where the new 22-dimensional 352-sphere configuration is obtained as a 22-dimensional section of a 23-dimensional 552-sphere configuration. 
The 17-dimensional double-sphere kissing configuration with 578 spheres follows a different but complementary pattern. 
As shown in Fig.~\ref{fig:3}\textcolor{royalblue}{c}, it is obtained by reusing and completing structures derived from a 16-dimensional 288-sphere configuration in the triple-sphere kissing setting. 
Structurally, it is composed of two copies of this 288-sphere block together with two additional spheres,
$
578=2\times 288+2.
$
This shows that motifs discovered in one kissing regime can be transferred and completed to produce stronger configurations in another.

These examples show that \textit{PackingStar} does not merely discover separate records for different values of \(K(n,\alpha)\). 
By sampling extremal configuration spaces, it identifies which realizations have the symmetry and structure needed to connect to higher-dimensional codes, which configurations arise as lower-dimensional sections, and which motifs can be reused across kissing regimes. 
These relations are not only explanatory but also constructive, because they guide subsequent searches, provide reusable seeds and reveal pathways to further breakthroughs that would remain hidden from an isolated-configuration view. 
More analysis can be found in the \textbf{Supplementary Information}.

\subsection*{Provable optimality under natural inner products}
\label{sec:results-pounip}

The construction laws above do more than generate large configurations. 
In several prescribed inner-product settings, the configurations generated by \textit{PackingStar} attain upper bounds. 
Thus, within these cosine-set settings, \textit{PackingStar} not only finds locally saturated or numerically large configurations, but also reaches the global optimum of the corresponding prescribed inner-product problem. 
For a prescribed inner-product set \(\mathcal C\), define
\begin{equation}
K_{\mathcal C}(n)^*
=
\max
\left\{
|X|:
X\subset \mathbb S^{n-1},\
\langle x,y\rangle\in \mathcal C
\ \text{for all distinct }x,y\in X
\right\}.
\end{equation}
The \(K_{\mathcal C}(n)^*\) denotes the optimal value of the generalized kissing problem restricted to configurations whose off-diagonal inner products are supported on \(\mathcal C\). It should be distinguished from the unrestricted value \(K(n,\alpha)\), where arbitrary inner products at most \(\alpha\) are allowed. We also distinguish the corresponding optimal part of the extremal configuration space: within the extremal configuration spaces, which can be viewed as galaxies, this optimal configuration space forms a distinguished constellation.
\begin{definition}[Optimal configuration space]
For a finite cosine set \(\mathcal C\subset[-1,1)\), define
\begin{equation}
\mathfrak O(n,\mathcal C)
=
\{X\in\Omega(n,\mathcal C): |X|=K_{\mathcal C}(n)^*\}/\!\sim .
\end{equation}
We call \(\mathfrak O(n,\mathcal C)\) the optimal configuration space
for the prescribed support \(\mathcal C\).
\end{definition}
Every element of \(\mathfrak O(n,\mathcal C)\) is extremal with respect to
\(\mathcal C\). Hence
\begin{equation}
\mathfrak O(n,\mathcal C)
\subseteq
\mathfrak E(n,\mathcal C,K_{\mathcal C}(n)^*).
\end{equation}
Thus the extremal configuration space is the natural sampling space for
\textit{PackingStar}, while the optimal configuration space is the portion
certified to have maximum possible cardinality.

For the triple-sphere family, \textit{PackingStar} repeatedly discovers configurations whose pairwise inner products lie in the same natural set
$
\mathcal C
=
\left\{-1/2,\ -1/8,\ 1/4\right\}.
$
This set appears along the construction path
$
12\text{D-}81
\longrightarrow
20\text{D-}405
\longrightarrow
21\text{D-}567,
$
suggesting that it captures an intrinsic algebraic structure of these extremal configuration spaces rather than a purely artificial restriction.

\begin{theorem}[Optimality under the prescribed triple-sphere inner products]
\label{thm:triple-natural-inner-products}
The prescribed inner-product triple-sphere kissing values are
\begin{equation}
K_{\mathcal C}(12)^*=81,\qquad
K_{\mathcal C}(20)^*=405,\qquad
K_{\mathcal C}(21)^*=567.
\end{equation}
\end{theorem}
where $\mathcal C = \left\{-1/2,\ -1/8,\ 1/4\right\}$. The lower bounds are realized by the \(81\)-sphere, \(405\)-sphere and \(567\)-sphere configurations discovered by \textit{PackingStar}. 
The matching upper bounds show that, within the prescribed set \(\{-1/2,-1/8,1/4\}\), these configurations are global optima rather than local search outcomes.

A similar certification appears in the quadruple-sphere kissing setting in dimension \(22\). 
Two natural prescribed inner-product sets arise:
$
\mathcal C_1
=
\left\{-1,\ -1/5,\ 1/5\right\},
$
$
\mathcal C_2
=
\left\{-3/5,\ -1/5,\ 1/5\right\}.
$
The second set cannot support a configuration as large as the one found by \textit{PackingStar}. To certify this, we use exact semidefinite-programming certificates based on the Bachoc--Vallentin three-point bound, following the framework of \cite{cohn2024optimality}. Our upper-bound computation gives
\begin{equation}
K_{\mathcal C_2}(22)^*\le 336.
\end{equation}
By contrast, \textit{PackingStar} finds a 352-sphere configuration supported on \(\mathcal C_1\). 
We therefore focus on \(\mathcal C_1\), for which the matching upper bound gives the prescribed inner-product optimum.

\begin{theorem}[Optimality under the prescribed quadruple-sphere inner products]
\label{thm:quadruple-natural-inner-products}
The prescribed-inner-product quadruple-sphere kissing value in dimension \(22\) is
\begin{equation}
K_{\mathcal C}(22)^*=352.
\end{equation}
\end{theorem}
where $\mathcal C = \left\{-1,\ -1/5,\ 1/5\right\}$.
To prove the matching upper bounds in Theorems~\ref{thm:triple-natural-inner-products} and~\ref{thm:quadruple-natural-inner-products}, we use the Delsarte linear-programming bound for spherical codes, following the harmonic-analysis framework of Delsarte, Goethals and Seidel~\cite{delsarte1977}. For the prescribed sets considered here, the resulting LP upper bounds match the sizes of the configurations discovered by \textit{PackingStar}. Hence no configuration supported on the stated inner-product sets can exceed \(81\), \(405\), \(567\) or \(352\) spheres in the corresponding dimensions.

This certification gives the mathematical guarantees of the extremal configuration space. 
It proves that, in these natural cosine-set settings, the configurations discovered by \textit{PackingStar} are set-constrained global optima. 
It does not prove unrestricted optimality for the full triple-sphere or quadruple-sphere kissing problems, where arbitrary inner products at most \(1/4\) or \(1/5\) are allowed. 
Nevertheless, the recurrence of these inner-product sets, the high symmetry of the certified configurations and the matching upper bounds suggest that the same structures may also be optimal in the unrestricted problems, although this remains open.

\begin{figure}[t] 
    \centering
    \includegraphics[width=0.95\textwidth]{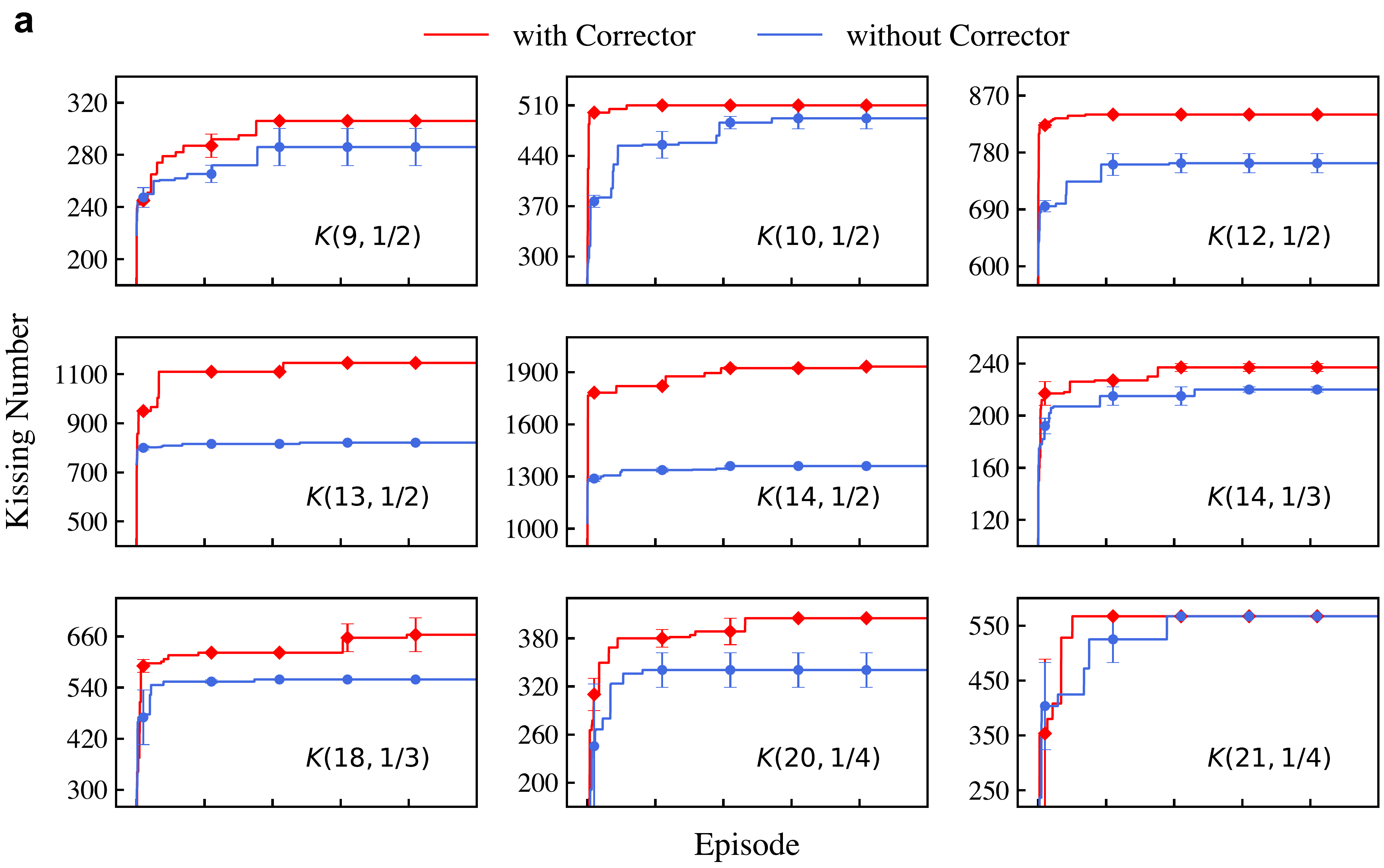} 
    \caption{\fontsize{11}{11}\selectfont
    \textbf{Ablation study of the Corrector in \textit{PackingStar}.}
    Learning curves compare \textit{PackingStar} with and without the Corrector across nine kissing-number and generalized kissing-number tasks.
    Red curves denote the full system with the Corrector, and blue curves denote the ablated system without the Corrector.
    Each panel shows the best kissing number found up to each training episode for one target setting.
    Error bars denote variation across independent runs.}
    \label{fig:5} 
\end{figure}

\subsection*{Improved rational configurations in 13 dimensions}

The 13-dimensional rational case provides a different test of extremal configuration spaces. 
Here the search is restricted not by a new multi-sphere kissing regime, but by rational Gram matrices and prescribed rational cosine sets. 
We apply \textit{PackingStar} to rational kissing configurations in dimension \(13\), where two cosine sets repeatedly arise:
$
\mathcal{C}_1=\{-1,\ 0,\ \pm 1/4,\ \pm 1/2\}
$,
$
\mathcal{C}_2=\{-1,\ -3/4,\ 0,\ \pm 1/4,\ \pm 1/2\}
$.
Within these rational extremal configuration spaces, \textit{PackingStar} discovers several inequivalent rational configurations with
$
K^{\mathrm r}_{\mathrm{new}}(13,1/2)=1146
$,
surpassing the previously known 1130-sphere rational construction introduced in 1971 \cite{leech1971sphere}. 
The new configurations have fully rational Gram matrices, allowing exact verification of all pairwise inner products without relying on numerical approximation.

The sampled 1146-sphere rational extremal configuration space reveals a striking internal diversity. 
One branch decomposes into a 1008-sphere component and a 138-sphere component, where the 1008-sphere component is completely non-antipodal. 
In this branch, no sphere in the 1008-sphere component has an antipodal counterpart, yet the configuration remains highly regular, with each sphere tightly contacting 98 neighboring spheres. 
Its cosine set is
$
\{-1,\ -3/4,\ 0,\ \pm 1/4,\ \pm 1/2\}.
$
This combines rationality, high local regularity and complete non-antipodality.

A second branch attains the same value \(1146\) through a different structural decomposition. 
It is formed by combining a fully antipodal 1008-sphere component with a non-antipodal 138-sphere component, while the full 1146-sphere configuration again has cosine set
$
\{-1,\ -3/4,\ 0,\ \pm 1/4,\ \pm 1/2\}.
$
Therefore, two configurations with the same size, the same dimension and the same cosine support can have fundamentally different internal organization. 
One branch is built around a completely non-antipodal 1008-sphere core, whereas another combines an antipodal 1008-sphere core with a non-antipodal residual component.

This shows that the 1146-sphere rational extremal configuration space contains multiple structural branches with the same extremal value but different decomposition principles. 
Such diversity would be invisible from a single-configuration view. 
By sampling the extremal configuration space, \textit{PackingStar} exposes hidden rational structures beyond the classical construction and shows that even rational kissing configurations can support rich, non-unique geometric organization.

\subsection*{Ablation of the Corrector}

We ablate the Corrector to evaluate its role in extremal-configuration search. Across nine kissing-number and generalized kissing-number tasks, removing the Corrector consistently weakens the search trajectory (Fig.~\ref{fig:5}). The ablated system often improves rapidly at early episodes, but then plateaus at lower kissing numbers, whereas the full system continues to find stronger configurations. The difference is especially pronounced for \(K(13,1/2)\), \(K(14,1/2)\), \(K(18,1/3)\) and \(K(20,1/4)\). This ablation shows that the Corrector is essential for sustaining high-quality search in the expanding Gram-matrix completion space. By suppressing low-quality partial configurations, it improves sampling quality and helps guide the system toward globally coherent extremal structures. Thus, the Corrector is a central component of \textit{PackingStar}, not an auxiliary refinement.

\section*{Discussion}

The significance of \textit{PackingStar} lies not only in the new bounds it obtains, but in the object on which discovery operates. By moving from isolated extremal configurations to extremal configuration spaces, it turns the Kissing Number Problem into an experimental study of structured families of solutions. Configurations with the same extremal value can now be sampled, compared, decomposed and reused, revealing which structures are accidental and which have generative power across dimensions. This shift changes the role of AI in mathematical discovery. Unlike many AI-for-mathematics systems evaluated on benchmark problems with known answers, \textit{PackingStar} produces new mathematical objects that change the best-known bounds in a classical open problem. Rather than merely optimizing within known construction templates, reinforcement learning makes richer geometric objects computationally explorable. It identifies recurring blocks, high-symmetry representatives and cross-dimensional pathways that can guide further searches. The resulting records are therefore not only larger configurations, but evidence of reusable laws inside extremal configuration spaces.

At the same time, machine discovery does not replace proof. 
The structures proposed by \textit{PackingStar} require rigorous certification through linear programming, semidefinite programming, symmetry analysis and computer-assisted proof. Several configurations are proved optimal within their natural prescribed inner-product settings, while unrestricted global optimality remains open. This suggests a productive division of labour: AI exposes the hidden geometry of large search spaces, and proof turns the resulting patterns into mathematical knowledge. More broadly, \textit{PackingStar} points toward AI-assisted mathematics centered on spaces of objects rather than isolated solutions. Future progress will depend on integrating generative search with algebraic invariants and rigorous certificates, so that experimentally discovered structure can become a source of new theorems.

\section*{Methods}

We describe the overall algorithmic mechanism by which \textit{PackingStar} samples extremal configuration spaces through Gram-matrix completion. For notational simplicity, we write $\mathbf{G}^{(m)}$ for the current Gram matrix of a partially constructed configuration with $m$ sphere-center vectors. Its diagonal entries are fixed to $1$, and its off-diagonal entries encode the assigned pairwise cosine relations. Let $\mathbf{G}^{(m_0)}\in\mathbb{R}^{m_0\times m_0}$ be the initial Gram matrix. We formulate Gram-matrix completion as a two-player Markov game, corresponding to Fig.~\ref{fig:pipelines}. At state $\mathbf{G}^{(m)}$, Player 1, the Filler, samples a candidate cosine vector $\mathbf{g}$, which specifies the inner products between a new sphere-center vector and the existing $m$ vectors (Fig.~\ref{fig:pipelines}\textcolor{royalblue}{b}). Player 2, the Corrector, selects a retained index set $I$ to remove suboptimal rows and columns through its complement (Fig.~\ref{fig:pipelines}\textcolor{royalblue}{b}). The two agents share a team reward given by the number of rows of the final matrix, and are jointly optimized to construct larger kissing configurations.

\paragraph{Player 1 (Matrix Filler).}  
For each step $m \ge m_0$, Player 1 samples an action according to policy $\pi_1$, $\mathbf{g} \sim \pi_1(\cdot \mid \mathbf{G}^{(m)})$, where $\mathbf{g} \in \mathcal{A}^{(m)}$ to extend the current matrix $\mathbf{G}^{(m)} \in \mathbb{R}^{m \times m}$:
\begin{equation}
\mathbf{G}^{(m+1)}(\mathbf{g}) =
\begin{bmatrix}
\mathbf{G}^{(m)} & \mathbf{g} \\
\mathbf{g}^\top & 1
\end{bmatrix}.
\end{equation}
The constraints for candidate action set $\mathcal{A}^{(m)}$ depends on $m$,
\noindent For \textbf{$m < n$}
\begin{equation}
\mathcal{A}^{(m)} = \Big\{ \mathbf{g} \in \mathcal{C}_1^m \;\Big|\; 
\mathbf{G}^{(m+1)}(\mathbf{g}) \succeq 0, \;
\mathrm{rank}(\mathbf{G}^{(m+1)}(\mathbf{g})) = m+1 \Big\}.
\end{equation}

\noindent For \textbf{$m \ge n$},
split $\mathbf{g}$ into two parts:
\begin{equation}
\mathbf{g} =
\begin{bmatrix} \mathbf{g}^{(1)} \\ \mathbf{g}^{(2)} \end{bmatrix}, \quad
\mathbf{g}^{(1)} \in \mathbb{R}^{n\times 1}, \ \mathbf{g}^{(2)} \in \mathbb{R}^{(m-n)\times 1}.
\end{equation}
Define the top-left block 
$\mathbf{G}^{(m)}_{1:n,\,1:n} \in \mathbb{R}^{n \times n}$ and its Cholesky factorization 
$\mathbf{G}^{(m)}_{1:n,\,1:n} = \mathbf{M}\mathbf{M}^\top$, with pseudo-inverse $\mathbf{M}^+$.   
Let
\begin{equation}
\mathbf{G}' = \mathbf{G}^{(m)}_{\,n+1:m,\,1:n} \in \mathbb{R}^{(m-n)\times n}.
\end{equation}
Compute the remaining part:
\begin{equation}
\mathbf{g}^{(2)} = {\mathbf{G}'} (\mathbf{G}^{(m)}_{1:n,\,1:n})^+ \mathbf{g}^{(1)}\, 
\end{equation}
Then the candidate action set is
\begin{equation}
\mathcal{A}^{(m)} = \Big\{ \mathbf{g} = [\mathbf{g}^{(1)}; \mathbf{g}^{(2)}] \;\Big|\;
\mathbf{g}^{(1)} \in \mathcal{C}_1^n, \;
\|\mathbf{M}^+ \mathbf{g}^{(1)}\|_2 = 1, \;
\mathbf{g}^{(2)} \in \mathcal{C}_2^{\, m-n} \Big\}.
\end{equation}
$\mathcal{C}_2$ is a predefined subset of allowed cosine values.  
$\mathcal{C}_2$ may coincide with $\mathcal{C}_1$, differ from it, or represent a continuous constraint set in different settings, e.g.,
\begin{equation}
\mathcal{C}_2 = \Big\{ \mathbf{g}^{(2)} \in \mathbb{R}^{\,m-n} \;\Big|\; g^{(2)}_i \le 0.5, \ i = 1, \dots, m-n \Big\}.
\end{equation}
In specific cases, an additional constraint set $\mathcal{C}_*$ can be directly applied to the candidate action set,
\begin{equation}
\mathcal{A}^{(m)} \gets \mathcal{A}^{(m)} \cap \mathcal{C}_*
\end{equation}

\paragraph{Player 2 (Matrix Corrector).}  
Once Player 1 completes $M$ steps (or no feasible $\mathbf{g}$ remains), Player 2 samples an index set
\begin{equation}
I \sim \pi_2(\cdot \mid \mathbf{G}^{(M)})
\end{equation}
where $\pi_2(\cdot \mid \mathbf{G}^{(M)})$ denotes a probability distribution over subsets of \([M]\), and constructs the refined matrix
\begin{equation}
\mathbf{G}^{*} = \mathbf{G}^{(M)}_{I,I}.
\end{equation}

\paragraph{Game Dynamics and Cooperative Learning Process.}
The policy $\pi_2$ is learned to identify suboptimal entries filled by Player 1 and select a retained index set $I$, so that the corrected matrix $\mathbf{G}^{*}=\mathbf{G}^{(M)}_{I,I}$ can support further expansion in subsequent filling steps. Player 1 then resumes filling from $\mathbf{G}^{*}$, producing a new feasible action set $\mathcal{A}^{(|I|)}$, and the alternation continues:
\begin{equation}
\mathbf{G}^{(M)} \xrightarrow{\pi_2} \mathbf{G}^{*} \xrightarrow{\pi_1} \mathbf{G}^{(M')} \xrightarrow{\pi_2} \cdots .  
\end{equation}
The filling-correcting process continues until no further feasible entries exist for Player 1 and Player 2 cannot produce a better correction. Let $\mathbf{G}^{(M^*)}$ denote the final matrix. The cooperative reinforcement learning objective of \textit{PackingStar} is to maximize the shared team reward $R_{\mathrm{team}}=M^*$, where $M^*$ is the number of rows of the final matrix. A higher reward therefore corresponds to a larger kissing configuration.
Formally, the optimal joint policies $(\pi_1^*,\pi_2^*)$ are obtained by solving
\begin{equation}
(\pi_1^*, \pi_2^*) 
=
\arg\max_{\pi_1,\pi_2}
\;
\mathbb{E}_{\mathbf{g}_t\sim\pi_1,\ I_t\sim\pi_2}
\left[
M^*(\{\mathbf{g}_t\},\{I_t\})
\right],
\end{equation}
where the expectation accounts for the stochasticity of both policies, and
$M^*(\{\mathbf{g}_t\},\{I_t\})$ denotes the final matrix size obtained after all filling actions $\{\mathbf{g}_t\}$ and correcting actions $\{I_t\}$ generated under $\pi_1$ and $\pi_2$. In short, the policy of Player 1 is implemented as a tree-search learner, where node values are updated based on the team reward obtained after each episode. The policy of Player 2 is parameterized by a neural network and optimized via policy gradient. After each round, both the tree-search statistics of Player 1 and the policy parameters of Player 2 are jointly updated under the same team reward, enabling the two players to mutually improve and construct larger kissing configurations.

\paragraph{Algebraic Abstraction.}

To make \textit{PackingStar} explore at the level of extremal configuration spaces rather than individual matrix states, we introduce an algebraic abstraction of the search tree. We use \(p\)-adic invariants of the rational Gram form, following the classification framework of Conway and Sloane \cite{conway2013sphere}. The algebraic invariant provides the computational link between the matrix game and extremal configuration spaces. Distinct algebraic labels correspond to different structural categories, which in turn identify different extremal configuration spaces or distinct branches within them (Fig.~\ref{fig:pipelines}\textcolor{royalblue}{a}). Specifically, let
\begin{equation}
\mathcal{G}_m(n,\mathcal{C})
=
\left\{
\mathbf{G}\in\mathbb{S}_+^m:
G_{ii}=1,\ 
G_{ij}\in\mathcal{C}\ (i\ne j),\
\operatorname{rank}(\mathbf{G})=n
\right\}
\end{equation}
where $\mathbb{S}_+^m$ denotes the set of $m\times m$ positive semidefinite matrices. For $\mathbf{G}\in\mathcal{G}_m(n,\mathcal{C})$, define the set of its full-rank cores by
\begin{equation}
\operatorname{Core}(\mathbf{G})
=
\left\{
J\subseteq [m]:
|J|=n,\ 
\operatorname{rank}(\mathbf{G}_{J,J})=n
\right\}.
\end{equation}
Let $\phi$ be an algebraic quantity defined on full-rank principal submatrices. We require $\phi$ to be invariant over the full-rank cores of the same Gram matrix, which means
$
\phi(\mathbf{G}_{J,J})=\phi(\mathbf{G}_{K,K}),
\forall J,K\in\operatorname{Core}(\mathbf{G}).
$
Thus $\phi$ induces a well-defined invariant of the whole matrix,
$
\Phi(\mathbf{G})=\phi(\mathbf{G}_{J,J}), J\in\operatorname{Core}(\mathbf{G}).
$
The label $z=\Phi(\mathbf{G})$ defines an abstract construction category. Two full-rank matrices are assigned to the same abstract branch if and only if $\Phi(\mathbf{G})=\Phi(\mathbf{H})$.
The concrete search tree is therefore compressed into algebraic categories
$
\mathcal{Z}_m=\Phi\bigl(\mathcal{G}_m(n,\mathcal{C})\bigr).
$
During Filler search, each episode produces a trajectory
\begin{equation}
\tau=\left(\mathbf{G}^{(m_0)},\mathbf{G}^{(m_0+1)},\ldots,\mathbf{G}^{(M)}\right),
\end{equation}
with terminal size $M(\tau)=M$. Whenever a full-rank matrix $\mathbf{G}^{(m)}\in\mathcal{G}_m(n,\mathcal{C})$ is reached, its label $z_m=\Phi(\mathbf{G}^{(m)})$ is computed. For each category $z$, the Filler records the number of visits $N_z$ to this category and the average terminal size
\begin{equation}
\widehat{V}(z)
=
\frac{1}{N_z}
\sum_{(\tau,m):\,\Phi(\mathbf{G}^{(m)})=z}
M(\tau),
\end{equation}
where the sum is taken over all visits to category $z$, and $M(\tau)$ is the terminal matrix size of trajectory $\tau$. The estimate $\widehat{V}(z)$ predicts the completion potential of category $z$. If a newly generated full-rank matrix has a well-sampled label $z$ with low $\widehat{V}(z)$, the Filler terminates or deprioritizes the branch and backpropagates $\widehat{V}(z)$ to update the search nodes. Therefore, Algebraic Abstraction Search converts node-level search into category-level search, compressing exponentially many Gram matrices into finitely many algebraic branches and enabling early recognition of low-potential routes.

\paragraph{Matrix Decomposition and Initialization for Subsequent Games.}
Once a terminal matrix $\mathbf{G}^{(M)}$ is obtained, \textit{PackingStar} decomposes it into principal submatrices
\begin{equation}
\mathbf{G}^{(M)}_{J,J}, \quad J \subset \{1,2,\dots,M\},
\end{equation}
which capture geometric blocks inherited from the terminal configuration. These blocks are used as structured seeds for subsequent games. For each seed $\mathbf{G}^{*}=\mathbf{G}^{(M)}_{J,J}$, Player 1 resumes filling according to $\pi_1$, and Player 2 applies corrections through $\pi_2$, continuing the alternating process. Multiple seeds can initialize parallel games, allowing \textit{PackingStar} to explore diverse regions of the search space and transfer structural information across dimensions or cosine constraints (Fig.~\ref{fig:pipelines}\textcolor{royalblue}{d}).
This step turns terminal configurations into sources of new searches. A terminal matrix representing an element of $\widehat{\mathfrak{E}}(n,\mathcal{C},N_*)$ may yield substructures that seed other sampled extremal configuration spaces. Thus, matrix decomposition induces directed structural transitions between extremal configuration spaces, so that \textit{PackingStar} learns not only isolated configurations but the relations among them. More algorithm details can be found in the \textbf{Supplementary Information}.





%

%
%
%
%
%
%


\section*{Acknowledgments}
We are grateful to Professor Henry Cohn for his influential work on the Kissing Number Problem and spherical codes, which has been an important source of inspiration for this research. We also thank him for helpful discussions, insightful comments, and constructive suggestions on generalized kissing configurations. We thank Ao Li and Xingmeng Zhang for assistance with program optimization and computing-cluster deployment. We thank Mingzhi Wang and Juntao Dai for coordinating GPU resources. We also thank Chiyuan Wang and Xinmian Sun for early discussions on this topic. We also thank Prof. Kai Du for valuable advice on manuscript presentation.

\section*{Author contributions}
C.M. conceived and led the project. C.M. established the matrix completion game and developed the game-theoretic reinforcement learning method. T.T.Z. established geometric metrics and interpreted the properties of the configurations. 
M.L., T.T.Z., C.M. and Z.M. explored cosine relationships between configurations of different dimensions.
H.C. explored the training of geometric generative models for high-dimensional structures.
C.M., H.C. and M.L. modeled the process of cosine feature simulation and designed the simulation algorithm.
M.L., C.M. and T.T.Z. discovered the cosine feature of constructions via numerical simulation.
C.M., H.C., T.T.Z., M.L., P.L., Z.M. completed the formulation of matrix completion.
Z.M. provided practical insights and metrics for exploring lattices and error-correcting codes, particularly for the 31-dimensional construction.
C.M., P.L., M.L., T.T.Z. and H.C. developed the project code. C.M. and P.L. optimized the efficiency of overall system.
C.M., T.T.Z., P.L. and H.C. conducted the experiments.
C.M., T.T.Z., P.L., Z.M. and Y.Y. wrote the manuscript. 
T.T.Z. and C.M. completed the mathematical analysis sections in the Supplementary Information.
Y.Q. and Y.C. led and coordinated the system-engineering team, providing directions for optimizing key operators, parallel execution pipelines and training stability in large-scale searches.
Y.Q., Y.Y. and Y.C. provided computational resources and planned directions for improving parallel efficiency and system scalability.

\section*{Competing interests}
The authors declare no competing interests.

\section*{Data and Code availability}
All discovered configurations reported in this work, together with the verification code used to check their structural and geometric properties, are available in the \textit{PackingStar} repository at
\href{https://github.com/CDM1619/PackingStar}{https://github.com/CDM1619/PackingStar}.


\newpage



\renewcommand{\thefigure}{S\arabic{figure}}
\renewcommand{\thetable}{S\arabic{table}}
\renewcommand{\theequation}{S\arabic{equation}}
\renewcommand{\thepage}{S\arabic{page}}
\setcounter{figure}{0}
\setcounter{table}{0}
\setcounter{equation}{0}
\setcounter{page}{1} 







\clearpage 

\appendix
\begin{center}
\section*{Supplementary Information}
\end{center}
\renewcommand{\contentsname}{}
\tableofcontents
\newpage

\section{AI for mathematics and the position of \textit{PackingStar}}

Artificial intelligence has recently begun to contribute to mathematics through several complementary routes. One major direction is AI-assisted reasoning and formal proof. Neural models have been used to guide proof search in systems such as Lean, Isabelle or Metamath, where each generated step can be checked by a formal verifier \cite{polu2020generative,polu2022formal,zheng2021minif2f,xin2024deepseekprover}. Related systems combine neural guidance with symbolic reasoning in structured mathematical domains. AlphaGeometry solves Olympiad-level Euclidean geometry problems by coupling neural search with symbolic deduction, while AlphaProof and AlphaGeometry 2 further demonstrate the potential of reinforcement learning and formal reasoning on International Mathematical Olympiad problems \cite{trinh2024alphageometry,hubert2026alphaproof,chervonyi2025alphageometry2}. These works show that AI can search large reasoning spaces while preserving independently checkable correctness.

A second direction uses machine learning for mathematical pattern recognition. In this setting, models are trained on existing mathematical data to identify hidden regularities, suggest useful invariants or guide conjecture formation. For example, machine learning has revealed patterns in knot theory and representation theory that led to new conjectures and human proofs \cite{davies2021advancing}. This line of work highlights the value of AI as a tool for detecting structures that may be difficult to see directly, while the mathematical objects and data to be analysed are typically already available.

A third direction formulates discovery as constructive search over algorithms, programs or algebraic decompositions. AlphaTensor uses reinforcement learning to discover faster matrix multiplication algorithms \cite{fawzi2022alphatensor}. FunSearch combines large language models with evolutionary program search to improve constructions in extremal combinatorics \cite{romera2024funsearch}. AlphaEvolve extends this program-search paradigm to broader algorithmic and mathematical settings, including a new lower bound of 593 for the 11-dimensional kissing number problem \cite{novikov2025alphaevolve}. This last result is especially relevant here: it shows that general-purpose AI search can produce new constructions in the Kissing Number Problem itself, a centuries-old open problem at the foundation of discrete geometry and the local analogue of Hilbert's 18th problem. Together, these advances establish AI as a powerful tool for mathematical reasoning, pattern recognition and constructive search.

\textit{PackingStar} is complementary to this line of work and targets a different level of mathematical discovery. Whereas AlphaEvolve shows that general-purpose program search can produce a new kissing-number construction, \textit{PackingStar} is built around the internal geometry of the Kissing Number Problem itself. For the broader Kissing Number Problem landscape, target values are unknown in almost all dimensions, large labelled datasets are unavailable, and progress requires not only improving lower bounds but also finding structures beyond existing human-designed templates. \textit{PackingStar} addresses this by moving from individual constructions to extremal configuration spaces, the spaces of inequivalent configurations attaining the same extremal value under a fixed geometric regime. The exploration reveals mathematically legible and reusable structures, including visible decompositions, high-symmetry representatives, natural inner-product sets and algebraic connections with the Leech lattice, the Conway group and the Fischer group. Thus, beyond numerical improvements, \textit{PackingStar} provides a new way to study the Kissing Number Problem and its generalized variants through structured spaces of extremal configurations and the geometric laws that connect them. More broadly, it points to a paradigm for AI-assisted mathematics in which difficult open problems are advanced by making richer mathematical object spaces explorable, rather than only optimizing within existing formulations.

\newpage

\section{Reference data for kissing configurations}
This section collects baseline data for the kissing configurations discussed in the paper, including known kissing numbers and cosine sets used for comparison and verification.

\begin{table}[ht]
  \centering
  \setlength{\belowcaptionskip}{14pt}
  \caption{\textbf{Kissing number $K(n)$ in dimension $n$ for $1 \le n \le 24$} \cite{cohn2024table}. The bold items indicate the kissing numbers that are known to be optimal.}
  \label{tab:kissing-1-24}
  \begin{tabular}{cc @{\hspace{2.1em}} cc @{\hspace{2.1em}} cc @{\hspace{2.1em}} cc @{\hspace{2.1em}} cc @{\hspace{2.1em}} cc}
    \toprule
    $n$ & $K(n)$ & $n$ & $K(n)$ & $n$ & $K(n)$ & $n$ & $K(n)$ & $n$ & $K(n)$ & $n$ & $K(n)$ \\
    \midrule
     1 & \textbf{2}    & 5 & 40             & 9 & 306           & 13 & 1154    & 17 & 5730       & 21 & 29768 \\
     2 & \textbf{6}    & 6 & 72             & 10 & 510          & 14 & 1932    & 18 & 7654       & 22 & 49896 \\
     3 & \textbf{12}   & 7 & 126            & 11 & 593          & 15 & 2564    & 19 & 11692      & 23 & 93150 \\
     4 & \textbf{24}   & 8 & \textbf{240}   & 12 & 840          & 16 & 4320    & 20 & 19448      & 24 & \textbf{196560} \\
    \bottomrule
  \end{tabular}
\end{table}

\begin{table}[ht]
    \centering
    \setlength{\belowcaptionskip}{-9pt}
    \caption{\fontsize{11}{11}\selectfont
    \textbf{Cosine sets of different kissing configurations.} 
    For each configuration, the table lists the set of cosine values between distinct unit vectors.}
    \renewcommand{\arraystretch}{1.15} 
    \setlength{\tabcolsep}{16pt}
    \begin{tabular}{cc}
            \\
            \hline
            Configuration                & 
            Cosine set (excluding $-1$ and $0$) \\
            \hline
            $K_{\mathrm{r}}(13)$                 
                & $\{\pm \tfrac{1}{4},\ \pm \tfrac{1}{2}\}$; variants: $\{-\tfrac{3}{4},\ \pm \tfrac{1}{4},\ \pm \tfrac{1}{2}\}$  \\
            $K_{\mathrm{new}}(25)$                 
                & $\{\pm \tfrac{1}{6},\ \pm \tfrac{\sqrt{6}}{12},\ \pm \tfrac{1}{4},\ \pm \tfrac{1}{3},\ \pm \tfrac{\sqrt{6}}{6},\ \pm \tfrac{1}{2}\}$  \\
            $K_{\mathrm{new}}(26)$             
                & $\{-\tfrac{5}{6},\ -\tfrac{2}{3},\ \pm \tfrac{1}{6},\ \pm \tfrac{\sqrt{6}}{12},\ \pm \tfrac{1}{4},\ \pm \tfrac{1}{3},\ \pm \tfrac{\sqrt{6}}{6},\ \pm \tfrac{1}{2}\}$           \\
            $K_{\mathrm{new}}(27)$  
                & $\{-\tfrac{5}{6},\ -\tfrac{2}{3},\ \pm \tfrac{2\sqrt{3} - \sqrt{6}}{12},\ \pm \tfrac{1}{6},\ \pm \tfrac{\sqrt{6}}{12},\ \pm \tfrac{1}{4},\ \pm \tfrac{1}{3},\ \pm \tfrac{\sqrt{6}}{6},\ \pm \tfrac{2\sqrt{3} + \sqrt{6}}{12},\ \pm \tfrac{1}{2}\}$          \\
            $K_{\mathrm{new}}(28)$     
                & $\{-\tfrac{5}{6},\ -\tfrac{2}{3},\ \pm \tfrac{1}{6},\ \pm \tfrac{\sqrt{6}}{12},\ \pm \tfrac{1}{4},\ \pm \tfrac{1}{3},\ \pm \tfrac{\sqrt{6}}{6},\ \pm \tfrac{1}{2}\}$                    \\
            $K_{\mathrm{new}}(29)$ 
                & $\{-\tfrac{5}{6},\ -\tfrac{2}{3},\ \pm \tfrac{2\sqrt{3} - \sqrt{6}}{12},\ \pm \tfrac{1}{6},\ \pm \tfrac{\sqrt{6}}{12},\ \pm \tfrac{1}{4},\ \pm \tfrac{\sqrt{3}}{6},\ \pm \tfrac{1}{3},\ \pm \tfrac{\sqrt{6}}{6},\ \pm \tfrac{2\sqrt{3} + \sqrt{6}}{12},\ \pm \tfrac{1}{2}\}$             \\
            $K_{\mathrm{new}}(30)$      
                &  $\{-\tfrac{5}{6},\ -\tfrac{2}{3},\ \pm \tfrac{1}{6},\ \pm \tfrac{\sqrt{6}}{12},\ \pm \tfrac{1}{4},\ \pm \tfrac{1}{3},\ \pm \tfrac{\sqrt{6}}{6},\ \pm \tfrac{1}{2}\}$        \\
            $K_{\mathrm{new}}(31)$        
                &  The full cosine set is available at the data repository.    \\
            \hline
    \end{tabular}
    \label{tab:cosset}
\end{table}

\section{Method details}
\label{sec:method-details}

This section provides supplementary methodological details of the \textit{PackingStar} system. We describe the construction of cosine sets, the cooperative matrix-completion game, Algebraic Abstraction Tree Search (AATS), the permutation-equivariant Corrector architecture and its training objective, the distributed implementation of the training pipeline, and the decomposition procedure that transforms completed Gram matrices into initializations for subsequent games.

\subsection{How to identify cosine set}

The matrix-completion game requires a finite set of cosine values to constrain the candidate matrix entries. As these values are generally not known a priori, we first perform a preliminary exploration in coordinate space to identify the empirical cosine patterns that emerge in large kissing configurations. The resulting observations are used to construct the downstream game constraints $\mathcal C_1$ and, when necessary, $\mathcal C_2$.

Given $m$ existing unit-sphere centers in $\mathbb{R}^n$ with $m \ge n-1$, we iteratively construct new candidate centers by examining all $(n-1)$-subsets of the current configuration. For a selected subset with coordinates $\mathbf{y}_1, \dots, \mathbf{y}_{n-1} \in \mathbb{R}^n$, we define
$\mathbf{A} = [\mathbf{y}_1^\top; \dots; \mathbf{y}_{n-1}^\top] \in \mathbb{R}^{(n-1)\times n}$,
where $\mathbf{A}$ is required to have full row rank. Candidate sphere centers $\mathbf{x} \in \mathbb{R}^n$ are then obtained by solving
\begin{equation}
\begin{cases}
\mathbf{A}\mathbf{x} = \mathbf{b}, \\
\|\mathbf{x}\|_2 = 1,
\end{cases}
\qquad
\mathbf{b} = \tfrac{1}{2}\mathbf{1}_{n-1}.
\end{equation}
The feasible solutions take the form
\begin{equation}
\mathbf{x}
=
\mathbf{A}^+\mathbf{b}
\pm
\sqrt{1-\|\mathbf{A}^+\mathbf{b}\|_2^2}\,\mathbf{z},
\qquad
\mathbf{z}\in\ker(\mathbf{A}),
\quad
\|\mathbf{z}\|_2=1.
\end{equation}

All feasible candidates generated in this manner enter the coordinate-level search space. Because the number of possible continuation paths grows rapidly, we use Monte Carlo Tree Search (MCTS) to determine which candidate should be accepted next. Each node in the search tree represents a partial configuration
$s_t=\{\mathbf{x}_1,\mathbf{x}_2,\dots,\mathbf{x}_{m_t}\}\subset\mathbb{R}^n$
containing $m_t$ sphere centers at search step $t$. MCTS selects the next candidate $\mathbf{x}_t$ according to the upper-confidence-bound criterion,
\begin{equation}
\mathbf{x}_t
=
\arg\max_{\mathbf{x}}
\left[
Q(s_t,\mathbf{x})
+
c\sqrt{\frac{\ln N(s_t)}{N(s_t,\mathbf{x})}}
\right],
\end{equation}
where $Q(s_t,\mathbf{x})$ denotes the estimated value, $N(s_t)$ and $N(s_t,\mathbf{x})$ denote visit counts, and $c>0$ controls the exploration-exploitation trade-off. The reward, defined as the total number of spheres in the resulting configuration, is backpropagated through the search tree to favor configurations with larger kissing numbers.

Exploration continues until no additional feasible $\mathbf{x}$ can be generated under the construction procedure. From the resulting configurations, we extract the empirical set of pairwise cosines observed between accepted sphere centers,
\begin{equation}
\widehat{C}_{\rm coord}
=
\{\langle \mathbf{x}_i,\mathbf{x}_j\rangle:
i\ne j,\ \mathbf{x}_i,\mathbf{x}_j
\text{ appear in the explored configurations}\}.
\end{equation}
These empirical regularities subsequently define the finite cosine constraints for the matrix-completion game instance,
\begin{equation}
\mathcal I=(n,\mathbf G^{(m_0)},\mathcal C_1,\mathcal C_2,\mathcal C_\ast),
\end{equation}
where $\mathcal C_1$ denotes the primary cosine set for Filler actions, and $\mathcal C_2$ specifies the constraint on the remaining entries after a full-rank core has been fixed (Section~\ref{sec:filler-corrector}). The coordinate-space exploration therefore establishes the cosine constraints for \textit{PackingStar} games.

\subsection{The Two-player Cooperative Game}
\label{sec:filler-corrector}

A game instance is specified by
\begin{equation}
\mathcal I = (n,\mathbf G^{(m_0)},\mathcal C_1,\mathcal C_2,\mathcal C_\ast).
\end{equation}
The game state is represented by the current partial Gram matrix $\mathbf G^{(m)}$.  The Filler and Corrector define two coupled transitions on this state space.
A Filler action is a cosine entry
$\mathbf g\in\textsc{CandidateActions}(\mathbf G^{(m)};\mathcal I)$ which
extends the current state by one row and column according to
\begin{equation}
\operatorname{Extend}(\mathbf G^{(m)},\mathbf g)
=
\begin{bmatrix}
\mathbf G^{(m)} & \mathbf g\\
\mathbf g^\top & 1
\end{bmatrix}.
\end{equation}
The candidate action set enforces the positive semi-definiteness, rank, cosine, and structural constraints for the current gram matrix and game instance.
A Corrector action is a retained index list $I$ that replaces the current matrix with submatrix $\mathbf G^{(m)}_{I,I}$.
Following the matrix-completion process, the leading block $n\times n$ naturally becomes a full-rank core once the matrix reaches size $n$. The corrector always keeps the core 
to avoid losing the full-rank property and deviating from the current algebraic class and the extremal configuration space. 

The game is cooperative in the sense that both agents are evaluated according to the size of the feasible configuration obtained after filling and correction. The Filler tree receives value backups from terminal row counts, whereas the Corrector receives a remove-and-repack reward determined by the change in row count after the corrected matrix is subsequently repacked by the Filler. The running best state $\mathbf G^*$ records the largest feasible Gram matrix discovered under the constraints of the current game instance.

For compact and consistent pseudocode, we group the game setting and
hyperparameters into configuration objects:
\begin{equation}
\begin{aligned}
\mathcal I
&=(n,\mathbf G^{(m_0)},\mathcal C_1,\mathcal C_2,\mathcal C_\ast),\\
\xi_{\rm AATS}
&=(\tau,N_{\rm prune},\mu_{\rm prune},\sigma_{\rm prune}^2,r),\\
\xi_{\rm net}
&=(L_{\rm att},L_{\rm enc},d,h,r_{\rm ffn},\mathcal J_0),\\
\xi_{\rm train}
&=(T,K,N_{\rm traj},\gamma,\lambda_{\rm GAE},
\epsilon_{\rm clip},\epsilon_V,c_V,c_H,\alpha_{\rm supcon},\tau_{\rm supcon},
E,N_{\rm mini},\eta).
\end{aligned}
\end{equation}
Here, $\mathcal I$ defines the matrix-completion game instance: $n$ is the ambient
dimension, $\mathbf G^{(m_0)}$ is the seed matrix,
$\mathcal C_1$ controls the cosine values that appear in the candidate entries, $\mathcal C_2$ constrains the remaining pairwise cosine values, and $\mathcal C_\ast$ denotes any additional
structural constraint.  The Algebraic Abstraction Tree Search(AATS) configuration $\xi_{\rm AATS}$ controls tree
exploration, algebraic-class pruning and best-of-$r$ rollout.  The network
configuration $\xi_{\rm net}$ controls the Corrector architecture and the
protected rows $\mathcal J_0$.  The training configuration $\xi_{\rm train}$
controls the finite-horizon remove-repack refinement episodes and the on-policy optimization
procedure.  Throughout the pseudocode, $\theta$ denotes the trainable
parameters of the Corrector policy $\pi_\theta$, and
$\mathcal S[z]=(N_z,\mu_z,\sigma_z^2)$ denotes the return statistics attached
to the algebraic label $z$. Table~\ref{tab:hyperparams} details how symbolic hyperparameters are grouped
throughout the pseudocode.

Algorithm~\ref{alg:cotrain} presents the sequential form of the cooperative game pipeline. Each trajectory attempt begins with AATS traversing the Filler tree. If the reached algebraic class has already entered the pruning regime, the corresponding class mean is immediately backpropagated and no Corrector episode is initiated. Otherwise, the Filler reaches a leaf node through rollout, yielding a matrix $\mathbf G_0$ that serves as the initial state of the subsequent Corrector--Filler refinement episode.
A total of $T$ remove-and-repack transitions are then performed. Each transition applies \textsc{Corrector.Remove} followed by \textsc{Filler.BestRollout}, producing the transition reward
\begin{equation}
\textsc{Reward}\bigl(
\mathrm{rows}(\mathbf G_{t+1})
-
\mathrm{rows}(\mathbf G_t)
\bigr).
\end{equation}
The observed terminal row count is subsequently backpropagated along the root-to-node trajectory. The distributed realization in Algorithm~\ref{alg:distributed} implements the same logical loop by serializing shared tree operations and updates while executing rollout and correction procedures across parallel workers.

Table~\ref{tab:hyperparams} summarizes how symbolic hyperparameters are grouped
throughout the pseudocode.

\begin{table}[ht]
    \centering
    \caption{\textbf{Configuration objects and symbolic hyperparameters.}}
    \label{tab:hyperparams}
    \begin{tabular}{p{0.18\linewidth}p{0.35\linewidth}p{0.37\linewidth}}
        \toprule
        \textbf{Object} & \textbf{Symbols} & \textbf{Role} \\
        \midrule
        $\mathcal I$ & $n,\mathbf G^{(m_0)},\mathcal C_1,\mathcal C_2,\mathcal C_\ast$ & Fixed matrix-completion problem instance and constraints \\
        $\xi_{\rm AATS}$ & $\tau,N_{\rm prune},\mu_{\rm prune},\sigma_{\rm prune}^2,r$ & Tree exploration, algebraic-class pruning, and best-of-$r$ rollout budget \\
        $\xi_{\rm net}$ & $L_{\rm att},L_{\rm enc},d,h,r_{\rm ffn},\mathcal J_0$ & Corrector architecture and deterministic protected rows \\
        $\xi_{\rm train}$ &
        \begin{tabular}{@{}l@{}}
        $T,K,N_{\rm traj},\gamma,\lambda_{\rm GAE},\epsilon_{\rm clip},\epsilon_V$\\
        $c_V,c_H,\alpha_{\rm supcon},\tau_{\rm supcon}$\\
        $E,N_{\rm mini},\eta$
        \end{tabular}
        & Episode horizon, trajectory batch size, PPO/GAE, contrastive learning loss, epochs, mini-batches, and learning rate \\
        $\mathcal W$ & worker pool and synchronization policy & Distributed sampling and serialized tree/statistics updates \\
        \bottomrule
    \end{tabular}
\end{table}

\begin{algorithm}[ht]
\LinesNumbered
\SetKwInOut{Require}{Require}
\SetKwInOut{Ensure}{Ensure}
\Require{Problem instance $\mathcal I$; AATS config $\xi_{\rm AATS}$; Corrector config $\xi_{\rm net}$; training config $\xi_{\rm train}$; algebraic-class-statistics table $\mathcal S$}
\Ensure{Trained Corrector policy $\pi_\theta$, best configuration $\mathbf G^*$}
\BlankLine
$\mathcal I=(n,\mathbf G^{(m_0)},\mathcal C_1,\mathcal C_2,\mathcal C_\ast)$\;
$\xi_{\rm train}=(T,K,N_{\rm traj},\ldots)$, $\xi_{\rm AATS}=(\ldots,r)$\;
Initialize Algebraic abstraction tree $\mathcal T$ with root $\mathbf G^{(m_0)}$\tcp*{Section~\ref{sec:mcts}}
Initialize Corrector policy $\pi_\theta$ using $\xi_{\rm net}$ \tcp*{Section~\ref{sec:corrector-arch}}
$\mathcal O\leftarrow\textsc{BuildOptimizer}(\theta;\xi_{\rm train})$\tcp*{Section~\ref{sec:optimization}}
$\mathbf G^* \leftarrow \mathbf G^{(m_0)}$\;

\BlankLine
\For{$\text{iter} = 1, 2, \ldots, K$}{
    $\mathcal B \leftarrow \emptyset$\;
\For{$j=1,\ldots,N_{\rm traj}$}{
            $\mathbf G_{\rm leaf},\,\mathcal P,\,h
            \leftarrow \textsc{Filler.Traverse}(\mathcal T,\mathcal S;\mathcal I,\xi_{\rm AATS})$\;
            \If{$h=\textsc{EarlyBackprop}$}{
            \textbf{continue} to the next trajectory attempt\;
            }
            $\mathbf G_0
            \leftarrow \textsc{Filler.BestRollout}(\mathbf G_{\mathrm{leaf}};\mathcal I,\xi_{\rm AATS})$\;
            $\tau_j\leftarrow[\,]$\;

            \For{$t = 0, \ldots, T-1$}{
                $\mathbf G_t', I_t,\log\pi_t^{\rm old},V_t^{\rm old}
                \leftarrow \textsc{Corrector.Remove}(\mathbf G_t;\pi_\theta,\mathcal I,\xi_{\rm net})$\;

                $\mathbf G_{t+1} \leftarrow \textsc{Filler.BestRollout}(\mathbf G_t';\mathcal I,\xi_{\rm AATS})$\;

                $\Delta m_t \leftarrow
                \mathrm{rows}(\mathbf G_{t+1})-\mathrm{rows}(\mathbf G_t)$\;
                $r_t \leftarrow \textsc{Reward}(\Delta m_t)$\;
                $d_t\leftarrow(t=T-1)$\;
                Append $(\mathbf G_t,I_t,\log\pi_t^{\rm old},
                V_t^{\rm old},r_t,d_t)$ to $\tau_j$\;
            }
            $\mathcal B\leftarrow\mathcal B\cup\{\tau_j\}$\;
            $\textsc{Filler.Backpropagate}(\mathcal T,\mathcal P,\mathrm{rows}(\mathbf G_T),\mathcal S;\mathcal I)$\;
            \If{$\mathrm{rows}(\mathbf G_T) > \mathrm{rows}(\mathbf G^*)$}{
                $\mathbf G^* \leftarrow \mathbf G_T$\;
            }
    }
    $(\pi_\theta,\mathcal O) \leftarrow
    \textsc{Corrector.Update}(\pi_\theta,\mathcal O,\mathcal B;\mathcal I,\xi_{\rm net},\xi_{\rm train})$ \tcp*{Section~\ref{sec:training}}
}

\caption{Filler-Corrector co-training pipeline.}
\label{alg:cotrain}
\end{algorithm}

\subsection{Algebraic Abstraction Tree Search}
\label{sec:mcts}

Algebraic Abstraction Tree Search (AATS) serves as the search algorithm for the Filler and consists of two coupled layers. The first layer is a prefix tree whose edges correspond to Filler actions and whose nodes represent partial Gram matrices obtained by replaying these actions from the seed state. This layer maintains the standard tree-search statistics and determines which partial matrix is expanded at each step.
The second layer is an algebraic-class statistics table indexed by the label
$z=\textsc{ComputeAlgebraicLabel}(\mathbf G)$,
which is computed once a trajectory reaches a full-rank Gram core. This layer aggregates terminal returns across paths sharing the same algebraic label and enables early backpropagation pruning. Together, the two layers form the UCT backbone while allowing statistical sharing across algebraically equivalent search paths.

\subsubsection{Tree data structure}
\label{sec:tree-structure}

The Filler search tree is a rooted prefix tree over Filler actions. A node $v$ corresponds to a matrix $\mathbf{G}(v)$ of size $m(v)$. The root $v_{m_0}$ represents the initial seed Gram matrix $\mathbf{G}^{(m_0)}$, with $m(v_{m_0}) = m_0$ and $\mathbf{G}(v_{m_0}) = \mathbf{G}^{(m_0)}$.

Each node can be uniquely represented by the ordered sequence of Filler actions applied from the root.
Specifically, every non-root node $v$ has a parent $\operatorname{par}(v)$ and is generated by an incoming action $\ell(v)$, yielding
\begin{equation}
\mathbf G(v)=
\operatorname{Extend}\!\bigl(\mathbf G(\operatorname{par}(v)),\ell(v)\bigr)
=
\begin{bmatrix}
\mathbf G(\operatorname{par}(v)) & \ell(v)\\
\ell(v)^\top & 1
\end{bmatrix}.
\label{eq:node-reconstruction}
\end{equation}
As a result, a node $v$ is completely determined by its root-to-node action sequence
\begin{equation}
\operatorname{path}(v)=
\bigl(\mathbf g_{m_0},\mathbf g_{m_0+1},\ldots,\mathbf g_{m(v)-1}\bigr),
\end{equation}
where \(\mathbf g_j\) denotes the action extending a size-\(j\) matrix to size \(j+1\). The root corresponds to the empty sequence, \(\operatorname{path}(v_{m_0})=\emptyset\).
This representation eliminates the need to store explicit Gram matrices at every node. Instead, \(\mathbf G(v)\) is reconstructed on demand by replaying the recorded action sequence from the seed matrix \(\mathbf G^{(m_0)}\).

The persistent state stored at each node is minimal.  For a node
$v_m$, the tree maintains only its visit count $N(v_m)$, accumulated value
$Q(v_m)$, and child-link map,
\begin{equation}
\mathrm{Ch}(v_m):\mathcal E(v_m)\to\mathcal V,
  \qquad
  \mathbf g\mapsto \mathrm{Ch}(v_m)[\mathbf g],
\end{equation}
where $\mathcal V$ denotes the set of child nodes of $v_m$, and
$\mathcal E(v_m)$ denotes the set of materialized actions available at  $v_m$.  For any
$\mathbf g\in\mathcal E(v_m)$, the corresponding child
$u_{\mathbf g}=\mathrm{Ch}(v_m)[\mathbf g]$ satisfies
\begin{equation}
\operatorname{par}(u_{\mathbf g})=v_m,\qquad
m(u_{\mathbf g})=m+1,\qquad
\ell(u_{\mathbf g})=\mathbf g,\qquad
\mathbf G(u_{\mathbf g})=\operatorname{Extend}(\mathbf G(v_m),\mathbf g).
\end{equation}

The tree is implemented as a recursive dictionary $\mathcal T$
rooted at $v_{m_0}$.  The symbol $v_m$ denotes the node identifier, whereas
$\mathcal T[v_m]$ denotes its stored record:
\begin{equation}
\mathcal T[v_m]=
\left\{
\begin{array}{ll}
\texttt{state}: & [\,N(v_m),Q(v_m)\,],\\[1mm]
\texttt{children}: &
\{\,\mathbf g:\mathcal T[u]
\;|\; \mathbf g\in\mathcal E(v_m), u=\mathrm{Ch}(v_m)[\mathbf g] \,\}.
\end{array}
\right.
\end{equation}
The recursion unfolds among existing edges:
\begin{equation}
\mathbf g_j  \in\mathcal E(v_j), \quad
v_{j+1}=\mathrm{Ch}(v_j)[\mathbf g_j], \quad
\mathcal T[v_{j+1}] = \mathcal T[v_j].\texttt{children}[\mathbf g_j].
\end{equation}
For example, traversing node $v_m$ along edge $\mathbf g_m$ leads to the child node $v_{m+1} = \mathrm{Ch}(v_m)[\mathbf g_m]$. The corresponding stored record $\mathcal T[v_{m+1}]$ is obtained via $\mathcal T[v_m].\texttt{children}[\mathbf g_m]$ and has the same internal structure as $\mathcal T[v_m]$.

\subsubsection{Child selection and node expansion}
\label{sec:traverse-expand}

Algorithms~\ref{alg:select-child} and \ref{alg:expansion} define the two basic
operations by which Filler traverses the prefix tree. Built on a UCT backbone, 
their purpose is to allocate search in a high-branching tree while
balancing exploitation and exploration.  At a node $v$ of size $m$, let
$\mathcal A_v=\textsc{CandidateActions}(\mathbf G(v);\mathcal I)$
denote the candidate actions defined by the current Gram matrix and
problem instance.  Each Filler action is a length-$m$ cosine vector $\ell(v)\in\mathbb R^m$.  Even after positive semi-definiteness, rank, cosine set, and
optional structural constraint in $\mathcal C_\ast$ are imposed, $\mathcal A_v$ can
remain combinatorially large.  AATS therefore cannot instantiate all
admissible children of $v$ when node $v$ is expanded.  Only the
sampled edge labels $\mathcal E(v)\subseteq\mathcal A_v$ are recorded, keeping the remaining child nodes implicit.

On this partially materialized tree, the Filler can either choose an unexplored action to materialize a new child node or select an existing child node. 
\textsc{SelectChild} uses a modified Upper Confidence Bound to decide between these two options.  
For each materialized edge label
$\mathbf g\in\mathcal E(v)$ with child
$u_{\mathbf g}=\mathrm{Ch}(v)[\mathbf g]$, the score is
\begin{equation}
\mathcal U_{\rm ch}(v,\mathbf g)
=
\frac{Q(u_{\mathbf g})}{N(u_{\mathbf g})+1}
+\tau
\sqrt{
\frac{\log(N(v)+1)}{N(u_{\mathbf g})+1}
}.
\end{equation}
For the unexplored part of the action space, we use the virtual
statistics $(N,Q)=(0,0)$ and compute the expansion score as
\begin{equation}
\mathcal U_{\rm ex}(v)=\tau\sqrt{\log(N(v)+1)} .
\end{equation}
The denominators $N(u_{\mathbf g})+1$ implement a finite priority
for the unmaterialized child nodes.  Without this convention, the search would
be forced to enumerate every possible child node before revisiting a promising
branch, effectively reducing to a breadth-first tree search of a large combinatorial action space.  In this way, a new child is materialized
only when that region becomes competitive with the best existing child.

When the node expansion move is selected, we sample candidate actions directly rather than constructing the exact unvisited set
$\mathcal A(\mathbf G(v))\setminus\mathcal E(v)$. If the sampled action $\mathbf g$  belongs to $\mathcal E(v)$ which has already been
materialized, the child map resolves it to an existing
node. Operationally, this amounts to an additional traversal step. If an unexplored action $\mathbf g$ is sampled, we materialize a new edge and a new node $v_{new}$. 
This shortcut trades off a small amount of exploration efficiency for a large reduction in computational cost in filtering out visited nodes, which is critical for a large candidate action space.

\SetKwProg{Fn}{Function}{:}{}
\begin{algorithm}[ht]
\caption{\textsc{Filler.SelectChild}}
\label{alg:select-child}
\KwIn{current node record $\mathcal T_{\rm cur}$; current gram matrix $\mathbf G$; current node $v$; problem instance $\mathcal I$; AATS config $\xi_{\rm AATS}$}
\KwOut{selected action $\mathbf g^\star$; selected child node $u_{\mathbf g^\star}$; child record $\mathcal T[u_{\mathbf g^\star}]$; child Gram matrix $\mathbf G_{\mathbf g^\star}$; operation $ \mathbf{o} \in\{\textsc{Selection},\textsc{Expansion}\}$}
\Fn{\textsc{Filler.SelectChild}$(\mathcal T_{\rm cur}, \mathbf G, v;\mathcal I,\xi_{\rm AATS})$}{
    $\mathcal I=(n,\mathbf G^{(m_0)},\mathcal C_1,\mathcal C_2,\mathcal C_\ast)$, $\xi_{\rm AATS}=(\tau,\ldots)$\;
    $N(v), Q(v) \leftarrow \mathcal T_{\rm cur}.\texttt{state}$\;
    $\mathcal U_{\rm ch} \leftarrow \emptyset$ \;
    \ForEach{$\mathbf g\in\mathcal E(v)$}{
      $u_{\mathbf g}\leftarrow \mathrm{Ch}(v)[\mathbf g]$\;
      $\mathcal T[u_{\mathbf g}]\leftarrow
      \mathcal T_{\rm cur}.\texttt{children}[\mathbf g]$\;
      $N(u_{\mathbf g}),Q(u_{\mathbf g}) \leftarrow \mathcal T[u_{\mathbf g}].\texttt{state}$\;
      $\mathcal U_{\rm ch}[\mathbf g]\leftarrow
      \dfrac{Q(u_{\mathbf g})}{N(u_{\mathbf g})+1}
      +\tau\sqrt{\dfrac{\log(N(v)+1)}{N(u_{\mathbf g})+1}}$\;
      }
    $\mathbf g^\star\leftarrow
      \arg\max_{\mathbf g\in\mathcal E(v)} \mathcal U_{\rm ch} [\mathbf g]$\;
    $s_{\rm ch}^\star \leftarrow \mathcal U_{\rm ch}[\mathbf g^\star]$\;
    
    $s_{\rm ex}  \leftarrow \tau\sqrt{\log(N(v)+1)}$\;

    \If{$s_{\rm ex}\ge s_{\rm ch}^\star$}{
        $\mathcal A\leftarrow \textsc{CandidateActions}(\mathbf G;\mathcal I)$\;
        Sample $\mathbf g^\star\sim\mathrm{Uniform}(\mathcal A)$\;
        \If{$\mathbf g^\star \notin \mathcal E(v)$}{
            $u_{\mathbf g^\star},\mathcal T[u_{\mathbf g^\star}], \mathbf G_{\mathbf g^\star} \leftarrow \textsc{Filler.Expansion}(\mathcal T_{\rm cur}, \mathbf G, v, \mathbf g^\star)$\;
            \Return $\mathbf g^\star,u_{\mathbf g^\star}, \mathcal T[u_{\mathbf g^\star}], \mathbf G_{\mathbf g^\star}, \textsc{Expansion}$\;
        }
    }
    $u_{\mathbf g^\star}\leftarrow \mathrm{Ch}(v)[\mathbf g^\star]$\;
    $\mathcal T[u_{\mathbf g^\star}] \leftarrow \mathcal T_{\mathrm{cur}}.\texttt{children}[\mathbf g^\star]$ \;
    $\mathbf G_{\mathbf g^\star} \leftarrow \operatorname{Extend}(\mathbf G,\mathbf g^\star)$\;
    \Return $\mathbf g^\star,u_{\mathbf g^\star},\mathcal T[u_{\mathbf g^\star}],\mathbf G_{\mathbf g^\star}, \textsc{Selection}$\;
}
\end{algorithm}

When \textsc{SelectChild} chooses the expansion move, the control passes to \textsc{Filler.Expansion}
(Algorithm~\ref{alg:expansion}), which materializes a new child node from the sampled action, registers it in the recursive child map, and extends the Gram matrix by one row and column.  

\begin{algorithm}[ht]
\caption{\textsc{Filler.Expansion}}
\label{alg:expansion}
\KwIn{current node record $\mathcal T_{\rm cur}$; current gram matrix $\mathbf G$; current node $v$; sampled action $\mathbf g$}
\KwOut{new child node $v_{\rm new}$; new child record $\mathcal T[v_{\rm new}]$; expanded Gram matrix $\mathbf G_{\rm new}$}
\Fn{\textsc{Filler.Expansion}$(\mathcal T_{\rm cur}, \mathbf G, v, \mathbf g)$}{
    Create a new node $v_{\rm new}$ with
    $\operatorname{par}(v_{\rm new})=v$ and $\ell(v_{\rm new})=\mathbf g$\;
    $\mathrm{Ch}(v)[\mathbf g]\leftarrow v_{\rm new}$,
    $\mathcal E(v)\leftarrow \mathcal E(v)\cup\{\mathbf g \}$\;
    $\mathcal T[v_{\rm new}]
    \leftarrow\{\texttt{state}:[0,0],\,\texttt{children}:\emptyset\}$\;
    $\mathbf G_{\rm new}\leftarrow
    \operatorname{Extend}(\mathbf G,\mathbf g)$\;
    $\mathcal T_{\rm cur}.\texttt{children}[\mathbf g] \leftarrow \mathcal T[v_{\rm new}]$\;
    \Return $v_{\rm new},\mathcal T[v_{\rm new}],\mathbf G_{\rm new}$\;
}
\end{algorithm}

\subsubsection{Rollout}
\label{sec:aats-rollout}

Rollout is the stochastic completion step that saturates a partial Gram
matrix.  It is invoked when AATS reaches a leaf node and after the Corrector has
removed rows from the current configuration. Starting from a
prefix $\mathbf G$, the Filler randomly samples a single action from
$\mathcal A(\mathbf G;\mathcal I)$  and extends the matrix.
The rollout terminates when the current gram matrix admits
no further extension.

The plain rollout in Algorithm~\ref{alg:rollout} performs a sequence of 
single-action extensions until no more candidate actions remain. At each step, the
candidate actions are recomputed for the updated Gram matrix, ensuring that all positive
semi-definiteness, rank, cosine, and structural constraints in $\mathcal I$ are satisfied.

\begin{algorithm}[ht]
\caption{\textsc{Filler.Rollout}: plain random greedy rollout}
\label{alg:rollout}
\KwIn{starting Gram matrix $\mathbf G$; problem instance $\mathcal I$; optional initial candidate action set $\mathcal A_0$}
\KwOut{terminal Gram matrix $\mathbf G$}
\Fn{\textsc{Filler.Rollout}$(\mathbf G;\mathcal I,\mathcal A_0=\mathrm{None})$}{
    $\mathcal I=(n,\mathbf G^{(m_0)},\mathcal C_1,\mathcal C_2,\mathcal C_\ast)$\;
    \eIf{$\mathcal A_0=\mathrm{None}$}{
        $\mathcal A\leftarrow \textsc{CandidateActions}(\mathbf G;\mathcal I)$\;
    }{
        $\mathcal A\leftarrow \mathcal A_0$\;
    }
    \While{$\mathcal A\neq\emptyset$}{
        Sample $\mathbf g\sim{\rm Uniform}(\mathcal A)$\;
        $\mathbf G\leftarrow\operatorname{Extend}(\mathbf G,\mathbf g)$\;
        $\mathcal A\leftarrow
        \textsc{CandidateActions}(\mathbf G;\mathcal I)$\;
    }
    \Return $\mathbf G$\;
}
\end{algorithm}

However, this stochastic completion process can terminate early due to an unfavorable random choice.
This randomness makes the value estimation for the root-to-node search paths given by the Filler 
and the corrected matrices $\mathbf G'$ given by the Corrector unreliable.
To address this problem,  Algorithm~\ref{alg:filler-best-rollout}  runs $r$ independent plain rollout
and returns the largest gram matrix.
By reducing the variance and allocating enough sampling budget,
the final reward and value estimation can reflect the quality of the correction and the potential of the leaf node more faithfully than a single random completion.

\begin{algorithm}[ht]
\caption{\textsc{Filler.BestRollout}: best-of-$r$ shared-prefix rollout}
\label{alg:filler-best-rollout}
\KwIn{post-Corrector Gram matrix $\mathbf G'$; problem instance $\mathcal I$; AATS config $\xi_{\rm AATS}$}
\KwOut{best terminal Gram matrix $\mathbf G_{\rm best}$}
\Fn{\textsc{Filler.BestRollout}$(\mathbf G';\mathcal I,\xi_{\rm AATS})$}{
    $\xi_{\rm AATS}=(\ldots,r)$\;
    $\mathcal A'\leftarrow \textsc{CandidateActions}(\mathbf G';\mathcal I)$\;
    $\mathbf G_{\rm best}\leftarrow\mathbf G'$\;
    \For{$j=1,\ldots,r$}{
        $\widetilde{\mathbf G}\leftarrow\operatorname{Clone}(\mathbf G')$\;
        $\widetilde{\mathcal A}\leftarrow\operatorname{Clone}(\mathcal A')$\;
        $\widetilde{\mathbf G}\leftarrow
        \textsc{Filler.Rollout}(\widetilde{\mathbf G};\mathcal I,\widetilde{\mathcal A})$\;
        \If{$\mathrm{rows}(\widetilde{\mathbf G}) > \mathrm{rows}(\mathbf G_{\rm best})$}{
            $\mathbf G_{\rm best}\leftarrow\widetilde{\mathbf G}$\;
        }
    }
    \Return $\mathbf G_{\rm best}$\;
}
\end{algorithm}

\subsubsection{Algebraic-invariant-based tree abstraction}
\label{sec:category-filter}

In the two-player matrix-completion game, a tree traversal and subsequent remove-and-repack transitions form an episode. The final configuration generated by an episode falls into a certain extremal configuration space. 
We use an algebraic invariant to distinguish episodes of different extremal configuration spaces in an early tree traversal phase. We find that episodes with the same invariants are likely to reach the same extremal configuration space or spaces with similar extremal value.
Thus, we can treat tree search paths with the same invariant as a single abstraction. By sharing statistics among episodes that belong to one abstraction, 
AATS can learn to identify and prune low-potential tree search paths that lead to configuration spaces with small extremal value.

We adopt the $p$-adic invariants, which is introduced in the Conway--Sloane classification framework for rational quadratic forms. Since Gram entries are rational in the problem instance
considered here, we can compute the invariants
\begin{equation}
z=\operatorname{\textsc{ComputeAlgebraicLabel}}\bigl(\mathbf G(v_n)\bigr),
\end{equation}
once the search reaches a node $v_n$ whose Gram matrix contains its full-rank core defining a non-singular quadratic form. 
Theoretically, the $p$-adic invariants separates different extremal configuration spaces.
AATS leverages this property to label different root-to-node paths as one abstraction and share statistics among these samples.

Specifically, the shared statistics for label $z$ is
$\mathcal S[z]=(N_z,\mu_z,\sigma_z^2)$, where $N_z$ is the number of backup
observations with label $z$, $\mu_z$ is their mean value, and $\sigma_z^2$ is
the empirical variance.  Each completed episode updates this statistics, and
each later episode with the same label queries it.  During traversal, when a path
reaches its first full-rank node, AATS computes $z=\operatorname{\textsc{ComputeAlgebraicLabel}}\bigl(\mathbf G(v_n)\bigr).
$  If the corresponding statistics has already been sampled sufficiently, with low mean and low variance,
\begin{equation}
N_z\ge N_{\rm prune},\qquad
\mu_z<\mu_{\rm prune},\qquad
\sigma_z^2\le\sigma_{\rm prune}^2
\end{equation}
then the current episode is treated as another representative of
an already-resolved configuration space with small extremal value.  
AATS immediately backpropagates $\mu_z$ along the recorded path without launching a new rollout.  
If any condition fails, the path remains active, with the traversal, expansion, rollout and subsequent remove-and-repack transitions continuing as usual. 
This early-backpropagation rule prevents repeated computation on episodes that belong to the stable low-return tree abstraction while keeping uncertain or high-return paths open. 
Thereby, configurations with diverse algebraic invariants are explored, and
the search is more concentrated on paths likely to reach better extremal configuration spaces.

The implementation of the $\textsc{ComputeAlgebraicLabel}$ function is described in Algorithm~\ref{alg:algebraic-class-label}.  
We first select a full-rank core $A$. 
The exact symmetric elimination then produces pivots $(d_1,\ldots,d_n)$ of a
$LDL^\top$-type congruence decomposition.  
For a nonzero rational pivot
$d_j=a_j/b_j$ in the lowest terms, define $e^{(j)}$ as the signed prime-factor
dictionary with $e^{(j)}_{-1}=1$ if $a_j<0$ and
$e^{(j)}_q=\nu_q(|a_j|b_j)$ for each prime $q\mid |a_j|b_j$; missing entries
are interpreted as zero.  Factoring numerator and denominator together is
equivalent for the retained square-class and odd prime mod-eight data, since
odd primes are self-inverse modulo $8$ up to parity.

For each odd prime $p\neq2$, define the helper
\begin{equation}
\textsc{PadicAntiSquare}(e,p)=
\begin{cases}
1, & e_p\ \text{is odd and }
\left(\frac{\prod_{q\in\operatorname{supp}(e)\setminus\{p\},\ e_q\ {\rm odd}}q}{p}\right)
=-1,\\
0, & \text{otherwise},
\end{cases}
\end{equation}
where $\left(\frac{\cdot}{p}\right)$ is the Legendre symbol, and $\operatorname{supp}(e)$ may contain the sign symbol $-1$, contributing the determinant sign to the Hasse--Witt invariant via $\left(\tfrac{-1}{p}\right)$.  The odd-prime
signature is
\begin{equation}
\eta_p=\sum_{j=1}^{n}
\left(p^{e^{(j)}_p}+4\,\textsc{PadicAntiSquare}(e^{(j)},p)-1\right)
\quad(\bmod\,8).
\end{equation}
We store only nonzero pairs $(p,\eta_p)$ in the sorted list $\Psi$.  In
addition, let $\delta=\sum_j e^{(j)}$ and let
\begin{equation}
\Delta_{\rm odd}=
\{q\in\operatorname{supp}(\delta):\delta_q\equiv1\pmod 2\}.
\end{equation}
This second component records the signed determinant square class.  The
algebraic class label is the ordered pair $z=(\Psi,\Delta_{\rm odd})$; both
components are part of the algebraic label.

\begin{algorithm}[ht]
\caption{\textsc{ComputeAlgebraicLabel}: $p$-adic algebraic label}
\label{alg:algebraic-class-label}
\KwIn{feasible Gram matrix $\mathbf G$ containing a rank-$n$ core; dimension $n$}
\KwOut{algebraic label $z=\Phi(\mathbf G)$}
\Fn{\textsc{ComputeAlgebraicLabel}$(\mathbf G,n)$}{
    $m\leftarrow \mathrm{rows}(\mathbf G)$\;
    $\operatorname{Core}(\mathbf G)\leftarrow
    \{J\subseteq[m]: |J|=n,\operatorname{rank}(\mathbf G_{J,J})=n\}$\;
    Select any $J^\star\in\operatorname{Core}(\mathbf G)$\;
    $\mathbf A\leftarrow\mathbf G_{J^\star,J^\star}$\;
    $(d_1,\ldots,d_n)\leftarrow\textsc{CongruencePivots}(\mathbf A)$\;
    $\mathcal D\leftarrow[\,]$, $\delta\leftarrow\emptyset$\;
    \For{$j=1,\ldots,n$}{
        $e^{(j)}\leftarrow\textsc{FactorSignature}(d_j)$\;
        Append $e^{(j)}$ to $\mathcal D$\;
        $\delta\leftarrow\delta+e^{(j)}$\;
    }
    $\mathcal P\leftarrow
    \operatorname{sort}\bigl((\bigcup_{e\in\mathcal D}\operatorname{supp}(e))\setminus\{-1,2\}\bigr)$\;
    $\Psi\leftarrow[\,]$\;
    \ForEach{$p\in\mathcal P$}{
        $\eta_p\leftarrow
        \sum_{e\in\mathcal D}
        \bigl(p^{e_p}+4\,\textsc{PadicAntiSquare}(e,p)-1\bigr)\quad(\bmod\,8)$\;
        \If{$\eta_p\neq0$}{
            Append $(p,\eta_p)$ to $\Psi$\;
        }
    }
    $\Delta_{\rm odd}\leftarrow
    \operatorname{sort}\{q\in\operatorname{supp}(\delta):\delta_q\equiv1\pmod 2\}$\;
    $z\leftarrow(\Psi,\Delta_{\rm odd})$\;
    \Return $z$\;
}
\end{algorithm}

The label produced by Algorithm~\ref{alg:algebraic-class-label} is consumed by the traversal
routine in Algorithm~\ref{alg:filler-traverse}.  At each step, \textsc{Filler.Traverse}
descends from the root using \textsc{SelectChild}, and once the path first reaches a full-rank
core it computes the algebraic class label and consults the shared statistics
$\mathcal S[z]$.  When the pruning condition holds, the traversal aborts and the cached mean is
backpropagated; otherwise it continues until expansion or saturation.

\begin{algorithm}[ht]
\caption{\textsc{Filler.Traverse}: traversal with algebraic abstraction}
\label{alg:filler-traverse}
\KwIn{search tree $\mathcal T$ rooted at $v_{m_0}$; algebraic-class-statistics table $\mathcal S[z]=(N_z,\mu_z,\sigma_z^2)$; problem instance $\mathcal I$; AATS config $\xi_{\rm AATS}$}
\KwOut{gram matrix at leaf node $\mathbf G_{\rm leaf}$; action path $\mathcal P=\operatorname{path}(v)$ from root to leaf; status $\mathbf{h}\in\{\textsc{ReachLeafNode},\textsc{EarlyBackprop}\}$}
\Fn{\textsc{Filler.Traverse}$(\mathcal T,\mathcal S;\mathcal I,\xi_{\rm AATS})$}{
    $\mathcal I=(n,\mathbf G^{(m_0)},\mathcal C_1,\mathcal C_2,\mathcal C_\ast)$, $\xi_{\rm AATS}=(\tau,N_{\rm prune},\mu_{\rm prune},\sigma_{\rm prune}^2,r)$\;

    $v\leftarrow v_{m_0}$, $\mathbf G\leftarrow \mathbf G^{(m_0)}$, $\mathcal T_{\rm cur}\leftarrow \mathcal T[v_{m_0}]$, $\mathcal P \leftarrow \emptyset$\;

    \While{$\mathcal E(v)\neq\emptyset$}{
        \If{$m(v)=n$}{
            $z\leftarrow \textsc{ComputeAlgebraicLabel}(\mathbf G,n)$\;
            \If{$z\notin \operatorname{dom}(\mathcal S)$}{
                $\mathcal S[z]\leftarrow (0,0,0)$\;
            }
            $(N_z,\mu_z,\sigma_z^2)\leftarrow \mathcal S[z]$\;
            \If{$N_z\ge N_{\rm prune}$ \emph{and} $\mu_z<\mu_{\rm prune}$ \emph{and} $\sigma_z^2\le\sigma_{\rm prune}^2$}{
                $\textsc{Filler.Backpropagate}(\mathcal T,\mathcal P,\mu_z,\mathcal S;\mathcal I)$\;
                \Return $\emptyset,\emptyset,\textsc{EarlyBackprop}$\;
            }
        }
        $\mathbf g^\star,v,\mathcal T_{\rm cur}, \mathbf G, o \leftarrow
        \textsc{Filler.SelectChild}(\mathcal T_{\rm cur}, \mathbf G, v;\mathcal I,\xi_{\rm AATS})$\;
        Append $\mathbf g^\star$ to $\mathcal P$\;
        \If {$o=\textsc{Expansion}$}{
            \Return $\mathbf G,\mathcal P,\textsc{ReachLeafNode}$\;
        }
    }

    $\mathcal A\leftarrow \textsc{CandidateActions}(\mathbf G;\mathcal I)$\;
    \If{$\mathcal A\neq\emptyset$}{
        Sample $\mathbf g^\star\sim{\rm Uniform}(\mathcal A)$\;
        $v_{\rm new}, \mathcal T_{\rm cur}, \mathbf G \leftarrow \textsc{Filler.Expansion}(\mathcal T_{\rm cur}, \mathbf G, v, \mathbf g^\star)$\;
        Append $\mathbf g^\star$ to $\mathcal P$\;
        }
    \Return $\mathbf G,\mathcal P,\textsc{ReachLeafNode}$\;
}
\end{algorithm}

After a trajectory finishes, the resulting return is propagated back along the recorded path by
\textsc{Filler.Backpropagate} in Algorithm~\ref{alg:filler-backpropagate}.  Each visited node has
its visit count and accumulated value updated, and whenever the running matrix attains a full-rank
core, the algebraic-class statistics $\mathcal S[z]$ are refreshed online using a Welford-style
running mean and variance.  This shared bookkeeping is what enables later traversals to prune
low-potential algebraic classes.

\begin{algorithm}[ht]
\caption{\textsc{Filler.Backpropagate}: node and algebraic-class updates}
\label{alg:filler-backpropagate}
\KwIn{search tree $\mathcal T$; ordered action path $\mathcal P=(\mathbf g_{m_0},\ldots,\mathbf g_{L-1})$ from root to leaf; value $r_{\rm bp}$; algebraic-class-statistics table $\mathcal S[z]=(N_z,\mu_z,\sigma_z^2)$; problem instance $\mathcal I$}
\KwOut{updated tree statistics $\mathcal T$ and algebraic-class statistics $\mathcal S$}
\Fn{\textsc{Filler.Backpropagate}$(\mathcal T,\mathcal P,r_{\rm bp},\mathcal S;\mathcal I)$}{
    $\mathcal I=(n,\mathbf G^{(m_0)},\mathcal C_1,\mathcal C_2,\mathcal C_\ast)$\;
    $\mathcal T_{\rm cur}\leftarrow\mathcal T[v_{m_0}]$, $\mathbf G\leftarrow\mathbf G^{(m_0)}$\;
    $N,Q\leftarrow\mathcal T_{\rm cur}.\texttt{state}$\;
    $\mathcal T_{\rm cur}.\texttt{state}\leftarrow [N+1,Q+r_{\rm bp}]$\;
    \ForEach{$\mathbf g\in\mathcal P$}{
        $\mathcal T_{\rm cur}\leftarrow\mathcal T_{\rm cur}.\texttt{children}[\mathbf g]$\;
        $\mathbf G\leftarrow\operatorname{Extend}(\mathbf G,\mathbf g)$\;
        $N,Q\leftarrow\mathcal T_{\rm cur}.\texttt{state}$\;
        $\mathcal T_{\rm cur}.\texttt{state}\leftarrow [N+1,Q+r_{\rm bp}]$\;
        \If{$\mathrm{rows}(\mathbf G)=n$}{
            $z\leftarrow \textsc{ComputeAlgebraicLabel}(\mathbf G,n)$\;
            \If{$z\notin \operatorname{dom}(\mathcal S)$}{
                $\mathcal S[z]\leftarrow (0,0,0)$\;
            }
            $(N_z,\mu_z,\sigma_z^2)\leftarrow \mathcal S[z]$\;
            $N_z'\leftarrow N_z+1$\;
            $\Delta\leftarrow r_{\rm bp}-\mu_z$\;
            $\mu_z'\leftarrow \mu_z+\Delta/N_z'$\;
            \eIf{$N_z=0$}{
                $\sigma_z^{2\prime}\leftarrow 0$\;
            }{
                $\sigma_z^{2\prime}\leftarrow
                \dfrac{(N_z-1)\sigma_z^2+\Delta(r_{\rm bp}-\mu_z')}
                {N_z'-1}$\;
            }
            $\mathcal S[z]\leftarrow (N_z',\mu_z',\sigma_z^{2\prime})$\;
        }
    }
    \Return $\mathcal T,\mathcal S$\;
}
\end{algorithm}

\subsection{Corrector: Permutation-Equivariant Architecture}
\label{sec:corrector-arch}

At each step, the Corrector observes a saturated Gram matrix
\(\mathbf G^{(m)}\) produced by the Filler and selects an action
\(I \sim \pi_\theta(\cdot \mid \mathbf G^{(m)})\), where \(I\) specifies the set
of rows and columns to retain.
This action induces a deterministic state transition by removing the
complementary rows and columns, yielding the reduced Gram matrix
\(\mathbf G^{(m)}_{I,I}\), which is returned to the Filler for subsequent
repacking.
The policy \(\pi_\theta\) is parameterized by a permutation-equivariant
architecture, reflecting the fact that any simultaneous row--column
reindexing \(\mathbf G^{(m)} \mapsto \mathbf P \mathbf G^{(m)} \mathbf P^\top\)
corresponds only to a relabeling of sphere centers and leaves the underlying
geometric configuration unchanged.
The following subsection describes the architecture in detail.

\subsubsection{Core-column feature map}
\label{sec:core-column-feature}

Let \(\mathbf G^{(m)} \in \mathbb R^{m \times m}\) denote the current partial Gram
matrix. Following the matrix-completion convention, the first \(n\) rows and columns of
\(\mathbf G^{(m)}\) are assumed to contain a full-rank core block.
When different initial configurations are used, this core block may appear at
different positions within the matrix. In such cases, we select an
\(n \times n\) full-rank block and apply a simultaneous row--column permutation
to relocate it to rows and columns \(1{:}n\), which does not alter the underlying
geometric configuration.

Since \(\mathbf G^{(m)}\) is a rank-\(n\) positive semidefinite Gram matrix, the
core block fully determines the entire matrix via
\begin{equation}
    \mathbf G^{(m)}
    =
    \mathbf G^{(m)}_{:,1:n}
    \bigl(\mathbf G^{(m)}_{1:n,1:n}\bigr)^{-1}
    \mathbf G^{(m)}_{1:n,:}.
\end{equation}
Accordingly, we use the core-column feature matrix
\begin{equation}
    \mathbf S
    =
    \Psi_{\mathrm{core}}(\mathbf G^{(m)})
    =
    \mathbf G^{(m)}_{:,1:n}
    \in \mathbb R^{m \times n}
\end{equation}
as the network input. Each of the \(m\) rows of \(\mathbf S\) represents one
sphere, and the policy operates over this set of sphere-level features.

\subsubsection{Permutation equivariance and invariant value}
\label{sec:permutation-invariance}

Let \(\mathbf P \in \{0,1\}^{m \times m}\) denote a permutation matrix acting on the
current Gram matrix. A simultaneous reindexing
\(\mathbf G^{(m)} \mapsto \mathbf P \mathbf G^{(m)} \mathbf P^\top\) represents the
same geometric configuration, differing only in the ordering of the
sphere-center vectors. Under the core-column convention introduced above, this
operation induces a row-wise reindexing of the feature matrix:
\begin{equation}
    \Psi_{\mathrm{core}}(\mathbf P \mathbf G^{(m)} \mathbf P^\top)
    =
    \mathbf P \, \Psi_{\mathrm{core}}(\mathbf G^{(m)}).
\end{equation}
Equivalently, the state representation transforms as
\(\mathbf S \mapsto \mathbf P \mathbf S\).

Respecting this symmetry, a scalar-valued function \(g\) defined on the state
space is required to be permutation invariant,
\begin{equation}
    g(\mathbf P \mathbf S) = g(\mathbf S),
\end{equation}
whereas a row-indexed function \(h\) must be permutation equivariant,
\begin{equation}
    h(\mathbf P \mathbf S) = \mathbf P \, h(\mathbf S).
\end{equation}

The proposed architecture is explicitly designed to satisfy these symmetry
constraints. In particular, the encoder applies a shared multilayer perceptron
to each row of \(\mathbf S\), and each attention block models pairwise
interactions among rows in a permutation-equivariant manner. The symmetry
properties of all architectural components are summarized in
Table~\ref{tab:symmetry}.

\subsubsection{Architecture overview}
\label{sec:arch-overview}

With \(\mathbf S = \Psi_{\mathrm{core}}(\mathbf G^{(m)})\), the Corrector network
is defined as
\begin{equation}
    f_\theta(\mathbf S)
    =
    \bigl(
        \boldsymbol\ell_\theta(\mathbf S),
        V_\theta(\mathbf S),
        \mathbf H^{\mathrm{pre}}_\theta(\mathbf S)
    \bigr).
\end{equation}
Here \(\boldsymbol\ell_\theta\) produces row-wise retention logits,
\(V_\theta\) estimates the state value, and
\(\mathbf H^{\mathrm{pre}}_\theta\) denotes the pre-attention encoder embeddings,
which map the input features into a latent space supervised by the spectral
contrastive objective described in
Section~\ref{sec:supcon}.

Algorithm~\ref{alg:corrector-forward} summarizes the complete forward pass. The
trainable parameters are partitioned into encoder, attention, policy, and value
components,
\begin{equation}
    \theta =
    (\theta_{\mathrm{enc}}, \theta_{\mathrm{att}},
     \theta_{\mathrm{pol}}, \theta_{\mathrm{val}}).
\end{equation}
During training, the PPO objective backpropagates through the encoder,
self-attention stack, policy head, and value head. The spectral contrastive loss
is applied to \(\mathbf H^{\mathrm{pre}}_\theta\), encouraging the encoder to
learn latent representations with geometrically and spectrally meaningful
structure.

\begin{algorithm}[ht]
\caption{\textsc{Corrector.Forward}: permutation-equivariant policy-value network}
\label{alg:corrector-forward}
\KwIn{Gram matrix $\mathbf G^{(m)}\in\mathbb R^{m\times m}$ with a full-rank core in rows and columns $1:n$; parameters $\theta$; problem instance $\mathcal I$; Corrector config $\xi_{\rm net}$}
\KwOut{row retention logits $\boldsymbol\ell\in\mathbb R^m$; value $v\in\mathbb R$; pre-attention embeddings $\mathbf H^{\rm pre}\in\mathbb R^{m\times d}$}
\Fn{\textsc{Corrector.Forward}$(\mathbf G^{(m)};\theta,\mathcal I,\xi_{\rm net})$}{
    $\mathcal I=(n,\ldots)$, $\xi_{\rm net}=(L_{\rm att}, L_{\rm enc},d,h,r_{\rm ffn},\mathcal J_0)$\;
    $\mathbf S\leftarrow\mathbf G^{(m)}_{:,1:n}$\;
    $\mathbf H^{\rm pre}\leftarrow
    \textsc{Corrector.Encoder}(\mathbf S;\theta_{\rm enc},\xi_{\rm net})$\;
    $\mathbf H\leftarrow\mathbf H^{\rm pre}$\;
    \For{$\lambda=1,\ldots,L_{\rm att}$}{
        $\mathbf H\leftarrow
        \textsc{Corrector.AttentionBlock}_\lambda(\mathbf H;\theta_{\rm att}^{(\lambda)},\xi_{\rm net})$\;
    }
    $\mathbf H^{\rm post}\leftarrow\mathbf H$\;
    $(\boldsymbol\ell,v)\leftarrow
    \textsc{Corrector.Heads}(\mathbf H^{\rm post};
    \theta_{\rm pol},\theta_{\rm val})$\;
    \Return $\boldsymbol\ell,v,\mathbf H^{\rm pre}$\;
}
\end{algorithm}

\subsubsection{Encoder}
\label{sec:encoder}

The encoder is implemented as a shared row-wise multi-layer perceptron.
The same function is applied independently to each feature row, so any reordering of the rows
induces the same reordering of the encoder outputs.
Permutation equivariance is guaranteed by this design. 
The depth and width of the encoder are treated as hyperparameters, selected
according to the specific game instance and the dimensionality of the
core-column input
\(\Psi_{\mathrm{core}}(\mathbf G)=\mathbf G_{:,1:n}\).
GELU activations are applied after each hidden affine layer, while the final
affine layer is left unactivated to produce the pre-attention embeddings.

\begin{algorithm}[ht]
\caption{\textsc{Corrector.Encoder}: shared row-wise MLP}
\label{alg:corrector-encoder}
\KwIn{core-column features $\mathbf S\in\mathbb R^{m\times n}$; encoder parameters $\theta_{\rm enc}$; Corrector config $\xi_{\rm net}$}
\KwOut{pre-attention embeddings $\mathbf H^{\rm pre}\in\mathbb R^{m\times d}$}
\Fn{\textsc{Corrector.Encoder}$(\mathbf S;\theta_{\rm enc},\xi_{\rm net})$}{
    $\xi_{\rm net}=(L_{\rm att}, L_{\rm enc},d,h,r_{\rm ffn},\mathcal J_0)$\;
    $\{(\mathbf W_q,\mathbf b_q)\}_{q=1}^{L_{\rm enc}}\leftarrow\theta_{\rm enc}$\;
    $\mathbf H\leftarrow\mathbf S$\;
    \For{$q=1,\ldots,L_{\rm enc}$}{
        $\mathbf H\leftarrow\mathbf H\mathbf W_q+\mathbf b_q$
        \If{$q<L_{\rm enc}$}{
            $\mathbf H\leftarrow\operatorname{GELU}(\mathbf H)$\;
        }
    }
    \Return $\mathbf H$\;
}
\end{algorithm}

\subsubsection{Self-attention stack}
\label{sec:attention}
The \(L_{\mathrm{att}}\) multi-head self-attention blocks operate on the latent
representations produced by the encoder. Each block adopts a pre-normalization
design, applying layer normalization prior to both the attention and
feed-forward sublayers:
\begin{equation}
    \widetilde{\mathbf X}
    =
    \operatorname{LayerNorm}_{\mathrm{att}}(\mathbf X),
    \qquad
    \mathbf X'
    =
    \mathbf X + \operatorname{Attention}(\widetilde{\mathbf X}),
    \qquad
    \mathbf X''
    =
    \mathbf X'
    +
    \operatorname{FFN}\!\left(
        \operatorname{LayerNorm}_{\mathrm{ffn}}(\mathbf X')
    \right).
\end{equation}
The output of the final attention block, denoted by \(\mathbf H^{\mathrm{post}}\),
is fed directly into the policy and value heads.

\begin{algorithm}[ht]
\caption{\textsc{Corrector.AttentionBlock}: pre-norm self-attention}
\label{alg:mabv2}
\KwIn{row embeddings $\mathbf H\in\mathbb R^{m\times d}$; block parameters $\theta_{\rm att}^{(\lambda)}$; Corrector config $\xi_{\rm net}$}
\KwOut{updated tensor $\mathbf O\in\mathbb R^{m\times d}$}
\Fn{\textsc{Corrector.AttentionBlock}$(\mathbf H;\theta_{\rm att}^{(\lambda)},\xi_{\rm net})$}{
    $\xi_{\rm net}=(L_{\rm att}, L_{\rm enc},d,h,r_{\rm ffn},\mathcal J_0)$, $d_h\leftarrow d/h$\;
    $\bigl(\{\mathbf W_Q^{(r)},\mathbf W_K^{(r)},\mathbf W_V^{(r)}\}_{r=1}^{h},
    \mathbf W_O,\theta_{\rm ffn}\bigr)\leftarrow\theta_{\rm att}^{(\lambda)}$\;
    $\widetilde{\mathbf H}\leftarrow
    \operatorname{LayerNorm}_{\rm att}(\mathbf H)$\;
    \For{$r=1,\ldots,h$}{
        $\mathbf Q^{(r)}\leftarrow\widetilde{\mathbf H}\mathbf W_Q^{(r)}$,\quad
        $\mathbf K^{(r)}\leftarrow\widetilde{\mathbf H}\mathbf W_K^{(r)}$,\quad
        $\mathbf V^{(r)}\leftarrow\widetilde{\mathbf H}\mathbf W_V^{(r)}$\;
        \ForEach{$i\in[m]$}{
            \ForEach{$j\in[m]$}{
                $s^{(r)}_{i,j}\leftarrow
                \langle \mathbf Q^{(r)}_{i,:},\mathbf K^{(r)}_{j,:}\rangle/\sqrt{d_h}$\;
            }
            $\mathbf A^{(r)}_{i,:}\leftarrow
            \operatorname{softmax}_{j\in[m]}(s^{(r)}_{i,j})$\;
            $\mathbf Z^{(r)}_{i,:}\leftarrow
            \sum_{j=1}^{m}A^{(r)}_{i,j}\mathbf V^{(r)}_{j,:}$\;
        }
    }
    $\mathbf Z\leftarrow
    \operatorname{Concat}(\mathbf Z^{(1)},\mathbf Z^{(2)},\ldots,\mathbf Z^{(h)})\mathbf W_O$\;
    $\mathbf R\leftarrow\mathbf H+\mathbf Z$\;
    $\widetilde{\mathbf R}\leftarrow
    \operatorname{LayerNorm}_{\rm ffn}(\mathbf R)$\;
    $\mathbf O\leftarrow
    \mathbf R+\operatorname{FFN}(\widetilde{\mathbf R};\theta_{\rm ffn}, r_{\rm ffn})$\;
    \Return $\mathbf O$\;
}
\end{algorithm}

\subsubsection{Policy head and value head}
\label{sec:heads}

Algorithm~\ref{alg:corrector-heads} specifies the two output heads. The value head is permutation-invariant and estimates the expected future remove-and-repack return from the current state. It first aggregates the post-attention row embeddings via mean pooling,
\begin{equation}
    \bar{\mathbf h}=\dfrac{1}{m}\sum_i\mathbf H^{\rm post}_{i,:},
\end{equation}
and then maps the resulting global configuration descriptor through an MLP to produce the scalar value estimate $v_\theta(\mathbf S)$.

The policy head is permutation-equivariant, as it applies a shared MLP independently to each row of $\mathbf H^{\rm post}$. The protected rows $\mathcal J_0$---typically the first $n$ rows constituting the full-rank core block---are retained deterministically and excluded from the stochastic policy. For every row, the policy head emits a real-valued logit $\ell_i$, which the sigmoid function converts into a Bernoulli retention probability,
\begin{equation}
    p_i=\sigma(\ell_i)=\frac{1}{1+\exp(-\ell_i)}.
\end{equation}
Protected rows are fixed to $a_i=1$, while actions on the remaining unprotected rows are sampled independently:
\begin{equation}
    a_i\sim{\rm Bernoulli}(p_i),
    \qquad
    p_i=\sigma(\ell_i).
\end{equation}
Consequently, the policy distribution factorizes over the unprotected rows:
\begin{equation}
    \pi_\theta(\mathbf a\mid \mathbf S)
    =
    \prod_{i\in[m]\setminus\mathcal J_0}
    {\rm Bernoulli}\bigl(a_i;\,p_i\bigr)
    =
    \prod_{i\in[m]\setminus\mathcal J_0}
    p_i^{a_i}(1-p_i)^{1-a_i},
\end{equation}
and the policy entropy decomposes as a sum of Bernoulli entropies over the same set of rows:
\begin{equation}
    H\!\left(\pi_\theta(\cdot\mid\mathbf S)\right)
    =
    \sum_{i\in[m]\setminus\mathcal J_0}
    H\!\left({\rm Bernoulli}(p_i)\right)
    =
    \sum_{i\in[m]\setminus\mathcal J_0}
    \bigl[-p_i\log p_i-(1-p_i)\log(1-p_i)\bigr].
\end{equation}

\begin{algorithm}[ht]
\caption{\textsc{Corrector.Heads}: row-equivariant policy and invariant value}
\label{alg:corrector-heads}
\KwIn{post-attention embeddings $\mathbf H^{\rm post}\in\mathbb R^{m\times d}$; policy-head parameters $\theta_{\rm pol}$; value-head parameters $\theta_{\rm val}$}
\KwOut{row retention logits $\boldsymbol\ell$; value estimate $v$}
\Fn{\textsc{Corrector.Heads}$(\mathbf H^{\rm post};\theta_{\rm pol},\theta_{\rm val})$}{
    $\bar{\mathbf h}\leftarrow
    \dfrac{1}{m}\sum_{i=1}^{m}\mathbf H^{\rm post}_{i,:}$\;
    $v\leftarrow\operatorname{MLP}_{\rm val}(\bar{\mathbf h};\theta_{\rm val})$\;
    \ForEach{$i\in[m]$}{
        $\ell_i\leftarrow\operatorname{MLP}_{\rm pol}(\mathbf H^{\rm post}_{i,:};\theta_{\rm pol})$\;
    }
    \Return $\boldsymbol\ell,v$\;
}
\end{algorithm}

Algorithm~\ref{alg:corrector-remove} realizes the sampling step that converts the policy head's output into the concrete remove-and-keep decision consumed by the cooperative game. The resulting index set $I$ of retained rows defines the post-removal submatrix $\mathbf G_{I,I}$, which is passed back to the Filler.

\begin{algorithm}[ht]
\caption{\textsc{Corrector.Remove}: sampling a retained index set}
\label{alg:corrector-remove}
\KwIn{current Gram matrix $\mathbf G\in\mathbb R^{m\times m}$ with a full-rank core; Corrector policy $\pi_\theta$; problem instance $\mathcal I$; Corrector config $\xi_{\rm net}$}
\KwOut{post-removal Gram matrix $\mathbf G'$;retained index list $I$; log-probability $\log\pi_\theta(I\mid\mathbf S)$; value estimate $V_\theta(\mathbf S)$}
\Fn{\textsc{Corrector.Remove}$(\mathbf G;\pi_\theta,\mathcal I,\xi_{\rm net})$}{
    $\xi_{\rm net}=(L_{\rm att}, L_{\rm enc},d,h,r_{\rm ffn},\mathcal J_0)$\;
    $(\boldsymbol\ell,V,\mathbf H^{\rm pre})
    \leftarrow\textsc{Corrector.Forward}(\mathbf G;\theta,\mathcal I,\xi_{\rm net})$\;
    $I \leftarrow$ ordered list of protected indices in $\mathcal J_0$\;
    \ForEach{$i\in [m]\setminus\mathcal J_0$}{
        Sample $a_i\sim{\rm Bernoulli}(\sigma(\ell_i))$\;
        \If{$a_i=1$}{
            Append $i$ to $I$\;
        }
    }
    $\log\pi\leftarrow
    \sum_{i\in [m]\setminus\mathcal J_0}
    \log{\rm Bernoulli}(a_i;\sigma(\ell_i))$\;
    $\mathbf G'\leftarrow\mathbf G_{I,I}$\;
    \Return $\mathbf G',I ,\log\pi,V$\;
}
\end{algorithm}

\subsubsection{Architecture summary}
\label{sec:arch-summary}

Table~\ref{tab:symmetry} summarizes the symmetry properties of each component. 
Table~\ref{tab:param-counts} reports the corresponding trainable parameter counts.  
The counts include affine biases and LayerNorm scale and
shift parameters. While the size of current gram matrix $m$ governs the activation sizes and the $O(L_{\rm att}m^2d)$ cost of attention, it does not affect the number of parameters.
For the implementation used in the reported runs
For the configuration used in the reported experiments ($d_{\rm in}=24$, $L_{\rm enc}=3$, $L_{\rm att}=12$, $d=512$, $h=8$, and $r_{\rm ffn}=4$, with policy and value MLPs of widths $512\to256\to64\to1$), the Corrector comprises $35.5$M trainable parameters.

\begin{table}[ht]
    \centering
    \caption{\textbf{Corrector component shapes and permutation symmetries.}
    Equivariance and invariance are separated according to whether the
    component output is row-indexed or scalar.}
    \label{tab:symmetry}
    \setlength{\tabcolsep}{3pt}
    \renewcommand{\arraystretch}{1.15}
    \footnotesize
    \begin{tabular}{@{}p{0.27\linewidth}p{0.24\linewidth}p{0.45\linewidth}@{}}
        \toprule
        \textbf{Component and map} & \textbf{Permutation equivariance} & \textbf{Permutation invariance} \\
        \midrule
        \begin{tabular}[t]{@{}l@{}}
        \textbf{Encoder}\\[-1pt]
        $E_{\theta_{\rm enc}}:\mathbb R^{m\times d_{\rm in}}\to\mathbb R^{m\times d}$
        \end{tabular}
        & $E_{\theta_{\rm enc}}(\mathbf P\mathbf S)=\mathbf P E_{\theta_{\rm enc}}(\mathbf S)$
        & --- \\
        \begin{tabular}[t]{@{}l@{}}
        \textbf{Self-attention stack ($\times L_{\rm att}$)}\\[-1pt]
        $A_{\theta_{\rm att}}:\mathbb R^{m\times d}\to\mathbb R^{m\times d}$
        \end{tabular}
        & $A_{\theta_{\rm att}}(\mathbf P\mathbf H)=\mathbf P A_{\theta_{\rm att}}(\mathbf H)$
        & --- \\
        \begin{tabular}[t]{@{}l@{}}
        \textbf{Policy head}\\[-1pt]
        $\boldsymbol\ell_{\theta_{\rm pol}}:\mathbb R^{m\times d}\to\mathbb R^m$
        \end{tabular}
        & $\boldsymbol\ell_{\theta_{\rm pol}}(\mathbf P\mathbf H)=\mathbf P\boldsymbol\ell_{\theta_{\rm pol}}(\mathbf H)$
        & --- \\
        \begin{tabular}[t]{@{}l@{}}
        \textbf{Value head}\\[-1pt]
        $V_{\theta_{\rm val}}:\mathbb R^{m\times d}\to\mathbb R$
        \end{tabular}
        & ---
        & $V_{\theta_{\rm val}}(\mathbf P\mathbf H)=V_{\theta_{\rm val}}(\mathbf H)$ \\
        \begin{tabular}[t]{@{}l@{}}
        \textbf{SupCon loss}\\[-1pt]
        $\textsc{SupConLoss}:$\\[-1pt]
        $\mathbb R^{m\times d}\times\{0,1,\ldots\}^{m}\to\mathbb R$
        \end{tabular}
        & --- &
        $\begin{aligned}
        &\textsc{SupConLoss}(\mathbf P\mathbf H^{\rm pre},\mathbf P\mathbf y;\tau_{\rm supcon})\\
        &=\textsc{SupConLoss}(\mathbf H^{\rm pre},\mathbf y;\tau_{\rm supcon})
        \end{aligned}$ \\
        \bottomrule
    \end{tabular}
\end{table}

\begin{table}[ht]
    \centering
    \caption{\textbf{Corrector parameter counts.}
    Let $d_{\rm in}=n$, let the encoder use $L_{\rm enc}$ affine layers of
    hidden width $d$, and let each attention block use FFN expansion
    $r_{\rm ffn}$.  The reported implementation uses the multi-head attention block with
    $Q,K,V$ projections, a two-layer FFN and two LayerNorms.  For the two head MLPs, write their widths as
    $u^{\rm pol}_0=d,\ldots,u^{\rm pol}_{Q_{\rm pol}}=1$ and
    $u^{\rm val}_0=d,\ldots,u^{\rm val}_{Q_{\rm val}}=1$.}
    \label{tab:param-counts}
    \setlength{\tabcolsep}{4pt}
    \renewcommand{\arraystretch}{1.15}
    \scriptsize
    \begin{tabular}{@{}p{0.20\linewidth}p{0.37\linewidth}p{0.15\linewidth}p{0.20\linewidth}@{}}
        \toprule
        \textbf{Component} & \textbf{Trainable parameters} & \textbf{Reported} & \textbf{Included terms} \\
        \midrule
        Encoder &
        $d_{\rm in}d+d+(L_{\rm enc}-1)(d^2+d)$ &
        $0.54$M &
        Shared row-wise affine layers \\
        One attention block &
        $(3+2r_{\rm ffn})d^2+(r_{\rm ffn}+8)d$ &
        $2.89$M &
        $Q,K,V$ projections, two-layer FFN, two LayerNorms \\
        Attention stack &
        $L_{\rm att}\bigl((3+2r_{\rm ffn})d^2+(r_{\rm ffn}+8)d\bigr)$ &
        $34.68$M &
        $L_{\rm att}$ identical-shape blocks with independent weights \\
        Final LayerNorm &
        $2d$ &
        $<0.01$M &
        Scale and shift after the attention stack \\
        Policy head &
        $\sum_{q=1}^{Q_{\rm pol}}\bigl(u^{\rm pol}_{q-1}u^{\rm pol}_{q}
        +u^{\rm pol}_{q}\bigr)$ &
        $0.15$M &
        Shared row-wise MLP ending in one logit \\
        Value head &
        $\sum_{q=1}^{Q_{\rm val}}\bigl(u^{\rm val}_{q-1}u^{\rm val}_{q}
        +u^{\rm val}_{q}\bigr)$ &
        $0.15$M &
        MLP after mean pooling \\
        Total &
        $P_{\rm enc}+P_{\rm att}+P_{\rm LN}+P_{\rm pol}+P_{\rm val}$ &
        $35.5$M &
        Sum of the component counts above \\
        \bottomrule
    \end{tabular}
\end{table}

\subsection{Training the Corrector}
\label{sec:training}

The Corrector is trained on the remove-and-repack trajectories collected by Algorithm~\ref{alg:cotrain}. Each transition stores the pre-correction matrix, the retained-index set sampled by the policy, the corresponding old log-probability and value estimate, and the reward obtained after Filler repacking. These trajectories drive a standard PPO policy-gradient update. In addition, each Gram matrix naturally exposes structural information that serves as auxiliary supervision: rows sharing the same spectral label are treated as positives in a supervised contrastive loss applied to the pre-attention embeddings.

\subsubsection{Recursive spectrum partition of Gram matrix}
\label{sec:spectral}

We define the \emph{spectrum} of $\mathbf G$ as the multiset of row-wise histograms of off-diagonal entries. Concretely, for a row subset $U=(u_1,\ldots,u_s)$ with induced submatrix $\mathbf G_U=\mathbf G_{U,U}$, the row spectrum of $u_a$ is the count vector
\begin{equation}
\boldsymbol\chi_a
=
\Bigl(\bigl|\{b\in[s]\setminus\{a\}:(\mathbf G_U)_{ab}=c\}\bigr|\Bigr)_{c\in\mathcal C_{\rm spec}},
\end{equation}
where $\mathcal C_{\rm spec}$ denotes the set of distinct off-diagonal values present in $\mathbf G_U$. Rows with identical row spectra are grouped into the same class, and the procedure is then applied recursively within each resulting submatrix until every row in the current submatrix shares the same spectrum. The full procedure is given in Algorithm~\ref{alg:spectral-partition}.

This recursive row-spectrum partition yields a rich geometric descriptor and is used in two places. During Corrector training, it supplies the positive/negative labels for the supervised contrastive loss. After a completed game, it provides candidate fragments for initializing the next game, as described in Section~\ref{sec:matrix-decomposition}.

\begin{algorithm}[ht]
\caption{\textsc{SpectralPartition}: recursive row-spectrum partition}
\label{alg:spectral-partition}
\KwIn{Gram matrix $\mathbf G\in\mathbb R^{m\times m}$}
\KwOut{row labels $\mathbf y\in\{0,1,\ldots\}^{m}$}
\Fn{\textsc{SpectralPartition}$(\mathbf G)$}{
    $\mathcal C_{\rm spec}\leftarrow
    \operatorname{Unique}\{G_{ij}:1\le i<j\le m\}$\;
    $\mathbf y\leftarrow(\bot,\ldots,\bot)$, $q\leftarrow0$\tcp*{$\bot$ denotes a temporary unassigned label}
    \Fn{\textsc{Refine}$(U,\mathbf G_U)$}{
        \If{$|U|\le1$}{
            $y_i\leftarrow q$ for all $i\in U$, $q\leftarrow q+1$\;
            \Return\;
        }
        Write $U=(u_1,\ldots,u_s)$ in the inherited row order\;
        \For{$a=1,\ldots,s$}{
            $\boldsymbol\chi_a\leftarrow
            \bigl(|\{b\in[s]\setminus\{a\}:(\mathbf G_U)_{ab}=c\}|\bigr)_{c\in\mathcal C_{\rm spec}}$\;
            $h_a\leftarrow\operatorname{Hash}(\boldsymbol\chi_a)$\;
        }
        $\{P_1,\ldots,P_K\}\leftarrow$ partition of $[s]$ induced by equal hash values $h_a$\;
        \eIf{$K=1$}{
            $y_i\leftarrow q$ for all $i\in U$, $q\leftarrow q+1$\;
        }{
            \For{$k=1,\ldots,K$}{
                $U_k\leftarrow(u_a:a\in P_k)$\;
                $\mathbf G_{U_k}\leftarrow(\mathbf G_U)_{P_k,P_k}$\;
                \textsc{Refine}$(U_k,\mathbf G_{U_k})$\;
            }
        }
    }
    $\mathbf G_{[m]}\leftarrow\mathbf G$\;
    \textsc{Refine}$([m],\mathbf G_{[m]})$\;
    \Return $\mathbf y$\;
}
\end{algorithm}

\subsubsection{Spectrum-supervised contrastive learning}
\label{sec:supcon}

The supervised contrastive objective operates on the pre-attention embeddings $\mathbf H^{\rm pre}$ returned by \textsc{Corrector.Forward}, regularizing the row-wise representation with explicit geometric features before contextual self-attention mixes information across rows. Algorithm~\ref{alg:supcon-loss} details the computation. Within a single Gram matrix, rows belonging to the same recursive row-spectrum class serve as positives for one another, while all remaining rows act as negatives. The loss is then averaged over the states in each training minibatch.

\begin{algorithm}[ht]
\caption{\textsc{SupConLoss}: spectral supervised contrastive loss}
\label{alg:supcon-loss}
\KwIn{row embeddings $\mathbf H\in\mathbb R^{m\times d}$; spectral labels $\mathbf y\in\{0,1,\ldots\}^{m}$; temperature $\tau_{\rm supcon}$}
\KwOut{contrastive loss $\mathcal L_{\rm supcon}$}
\Fn{\textsc{SupConLoss}$(\mathbf H,\mathbf y;\tau_{\rm supcon})$}{
    $\mathcal L\leftarrow0$, $N_{\rm anc}\leftarrow0$\;
    $\mathbf z_i\leftarrow \mathbf H_i/\|\mathbf H_i\|_2$ for all $i\in[m]$\;
    \For{$i=1,\ldots,m$}{
        $\Omega_i\leftarrow\{\omega \in[m]\setminus\{i\}:y_\omega=y_i\}$\;
        \If{$\Omega_i=\emptyset$}{\textbf{continue}}
        $\ell_i\leftarrow
        -\dfrac{1}{|\Omega_i|}
        \sum_{\omega \in\Omega_i}
        \log
        \dfrac{\exp(\langle\mathbf z_i,\mathbf z_\omega\rangle/\tau_{\rm supcon})}
        {\sum_{j\in[m]\setminus\{i\}}
        \exp(\langle\mathbf z_i,\mathbf z_j\rangle/\tau_{\rm supcon})}$\;
        $\mathcal L\leftarrow\mathcal L+\ell_i$,
        $N_{\rm anc}\leftarrow N_{\rm anc}+1$\;
    }
    \Return $\mathcal L/\max(N_{\rm anc},1)$\;
}
\end{algorithm}

\subsubsection{Reward and Generalized Advantage Estimation}
\label{sec:reward}

The Corrector receives reward from the outcome of a remove-and-repack
transition.  If $\mathbf G_t$ is the matrix before removal and
$\mathbf G_{t+1}$ is the best rollout completion, the raw progress signal is
\begin{equation}
\Delta m_t=\mathrm{rows}(\mathbf G_{t+1})-\mathrm{rows}(\mathbf G_t).
\end{equation}
The reward $r_t$ for this transition is a function of $\Delta m_t$; in practice,
we simply use the configured scalar function
\begin{equation}
r_t=\textsc{Reward}(\Delta m_t),\qquad
\textsc{Reward}(\Delta m)=\beta_{\rm rew}\Delta m,
\end{equation}
where $\beta_{\rm rew}>0$ is an experiment-level scaling constant.  Generalized
Advantage Estimation is computed on the finite-horizon trajectories in
$\mathcal B$, as described in Algorithm~\ref{alg:supcon-loss} .

\begin{algorithm}[ht]
\caption{\textsc{ComputeGAE}: returns and advantages for PPO}
\label{alg:compute-gae}
\KwIn{trajectory batch $\mathcal B$; training config $\xi_{\rm train}$}
\KwOut{transition dataset $\mathcal D$ containing advantages $\widehat A_t$ and returns $\widehat R_t$}
\Fn{\textsc{ComputeGAE}$(\mathcal B;\xi_{\rm train})$}{
    $\xi_{\rm train}=(\ldots,\gamma,\lambda_{\rm GAE},\ldots)$\;
    $\mathcal D\leftarrow\emptyset$\;
    \ForEach{episode trajectory $(\mathbf G_t,I_t,\log\pi_t^{\rm old},V_t^{\rm old},r_t,d_t)_{t=0}^{T-1}\subset\mathcal B$}{
        $\widehat A_T\leftarrow0$\;
        \For{$t=T-1,T-2,\ldots,0$}{
            \eIf{$d_t=1$}{
                $\bar V_{t+1}^{\rm old}\leftarrow0$\;
            }{
                $\bar V_{t+1}^{\rm old}\leftarrow V_{t+1}^{\rm old}$ from the next transition in the same episode\;
            }
            $\delta_t\leftarrow
            r_t+\gamma(1-d_t)\bar V_{t+1}^{\rm old}-V_t^{\rm old}$\;
            $\widehat A_t\leftarrow
            \delta_t+\gamma\lambda_{\rm GAE}(1-d_t)\widehat A_{t+1}$\;
            $\widehat R_t\leftarrow\widehat A_t+V_t^{\rm old}$\;
            Add $(\mathbf G_t,I_t,\log\pi_t^{\rm old},V_t^{\rm old},\widehat A_t,\widehat R_t)$ to $\mathcal D$\;
        }
    }
    Normalize $\widehat A_t$ over $\mathcal D$ to zero mean and unit standard deviation\;
    \Return $\mathcal D$\;
}
\end{algorithm}

\subsubsection{Policy optimization}
\label{sec:ppo}

Algorithm~\ref{alg:corrector-update} specifies
\textsc{Corrector.Update} in Algorithm~\ref{alg:cotrain}.  The update first
converts collected trajectories into advantages and returns, and then performs
minibatch optimization of a combined loss. In the combined loss, the policy term is the PPO clipped
surrogate, the value term uses clipped value regression, the entropy term
encourages exploration over retention actions, and the spectrum-supervised contrastive loss term regularizes the pre-attention representation. The optimizer is initialized with \textsc{BuildOptimizer} and takes a step on the combined loss for each minibatch.

\begin{algorithm}[ht]
\caption{\textsc{Corrector.Update}: PPO with spectral contrastive auxiliary loss}
\label{alg:corrector-update}
\KwIn{Corrector policy $\pi_\theta$; optimizer $\mathcal O$; on-policy trajectory batch $\mathcal B$; problem instance $\mathcal I$; Corrector config $\xi_{\rm net}$; training config $\xi_{\rm train}$}
\KwOut{updated policy $\pi_\theta$ and optimizer $\mathcal O$}
\Fn{\textsc{Corrector.Update}$(\pi_\theta,\mathcal O,\mathcal B;\mathcal I,\xi_{\rm net},\xi_{\rm train})$}{
    $\xi_{\rm train}=(T,K,N_{\rm traj},\gamma,\lambda_{\rm GAE},
    \epsilon_{\rm clip},\epsilon_V,c_V,c_H,\alpha_{\rm supcon},\tau_{\rm supcon},
    E,N_{\rm mini},\eta)$\;
    $\mathcal D\leftarrow
    \textsc{ComputeGAE}(\mathcal B;\xi_{\rm train})$\;
    \For{$e=1,\ldots,E$}{
        \ForEach{minibatch $\mathcal M\subset\operatorname{Shuffle}(\mathcal D)$ with $|\mathcal M|=N_{\rm mini}$}{
            $\boldsymbol\ell_\theta,V_\theta,\mathbf H^{\rm pre}            \leftarrow\textsc{Corrector.Forward}(\mathbf G;\theta,\mathcal I,\xi_{\rm net})$
            for states $\mathbf G$ in $\mathcal M$\;
            $p_{\theta,i}\leftarrow\sigma(\ell_{\theta,i})$
            for $i\in[m]\setminus\mathcal J_0$\;
            $\log\pi_\theta\leftarrow
            \sum_{i\in[m]\setminus\mathcal J_0}
            \bigl[\mathbf 1[i\in I]\log p_{\theta,i}
            +\mathbf 1[i\notin I]\log(1-p_{\theta,i})\bigr]$\;
            $\mathcal H_\theta\leftarrow
            \sum_{i\in[m]\setminus\mathcal J_0}
            \bigl[-p_{\theta,i}\log p_{\theta,i}
            -(1-p_{\theta,i})\log(1-p_{\theta,i})\bigr]$\;
            $\bar{\mathcal H}_\theta\leftarrow\operatorname{mean}(\mathcal H_\theta)$\;
            $\mathcal L_{\rm pol}\leftarrow-\operatorname{mean}\bigl[\min(r_\theta\widehat A,\operatorname{clip}(r_\theta,1-\epsilon_{\rm clip},1+\epsilon_{\rm clip})\widehat A)\bigr]$, where $r_\theta=\exp(\log\pi_\theta-\log\pi^{\rm old})$\;
            $V_{\rm clip}\leftarrow
            V^{\rm old}+\operatorname{clip}(V_\theta-V^{\rm old},
            -\epsilon_V,\epsilon_V)$\;
            $\mathcal L_V\leftarrow
            \operatorname{mean}\bigl[\max((V_\theta-\widehat R)^2,
            (V_{\rm clip}-\widehat R)^2)\bigr]$\;
            $\mathbf y\leftarrow
            \textsc{SpectralPartition}(\mathbf G)$
            for each state in $\mathcal M$\;
            $\mathcal L_{\rm supcon}\leftarrow$ average of
            \textsc{SupConLoss}$(\mathbf H^{\rm pre},\mathbf y;\tau_{\rm supcon})$
            over states in $\mathcal M$\;
            $\mathcal L\leftarrow
            \mathcal L_{\rm pol}+c_V\mathcal L_V
            -c_H\bar{\mathcal H}_\theta
            +\alpha_{\rm supcon}\mathcal L_{\rm supcon}$\;
            \textsc{OptimizerStep}$(\mathcal O,\mathcal L)$\;
        }
    }
    \Return $\pi_\theta,\mathcal O$\;
}
\end{algorithm}

\subsubsection{Optimizer}
\label{sec:optimization}
The Corrector contains a deep stack of matrix-valued transformations in the
encoder and self-attention layers.  For these interior matrices, Muon-style
orthogonalized updates can improve conditioning by normalizing the update
geometry of weight matrices rather than treating every entry independently.  We
therefore allow for a Muon update on the matrix-valued weights of the encoder and
attention stack.

The output heads and scalar-like parameters are treated differently.  Policy
and value heads directly map the representation to action logits and scalar
value estimates, so preserving the ordinary gradient direction is important,
especially for the value regression objective.  Biases, normalization
parameters, vectors and scalars also do not have the same matrix-update
geometry.  These parameters are therefore optimized with AdamW.  When Muon is
disabled, \textsc{BuildOptimizer} simply returns AdamW over all parameters.

Gradient routing is determined by the computational graph: the PPO policy and
entropy terms flow through the encoder, attention stack and policy head; the
value term flows through the encoder, attention stack and value head; and the
SupCon term, being attached to $\mathbf H^{\rm pre}$, contributes only to the
encoder parameters. 

\begin{algorithm}[ht]
\caption{\textsc{BuildOptimizer} and \textsc{OptimizerStep}}
\label{alg:optimizer}
\KwIn{parameters $\theta$; training config $\xi_{\rm train}$}
\KwOut{optimizer object $\mathcal O$}
\Fn{\textsc{BuildOptimizer}$(\theta;\xi_{\rm train})$}{
    $\xi_{\rm train}=(\ldots,\eta)$\;
    Partition $\theta$ into $(\theta_{\rm enc},\theta_{\rm att},\theta_{\rm pol},\theta_{\rm val})$\;
    \If{Muon is disabled}{
        \Return $\textsc{AdamW}(\theta;\eta)$\;
    }
    $\Theta_{\rm Muon}\leftarrow$ matrix-valued weights in $\theta_{\rm enc}\cup\theta_{\rm att}$\;
    $\Theta_{\rm Adam}\leftarrow\theta\setminus\Theta_{\rm Muon}$
    $\mathcal O_{\rm Muon}\leftarrow\textsc{Muon}(\Theta_{\rm Muon};\eta)$\;
    $\mathcal O_{\rm Adam}\leftarrow\textsc{AdamW}(\Theta_{\rm Adam};\eta)$\;
    \Return $\operatorname{CompositeOptimizer}(\mathcal O_{\rm Muon},\mathcal O_{\rm Adam})$\;
}
\Fn{\textsc{OptimizerStep}$(\mathcal O,\mathcal L)$}{
    $\mathcal O.\textsc{ZeroGrad}()$\;
    Backpropagate $\nabla_\theta\mathcal L$ through the computational graph\;
    $\mathcal O.\textsc{Step}()$\;
}
\end{algorithm}

\subsection{Distributed Training Architecture}
\label{sec:distributed}

The distributed architecture implements the conceptual pipeline in Algorithm~\ref{alg:cotrain} as an execution protocol. It decouples shared tree management and policy learning from computationally expensive sampling, while leaving the underlying mathematical transition rules unchanged. The tree server owns the AATS tree state and serializes all operations that read from or modify the shared state. In parallel, workers execute Filler rollouts and Corrector refinement episodes, while the learner aggregates the resulting trajectories to update the policy. In Algorithm~\ref{alg:distributed}, \textsc{SerializedTraverse}, \textsc{SerializedBackpropagate}, and \textsc{SerializedUpdateBest} denote tree-server critical sections over $(\mathcal T,\mathcal S,\mathbf G^*)$; all rollout execution and neural-network computation are performed on the worker side. Table~\ref{tab:distributed} summarizes the state ownership and functional role of each component in the distributed architecture.

\begin{table}[ht]
    \centering
    \caption{\textbf{State ownership and roles in the distributed PackingStar architecture.}}
    \label{tab:distributed}
    \small
\begin{tabular}{@{}p{0.17\linewidth}p{0.26\linewidth}p{0.48\linewidth}@{}}
    \toprule
    \textbf{Component} & \textbf{State owned} & \textbf{Role in the protocol} \\
    \midrule
    Tree server & $\mathcal T,\mathcal S,\mathbf G^*$ &
    serializes traversal, expansion and backpropagation of shared tree statistics \\
    Workers $\mathcal W$ & local rollout state and local trajectories &
    run Filler rollout and $T$ Corrector refinement steps in parallel \\
    Learner & policy $\pi_\theta$ and optimizer state &
    aggregates trajectories and performs minibatch policy updates \\
    \bottomrule
\end{tabular}
\end{table}

\begin{algorithm}[ht]
\caption{\textsc{DistributedPackingStar}: distributed training architecture}
\label{alg:distributed}
\KwIn{problem instance $\mathcal I$; configs $\xi_{\rm AATS},\xi_{\rm net},\xi_{\rm train}$; worker pool $\mathcal W$}
\KwOut{trained policy $\pi_\theta$; best feasible Gram matrix $\mathbf G^*$}
\Fn{\textsc{DistributedPackingStar}$(\mathcal I,\xi_{\rm AATS},\xi_{\rm net},\xi_{\rm train},\mathcal W)$}{
    Initialize shared tree state $(\mathcal T,\mathcal S,\mathbf G^*)$\;
    Initialize Corrector policy $\pi_\theta$ using $\xi_{\rm net}$\;
    $\mathcal O\leftarrow\textsc{BuildOptimizer}(\theta;\xi_{\rm train})$\;
    \For{$\text{iter}=1,\ldots,K$}{
        Broadcast $\theta$ to workers\;
        $\mathcal B\leftarrow\emptyset$\;
        \While{$|\mathcal B|<N_{\rm traj}$}{
            \ForEach{$u\in\mathcal W$ \textbf{in parallel}}{
                $\mathbf G_{\rm leaf},\mathcal P,h\leftarrow
                \textsc{SerializedTraverse}(\mathcal T,\mathcal S;\mathcal I,\xi_{\rm AATS})$\;
                \If{$h=\textsc{EarlyBackprop}$}{\textbf{continue}}
                $\tau_u,\mathbf G_{u,T}\leftarrow
                \textsc{WorkerEpisode}(\mathbf G_{\rm leaf},\pi_\theta;\mathcal I,\xi_{\rm AATS},\xi_{\rm net},\xi_{\rm train})$\;
                \textsc{SerializedBackpropagate}$(\mathcal T,\mathcal P,\mathrm{rows}(\mathbf G_{u,T}),\mathcal S;\mathcal I)$\;
                $\mathbf G^*\leftarrow
                \textsc{SerializedUpdateBest}(\mathbf G^*,\mathbf G_{u,T})$\;
                $\mathcal B\leftarrow\mathcal B\cup\{\tau_u\}$\;
            }
        }
        $\pi_\theta,\mathcal O\leftarrow
        \textsc{Corrector.Update}(\pi_\theta,\mathcal O,\mathcal B;\mathcal I,\xi_{\rm net},\xi_{\rm train})$\;
    }
    \Return $\pi_\theta,\mathbf G^*$\;
}
\end{algorithm}

\subsection{Matrix Decomposition and Initialization for Subsequent Games}
\label{sec:matrix-decomposition}

The output of a completed game is a saturated
Gram matrix which can be decomposed into subconfigurations and reused to initialize subsequent games.
This provides the mechanism by which \textit{PackingStar} transfers discovered
motifs across dimensions and problem instances.
The full procedure is summarized
in Algorithm~\ref{alg:decomposition-seeded-packingstar}. 
The partitioning rule is given
by the recursive row-spectrum partition in Algorithm~\ref{alg:spectral-partition}.
Rows assigned the same terminal row-spectrum label define a relational motif.
The corresponding submatrix can then be used as a seed Gram matrix
for a subsequent game, biasing the search toward better configurations.

\begin{algorithm}[ht]
\caption{\textsc{DecompositionSeededPackingStar}: spectral decomposition and subsequent games}
\label{alg:decomposition-seeded-packingstar}
\KwIn{initial seed bank $\mathcal Q_0$ of records $\mathbf G_{\rm seed}$; configs $\xi_{\rm AATS},\xi_{\rm net},\xi_{\rm train}$; worker pool $\mathcal W$; game budget $B$}
\KwOut{completed configuration library $\mathcal L$}
\Fn{\textsc{DecompositionSeededPackingStar}$(\mathcal Q_0,\xi_{\rm AATS},\xi_{\rm net},\xi_{\rm train},\mathcal W,B)$}{
     $\mathcal Q\leftarrow \mathcal Q_0$, $\mathcal L\leftarrow\emptyset$\;
    \While{$\mathcal Q\neq\emptyset$ and fewer than $B$ games have been launched}{
        Select a seed record $\mathbf G_{\rm seed}$ from $\mathcal Q$\;
        Set the problem instance $\mathcal I\leftarrow(n,\mathbf G^{(m_0)}, \mathcal C_1,\mathcal C_2,\mathcal C_\ast)$ according to the seed\;
        $\pi_\theta,\overline{\mathbf G}\leftarrow
        \textsc{DistributedPackingStar}(\mathcal I,\xi_{\rm AATS},\xi_{\rm net},\xi_{\rm train},\mathcal W)$\;
        $\mathbf y\leftarrow\textsc{SpectralPartition}(\overline{\mathbf G})$\;
        Insert $\overline{\mathbf G}$ into $\mathcal L$\;
        \ForEach{spectral label $a$ appearing in $\mathbf y$}{
            $J_a\leftarrow(i\in[\mathrm{rows}(\overline{\mathbf G})]:y_i=a)$ in inherited row order\;
            $\mathbf H_a\leftarrow\overline{\mathbf G}_{J_a,J_a}$\;
            $d_a\leftarrow\operatorname{rank}(\mathbf H_a;\epsilon_{\rm rank})$\;
            $\widetilde{\mathbf H}_a\leftarrow\textsc{CanonicalSeedOrder}(\mathbf H_a,d_a)$\;
            Insert $\widetilde{\mathbf H}_a$ into $\mathcal Q$\;
        }
    }
    \Return $\mathcal L$\;
}
\end{algorithm}

Let $\overline{\mathbf G}\in\mathbb R^{M\times M}$ be a terminal Gram
matrix, and let
\begin{equation}
    \mathbf y=\textsc{SpectralPartition}(\overline{\mathbf G})
\end{equation}
denote the recursive row-spectrum labels produced by
Algorithm~\ref{alg:spectral-partition}.  For each label $a$, define
\begin{equation}
    J_a=\{i\in[M]:y_i=a\},\qquad
    \mathbf H_a=\overline{\mathbf G}_{J_a,J_a},\qquad
    d_a=\operatorname{rank}(\mathbf H_a).
\end{equation}
The fragment $\mathbf H_a$ is treated as a candidate seed
with rank $d_a$ showing its intrinsic dimension.  
Before insertion into the seed bank,
we apply \textsc{CanonicalSeedOrder} to expose a full-rank core in the leading
rows and columns. This produces a stable representation of the fragment for
use in later games.

The seed bank is a catalog of promising substructures.
For a subsequent game, a seed is selected from the bank and the new problem instance
$\mathcal I=(n,\mathbf G^{(m_0)},\mathcal C_1,\mathcal C_2,\mathcal C_\ast)$ is
specified according to the intended continuation experiment.   
Iterating this construction yields the exploration scheme
\begin{equation}
    \overline{\mathbf G}
    \longmapsto
    \{\mathbf H_a,d_a\}_{a}
    \longmapsto
    \{\mathcal I_a=(n_a,\mathbf G^{(m_0)}_a,\mathcal C_{1,a},
       \mathcal C_{2,a},\mathcal C_{\ast,a})\}_{a}
    \longmapsto
    \{\overline{\mathbf G}'\},
\end{equation}
where typically $n_a=d_a$. The loop combines mathematical bias with 
algorithmic exploration.  Within each game, the Filler and Corrector improve the
quality of local completions; across games, the decomposition step transfers
discovered motifs between dimensions, problem instances, and cosine regimes.
The repeated cycle of seed decomposition, game initialization, and
training moves the search through related extremal configuration spaces, rather
than repeatedly restarting from empty or unrelated seeds.

\section{Optimality certificates for prescribed inner-product sets}
\label{app:prescribed-inner-product-optimality}
We follow the notation and harmonic-analysis framework of Delsarte, Goethals, and Seidel \cite{delsarte1977}. Let $\Omega_n$, with measure $\omega_n$, denote the unit sphere in the Euclidean space $ \mathbb{R}^n$
of dimension $n$, endowed with the inner product $\langle,\rangle$. For any $k\geq0$, let $ Hom(k)= Hom_n(k)$ denote the linear space of all functions $V: \Omega_n\rightarrow \mathbb{R}$ which are represented by polynomials $V(\zeta) = V(\zeta_1, ..., \zeta_n)$, homogeneous of total degree $k$ in the $n$ variables $\zeta_i$. Let $Harm(k)$ denote the subspace of $Hom(k)$ consisting of all functions represented by harmonic polynomials of degree $k$. Then $Harm(k)$ is invariant under the orthogonal group $O(n)$ of $\mathbb{R}^n$. Any function $V \in Hom(k)$ can be uniquely written as:

\begin{equation}
V(\zeta)
=
\sum_{i=0}^{\left\lfloor k/2\right\rfloor}
\langle\zeta,\zeta\rangle^i
W_{k-2i}(\zeta),
\qquad
W_{k-2i}\in Harm(k-2i).
\end{equation}

\begin{definition}
	For any finite non-empty set $X\subset \Omega_n$ of size $m$, for any orthogonal basis $\{W_{k,i}\}$ of $Harm(k)$, with norm $W_{k,i}=\omega_n^{1/2}$ and for any fixed numbering of these, the $m \times \dim\ Harm(k)$ matrix 
	\begin{equation}
		H_k := [ W_{k,i}(\xi) ],\  \xi\in X,\ i\in\{1, 2, ..., \dim\ Harm(k)\},
	\end{equation}
	is called the $k$-th characteristic matrix. Thus, $H_0$ is the all-one vector of size $m$. 
\end{definition}

\begin{definition}[Spherical $\mathcal{C}$-code]
	Let $\mathcal{C}$ be a subset of the interval $[-1,1]$. A spherical $\mathcal{C}$-code, for short a $\mathcal{C}$-code, is a non-empty subset $X$ of the unit sphere in $ \mathbb{R}^n$, satisfying $\langle\xi,\eta\rangle \in \mathcal{C}$, for all $\xi\neq\eta \in X$.
\end{definition}
Thus, a $\mathcal{C}$-code is a set of unit vectors with angles from the prescribed set $\arccos \mathcal{C}$, or a set of points on $\Omega_n$ with distances from the prescribed set $\{
\sqrt{2(1-\alpha)}
:
\alpha\in \mathcal{C}
\}$.

\begin{definition}[Compatible]
	A polynomial
	$F(x)\in\mathbb R[x]$
	is said to be compatible with
	$\mathcal{C}$
	if
	\begin{equation}
	F(\alpha)\le0
	\quad
	\text{for all }\alpha\in \mathcal{C}.
	\end{equation}
\end{definition}
We now recall the classical Delsarte linear-programming bound.

\begin{theorem}[Delsarte \cite{delsarte1977}]\label{thm:lp-bound}
	Let $F(x)$, with Gegenbauer coefficients $f_0 > 0$ and $f_k \geq 0$, for all $k$, be compatible with the set $\mathcal{C}$. Then the cardinality $m=|X|$ of any $\mathcal{C}$-code $X$ satisfies
	\begin{equation}
		m \leq F(1)/f_0
	\end{equation}
	Equality holds if and only if, for all $\xi\neq\eta \in X$, and for all $k\geq 1$:
	\begin{equation}
		F\left(\langle\xi,\eta\rangle\right)= 0, \ f_kH_k^\top H_0= \boldsymbol{0}
	\end{equation}
	
\end{theorem}
The bound of Theorem~\ref{thm:lp-bound} is the foundation upon which essentially all subsequent linear- and semidefinite-programming bounds for spherical codes are built. Its starting point, the addition formula and the nonnegative-coefficient expansion of $O(n)$-invariant continuous positive-definite kernels on $\Omega_n$, is due to Schoenberg \cite{schoenberg1942}; the linear-programming method itself originates with Delsarte's earlier work on association schemes \cite{delsarte1973}. For the half-infinite case $\mathcal{C} = [-1, \cos\theta]$, Levenshtein \cite{levenshtein1979, levenshtein1992} obtained explicit analytic optimisers, leading to the asymptotic Kabatyanski\u\i--Levenshtein bound on the density of spherical codes \cite{kabatyanskii1978}. Bachoc and Vallentin \cite{bachoc2008} sharpened the bound by introducing matrix-valued kernels and three-point semidefinite programmes, which were subsequently unified into a hierarchy of $k$-point semidefinite bounds by de~Laat and Vallentin \cite{delaat2015} and by de~Laat, Machado, Oliveira, and Vallentin \cite{delaat2021}. Most recently, Cohn, de~Laat, and Leijenhorst \cite{cohn2024} have demonstrated how high-precision SDP solutions, combined with rounding to algebraic numbers, yield exact optimality proofs for several sharp configurations.

\subsection{Specialisation to finite \texorpdfstring{$\mathcal{C}$}{C}}

In our work, the inner-product set of every $\mathcal{C}$-code under consideration is a known finite subset
\begin{equation}
	\mathcal{C} \;=\; \{\alpha_1,\, \alpha_2,\, \ldots,\, \alpha_s\}
	\;\subset\; [-1, 1).
\end{equation}
This regime arises naturally whenever the inner products of a candidate code $X$ are completely determined by a small set of algebraic values --- as is the case, for instance, for codes with prescribed automorphism group or codes derived from association schemes. In this finite-$\mathcal{C}$ setting the compatibility constraint $F(\alpha) \leq 0$ for all $\alpha \in \mathcal{C}$ collapses to $s$ scalar inequalities $F(\alpha_i) \leq 0$, $i = 1, \ldots, s$, and Theorem~\ref{thm:lp-bound} reduces to a small linear program in the Gegenbauer coefficients of~$F$.

We adopt throughout the standard normalisation $Q_k(1) = 1$ for all $k \geq 0$, so that $F(1) = \sum_{k} f_k$. Fix a degree bound $r \in \mathbb{N}$ on $F$. The Delsarte primal LP for the size of any $\mathcal{C}$-code in $\Omega_n$ is

\begin{equation}\label{eq:lp-primal}
	\mathrm{LP}_r(n, \mathcal{C}) \;:=\;
	\begin{aligned}[t]
		\min_{f_0, \ldots, f_r}\quad & \sum_{k=0}^{r} f_k \\
		\text{subject to}\quad
		& \sum_{k=0}^{r} f_k\, Q_k(\alpha_i) \;\leq\; 0,
		\qquad i = 1, \ldots, s, \\
		& f_0 \;=\; 1, \\
		& f_k \;\geq\; 0, \qquad k = 1, \ldots, r.
	\end{aligned}
\end{equation}
The normalisation $f_0 = 1$ is without loss of generality, since the bound $F(1)/f_0$ in Theorem~\ref{thm:lp-bound} is invariant under positive scaling of~$F$. Both \eqref{eq:lp-primal} and its dual are feasible and bounded, so strong LP duality applies; the dual program reads

\begin{equation}\label{eq:lp-dual}
	\mathrm{LP}_r(n, \mathcal{C}) \;=\;
	\begin{aligned}[t]
		\max_{\lambda \in \mathbb{R}^s_{\geq 0}}\quad
		& 1 \,+\, \sum_{i=1}^{s} \lambda_i \\
		\text{subject to}\quad
		& \sum_{i=1}^{s} \lambda_i\, Q_k(\alpha_i) \;\geq\; -1,
		\qquad k = 1, \ldots, r.
	\end{aligned}
\end{equation}
The dual variables $\lambda_i$ admit a transparent geometric interpretation, which we record next.

\begin{theorem}[Distance-distribution interpretation of the dual]
	\label{thm:dual}
	Let $X \subset \Omega_n$ be a $\mathcal{C}$-code of size $m$. Define its (normalised) distance distribution by
	\begin{equation}\label{eq:dist-dist}
		\lambda_i^{X}
		\;:=\; \frac{1}{m}\,
		\bigl|\,\{\,(\xi, \eta) \in X \times X \,:\,
		\xi \neq \eta,\ \langle \xi, \eta \rangle = \alpha_i \,\}\,\bigr|,
		\qquad i = 1, \ldots, s.
	\end{equation}
Then $\lambda^{X} = (\lambda_1^{X}, \ldots, \lambda_s^{X})$ is feasible for the dual program \eqref{eq:lp-dual}, and the corresponding dual objective value equals~$m$. In particular, when $\mathrm{LP}_r(n, \mathcal{C})$ is integer-valued and is attained by some code $X^{\star}$, the optimal dual measure $\lambda^{\star}$ coincides with the distance distribution of~$X^{\star}$.
\end{theorem}

\begin{proof}
	The addition formula and Schoenberg's positivity together yield
	\begin{equation}
		\sum_{\xi, \eta \in X} Q_k(\langle \xi, \eta \rangle) \,=\, \tfrac{1}{h_k(n)}\, H_k^{\top} H_k\,(\mathbf{1}, \mathbf{1}) \,\geq\, 0
	\end{equation}
	for every $k \geq 0$, where $h_k(n) = \dim \mathrm{Harm}(k)$. Splitting the diagonal contribution from the off-diagonal part and using $Q_k(1) = 1$ gives
	\begin{equation}
		0 \;\leq\; \sum_{\xi, \eta \in X} Q_k(\langle \xi, \eta \rangle)
		\;=\; m \,+\, m \sum_{i=1}^{s} \lambda_i^{X}\, Q_k(\alpha_i),
	\end{equation}
	which on rearrangement is exactly the dual constraint of \eqref{eq:lp-dual}. The dual objective at $\lambda^{X}$ equals $1 + \sum_i \lambda_i^{X} = 1 + (m - 1) = m$ by counting all ordered pairs in $X \times X$. The final assertion follows from complementary slackness applied to optimal primal-dual pairs.
\end{proof}

Theorem~\ref{thm:dual} demonstrates that \emph{the LP dual extracts a complete combinatorial fingerprint --- the inner-product multiplicity profile --- of any extremal $\mathcal{C}$-code, without ever having access to its coordinates}.

\subsection{Three-point semidefinite programming bound}

The bound of Theorem~\ref{thm:lp-bound} captures only the pairwise correlations of a $\mathcal{C}$-code $X$. A stronger bound is obtained by incorporating semidefinite constraints arising from triples of points in $X$, yielding the \emph{three-point bound} of Bachoc and Vallentin \cite{bachoc2008}. In this work we adopt the explicit formulation and the exact-arithmetic implementation framework of Cohn, de~Laat, and Leijenhorst \cite{cohn2024}.

For each pair of nonnegative integer degree bounds $(d_2, d_3)$, the three-point semidefinite programme --- which we denote by $\mathrm{SDP}_{d_2, d_3}(n, \mathcal{C})$ --- optimises over a tuple $\bigl(\,(a_k)_{k = 0}^{2 d_2},\ (F_k)_{k = 0}^{d_3}\,\bigr)$ consisting of nonnegative scalars $a_k \geq 0$ (the two-point part) and positive semidefinite matrices $F_k \succeq 0$ of size $(d_3 - k + 1) \times (d_3 - k + 1)$ (the three-point part), subject to two families of constraints:

\begin{itemize}
	\item a \emph{two-point constraint} for each $\alpha \in \mathcal{C}$, generalising the inequality $F(\alpha) \leq 0$ of \eqref{eq:lp-primal} from a scalar polynomial to a sum of scalar Gegenbauer terms and a matrix-valued kernel evaluated at the degenerate triple $(\alpha, \alpha, 1)$;
	\item a \emph{three-point constraint} for each \emph{feasible triple}
	\begin{equation}\label{eq:feasible-triples}
		(u, v, t) \,\in\, T(\mathcal{C}) \;:=\; \bigl\{\, (u, v, t) \in (\mathcal{C} \cup \{1\})^3 \;:\; 1 + 2uvt - u^2 - v^2 - t^2 \,\geq\, 0 \,\bigr\},
	\end{equation}
	where the determinantal inequality is the condition for the $3 \times 3$ Gram matrix of any three points realising those pairwise inner products to be positive semidefinite, i.e., for the triple $(u, v, t)$ to actually arise from three points of $\Omega_n$.
\end{itemize}
Both constraint families involve the $S_3$-symmetric matrix-valued kernel $S^{(n)}_k(u, v, t)$ constructed in \cite{cohn2024} from the Gegenbauer polynomials in dimension $n - 1$ via the cleared-radicals substitution
\begin{equation}\label{eq:cleared-radicals}
	Q^{(n-1)}_k\!\bigl(\tfrac{t - uv}{\sqrt{(1 - u^2)(1 - v^2)}}\bigr)
	\cdot \bigl[(1 - u^2)(1 - v^2)\bigr]^{k/2}
	\;\in\; \mathbb{Q}[u, v, t],
\end{equation}
which removes the irrational factors that would otherwise arise in the conditional inner product of a third point relative to the first two; we refer to \cite[Sec.~3]{cohn2024} for the explicit derivation. The bound satisfies
\begin{equation}
	m \;\leq\; \mathrm{SDP}_{d_2, d_3}(n, \mathcal{C}) \;\leq\; \mathrm{LP}_{2 d_2}(n, \mathcal{C})
\end{equation}
for any $\mathcal{C}$-code $X \subset \Omega_n$ of size $m$, with the second inequality typically strict.

\paragraph{Implementation.}
We implement the three-point SDP in Julia, using the package \href{https://github.com/nanleij/ClusteredLowRankSolver.jl}{ClusteredLowRankSolver} of Leijenhorst \cite{cohn2024} as the SDP backend. The polynomial coefficients of the kernel $S^{(n)}_k$, the cleared-radicals expansion \eqref{eq:cleared-radicals}, and the Gegenbauer basis used in the two-point part are constructed symbolically over $\mathbb{Q}$ via \href{https://nemocas.github.io/Nemo.jl/stable/}{Nemo.jl}, so that the SDP coefficient matrices are exact rationals. The solver itself is run at $256$-bit floating-point precision with a duality gap threshold of $10^{-20}$ --- well below the tolerance required to identify the integer ceiling of the bound for the cases studied in this paper.

\begin{table}[t]
	\centering
	\small
	\setlength{\tabcolsep}{4pt}
	\begin{tabular}{ccccc}
		\toprule
		dimension $n$ & inner-product set $\mathcal C$ & primal objective & dual objective & integer upper bound \\
		\midrule
		$12$ &
		$\{-1/2,-1/8,1/4\}$ &
		$81 + 9.51 \times 10^{-19}$ &
		$81 - 2.6 \times 10^{-20}$ &
		$81$ \\

		$20$ &
		$\{-1/2,-1/8,1/4\}$ &
		$405 + 1.29 \times 10^{-18}$ &
		$405 - 3.5 \times 10^{-19}$ &
		$405$ \\

		$21$ &
		$\{-1/2,-1/8,1/4\}$ &
		$567 + 6.08 \times 10^{-18}$ &
		$567 - 1.66 \times 10^{-18}$ &
		$567$ \\

		$22$ &
		$\{-1,-1/5,1/5\}$ &
		$352 + 6.72 \times 10^{-19}$ &
		$352 - 4.67 \times 10^{-19}$ &
		$352$ \\
		\bottomrule
	\end{tabular}
	\caption{
		Three-point SDP upper bounds on $|X|$ for prescribed
		inner-product sets $\mathcal C$, computed at $256$-bit precision.
		The primal-dual gap is below $7\times10^{-21}$ for the first three
		rows and below $2\times10^{-21}$ for the 22-dimensional case.
	}
	\label{tab:sdp-results}
\end{table}

\paragraph{Application to the discovered configurations.}
For the constraint sets
$
\mathcal{C}_1=\{-1/2,\,-1/8,\,1/4\}$ and 
$
\mathcal{C}_2=\{-1,\,-1/5,\,1/5\},
$
we use degree parameters $(d_2,d_3)=(6,4)$ for
$\mathcal C_1$ and $(d_2,d_3)=(10,10)$ for
$\mathcal C_2$. The resulting three-point SDP upper bounds are shown in
Table~\ref{tab:sdp-results}. The high-precision primal--dual solutions identify
the integer ceilings unambiguously, giving upper bounds
$|X|\le81$, $|X|\le405$, $|X|\le567$, and $|X|\le352$
in dimensions $12$, $20$, $21$, and $22$, respectively.
Together with the explicit constructed $\mathcal C$-codes of sizes
$81$, $405$, $567$, and $352$, this yields:
\begin{equation}
\begin{alignedat}{2}
&\max\bigl\{\,|X| : X \subset \Omega_{12},\
\langle \xi,\eta\rangle \in \{-1/2,-1/8,1/4\}
\text{ for all } \xi\neq\eta\in X\,\bigr\}
&&= 81, \\
&\max\bigl\{\,|X| : X \subset \Omega_{20},\
\langle \xi,\eta\rangle \in \{-1/2,-1/8,1/4\}
\text{ for all } \xi\neq\eta\in X\,\bigr\}
&&= 405, \\
&\max\bigl\{\,|X| : X \subset \Omega_{21},\
\langle \xi,\eta\rangle \in \{-1/2,-1/8,1/4\}
\text{ for all } \xi\neq\eta\in X\,\bigr\}
&&= 567, \\
&\max\bigl\{\,|X| : X \subset \Omega_{22},\
\langle \xi,\eta\rangle \in \{-1,-1/5,1/5\}
\text{ for all } \xi\neq\eta\in X\,\bigr\}
&&= 352.
\end{alignedat}
\end{equation}

\begin{remark}[Towards fully exact bounds]
	Because the primal SDP objective is itself a feasible upper bound on $|X|$ at $256$-bit precision, the integer bounds in Table~\ref{tab:sdp-results} are rigorous as stated, no symbolic post-processing being strictly necessary to identify the integer ceiling. Should an exact rational (or algebraic) certificate be desired --- for instance, to verify the bound at machine-checkable precision, or to apply the framework over a number field other than $\mathbb{Q}$ --- one applies the rounding-and-verification procedure of \cite{cohn2024}, which projects the BigFloat primal solution onto a feasible point over a prescribed number field and certifies the resulting bound symbolically. In each instance of Table~\ref{tab:sdp-results} the projection step is straightforward, since the optimum is integer-valued and the corresponding optimal face is one-dimensional.
\end{remark}

\section{Mathematical landscape}

The AI system \textit{PackingStar} has brought back from the high-dimensional
combinatorial world a large and varied collection of spherical codes, some of
which yield new records. But records are only the first way in which these
configurations should be read. They exhibit rich and diverse mathematical
features, and form an exceptional corpus of objects to be observed and
understood. In this part, we study the mathematical structures behind the
configurations found by \textit{PackingStar}. Its record configurations often organize
into families of non-isometric codes, sometimes with common local statistics,
while special members of these families carry large symmetry groups and reveal
classical finite geometries. The families themselves also exhibit structured
relations with one another. The purpose of this part is to describe this
structural layer.

The first layer consists of individual configurations and the construction
mechanisms which explain them. Here the central object is still a spherical
code: a finite subset of a Euclidean sphere satisfying a prescribed
inner-product constraint. In some cases, the code is governed by classical
finite geometry, such as the \(W(5,2)\)--\(E_7\)--\(Q^+(7,2)\) dictionary. In
other cases, it is explained by new construction mechanisms such as the fiber
tensor product. This layer includes several explicit record configurations,
such as \(\texttt{14d252k2}\), \(\texttt{13d1146k1}\),
\(\texttt{17d578k2}\), and the Leech-based kissing number configurations in dimensions
\(25\)--\(31\).

The second layer concerns regularity without symmetry. For a fixed cosine set,
\textit{PackingStar} often reveals not a unique configuration, but a structured region
of extremal configurations. This motivates the
\emph{extremal configuration space}. When a supported version of the Delsarte--Goethals--Seidel linear-programming
bound~\cite{DelsarteGoethalsSeidel} or some other method proves optimality within the same cosine set, we obtain an
\emph{optimal configuration space}, which we shall call a
\emph{constellation}. The main examples come from the cosine set
\(\{1/4,-1/8,-1/2\}\), where \(\texttt{12d81k3}\),
\(\texttt{20d405k3}\), and \(\texttt{21d567k3}\) are certified by the
supported-LP method. Related examples such as \(\texttt{15d135k3}\) and
\(\texttt{18d243k3}\) are governed by the same LP polynomial and an associated
integrality condition. Every configuration in these five constellations is also balanced in Leech's mechanical sense~\cite{Leech1957,CohnElkiesKumarSchuermann2010}: if the points are
viewed as particles constrained to the sphere, then for every pairwise force law
depending only on distance, the net tangential force at each point vanishes. The
\(\texttt{12d81k3}\) constellation contains at least \(31\) non-isomorphic
configurations with the same first statistic (distance multiset by points) but different global symmetries.
Thus the order observed here does not simply come from a large automorphism
group, but from a form of extremal regularity produced by the geometry of the
constraint itself. We also give a list of such constellations currently
known to us.

The third layer concerns the special symmetric representatives that sometimes
exist inside constellations. We call such representatives \emph{Stars}. In the
\(\{1/4,-1/8,-1/2\}\) series, the Stars arise as subspace configurations of the
universally optimal \(\texttt{22d891k3}\) code~\cite{CohnKumar2007}. The Fischer geometry of
\(U_6(2)=\mathrm{Fi}_{21}\) provides a natural language for these structures,
and we call the resulting family of subspace codes the \emph{Fischer tower}.
This observation leads to the \emph{Genealogy} method for spherical codes.
Instead of studying a code only as an isolated extremal object, we study it
through its subspace descendants. Using the genetic subspace algorithm, we
explore the genealogy of the minimal vectors of the Leech lattice and other
special spherical codes, and obtain many new lower-bound records.

The fourth layer is Euclidean representations and subgroup geometry. This
viewpoint is motivated by Conway's geometric use of the Leech lattice~\cite{ConwaySloaneSPLAG} and by
the ATLAS tradition of making finite simple groups concrete through character
tables, explicit representations, and maximal-subgroup data~\cite{Atlas1985}. In our setting, a
spherical code gives a Euclidean realization of representation-theoretic and
subgroup-geometric data: its automorphism group acts on the code, giving at
once an orthogonal representation and a permutation representation, while its
metric subconfigurations realize and organize subgroup structure. We begin with
the baby example of the icosahedral code and \(A_5\), and then develop
Euclidean representations for \(U_6(2)=\mathrm{Fi}_{21}\), \(U_4(2)\), and
\(Sp_6(2)\). In these examples, many maximal subgroups admit simple metric
descriptions. Finally, we construct a \(78\)-dimensional
\(1/4\)-spherical code representing the sporadic simple group
\(\mathrm{Fi}_{22}\), and conjecture a characterization of \(\mathrm{Fi}_{22}\)
by purely Euclidean-geometric conditions.

The labels used above, such as \(\texttt{14d252k2}\), are shorthand for the
generalized kissing number configurations studied throughout this part. We now fix
this notation.

\begin{definition}[Generalized kissing number configuration and label]
\label{def:generalized-kissing}
Let \(D,N,K\) be positive integers. A \emph{\(D\mathrm{d}N\mathrm{k}K\)
configuration} is a finite set
\begin{equation}
C\subset S^{D-1}\subset \mathbb R^D,
\qquad |C|=N,
\end{equation}
such that
\begin{equation}
\langle x,y\rangle \le \frac{1}{K+1}
\qquad
\text{for all distinct }x,y\in C.
\end{equation}
Equivalently, it is a \((D,N,1/(K+1))\)-spherical code. The symbols
\(\mathrm d\) and \(\mathrm k\) are separators in the label, not operations.
Thus \(\texttt{14d252k2}\) means \(252\) unit vectors in
\(S^{13}\subset\mathbb R^{14}\) with maximal inner product at most \(1/3\).
\end{definition}

The thresholds \(1/(K+1)\) have a direct kissing interpretation. If \(K\)
mutually tangent unit spheres are fixed in \(\mathbb R^{D+K-1}\), then the
centers of all further unit spheres touching every fixed sphere lie, after
translation and scaling, on \(S^{D-1}\) \cite{BannaiSloane}. Under this
normalization, pairwise non-overlap is exactly
\begin{equation}
\langle x,y\rangle \le \frac{1}{K+1}.
\end{equation}
For \(K=1\) this is the ordinary kissing number configuration; for larger \(K\) we call
it a \emph{generalized kissing number configuration}. A secondary reason for this
parameterization is that
\begin{equation}
K=c^{-1}-1
\end{equation}
is the simplest function sending the non-trivial cosine domain of spherical codes, i.e.
\(c\in(0,1)\) to $(0,+\infty)$, with the ordinary kissing threshold
\(c=1/2\) corresponding to \(K=1\). Thus the labels \(\mathrm{k}1\),
\(\mathrm{k}2\), and \(\mathrm{k}3\) correspond respectively to the thresholds
\(1/2\), \(1/3\), and \(1/4\). The label records only the parameters
\((D,N,K)\); it does not assert uniqueness or optimality.

Let \(C=\{x_1,\dots,x_N\}\subset S^{D-1}\) be a spherical code, and let
\begin{equation}
G_{ij}=\langle x_i,x_j\rangle
\end{equation}
be its Gram matrix. For \(x\in C\), define the \emph{local distance distribution} by
\begin{equation}
A_t(x)=\#\{y\in C:\langle x,y\rangle=t\}.
\end{equation}
We include \(y=x\), so \(A_1(x)=1\). Equivalently, \(A_\bullet(x)\) is the multiplicity function of the Gram row indexed by \(x\); it records the number of code points at each inner product, or equivalently at each spherical distance, from \(x\) \cite{DelsarteGoethalsSeidel}.

\begin{definition}[First Statistic]
\label{def:first-stat}
The \emph{first statistic} of \(C\) is the multiset
\begin{equation}
\mathsf{Stat}_1(C)
=
\bigl\{\!\bigl\{A_\bullet(x):x\in C\bigr\}\!\bigr\}.
\end{equation}
\end{definition}
It induces the coarse partition
\begin{equation}
x\equiv_1 x'
\quad\Longleftrightarrow\quad
A_t(x)=A_t(x')\quad\text{for all }t.
\end{equation}
Every orbit of \(\operatorname{Aut}(C)\) lies inside one \(\equiv_1\)-class, but the converse need not hold. Here
\begin{equation}
\operatorname{Aut}(C)=\{U\in O(\mathbb{R}^D):U(C)=C\}.
\end{equation}
For the three antipodal \(\texttt{14d252k2}\) configurations constructed below, the point-orbit counts are \(4,20,20\), but their first statistic is the same:
\begin{equation}\label{eq:first-stat-14d252k2}
\mathsf{Stat}_1(C)=
\left\{\!\left\{
\begin{aligned}
&(0^{112},(\pm\tfrac13)^{69},(\pm1)^1)^{56},\\
&(0^{104},(\pm\tfrac13)^{73},(\pm1)^1)^{196}.
\end{aligned}
\right\}\!\right\}.
\end{equation}
Here \(t^m\) means \(A_t=m\), \((\pm t)^m\) means \(A_t=A_{-t}=m\), and \(T^m\) outside a row means that \(m\) points have local distribution \(T\). Thus the first row says that, for \(56\) points \(x\),
\begin{equation}
A_0(x)=112,\qquad A_{1/3}(x)=A_{-1/3}(x)=69,\qquad A_1(x)=A_{-1}(x)=1,
\end{equation}
and \(112+2\cdot69+2=252\).
\\
Note that having a single first statistic is precisely distance invariance in the sense of DGS~\cite{DelsarteGoethalsSeidel}.
\\
We begin with the basic dictionary behind the decomposition of the \(126\)-point \(E_7\) kissing number configuration into \(9\) copies of the \(7\)-dimensional cross polytope \(X_7\). In kissing-number language this is the identity \(126=9\cdot14\). In finite-geometric language it is governed by the spread geometry of \(W(5,2)\). This hidden geometry reappears throughout our discoveries, so we record the dictionary here in a form suited for later use.

\section{Construction Mechanisms and New Records}
This section extracts construction mechanisms from the configurations discovered by \textit{PackingStar}. Finite-geometric structures such as \(W(5,2)\), \(E_7\) spreads, Barnes--Wall/Nordstrom--Robinson subcodes and Leech-lattice sections reveal reusable blocks and assembly rules behind \(\texttt{14d252k2}\), \(\texttt{13d1146k1}\), \(\texttt{17d578k2}\) and the improved kissing-number lower bounds in dimensions \(25\)--\(31\).

\subsection{The \texorpdfstring{\(W(5,2)\)}{W(5,2)} structure of the \texorpdfstring{\(E_7\)}{E7} roots}

Let \(V=\mathbb F_2^6\), and define
\begin{equation}
b(x,y)=x_1y_4+x_2y_5+x_3y_6+x_4y_1+x_5y_2+x_6y_3 .
\end{equation}
This is the standard symplectic form on \(V\). Since \(|\mathbb F_2^\times|=1\), the points of \(PG(V)\) are the nonzero vectors of \(V\). We write \(W=W(5,2)\). Its lines are
\begin{equation}
\{x,y,x+y\},\qquad x,y\in V\setminus\{0\},\ x\ne y,\ b(x,y)=0,
\end{equation}
and its generators are the \(3\)-dimensional totally isotropic subspaces of \(V\), viewed projectively as Fano planes. Thus \(W\) has \(63\) points, \(315\) lines, and \(135\) generators \cite{PayneThas}.

Let \(R=R(E_7)\) be the \(126\) roots of \(E_7\), and write \(\bar r=\{\pm r\}\) for a root line.

\begin{theorem}
There is a bijection
\begin{equation}
\phi:R/\{\pm1\}\longrightarrow W
\end{equation}
such that, for distinct root lines \(\bar r,\bar s\),
\begin{equation}
\langle r,s\rangle=0
\qquad\Longleftrightarrow\qquad
b(\phi(\bar r),\phi(\bar s))=0 .
\end{equation}
\end{theorem}

\begin{proof}
Let \(L=L(E_7)\). Reduction modulo \(2\) gives an alternating form on \(L/2L\). Its radical is \(1\)-dimensional, so
\begin{equation}
\overline V=(L/2L)/\operatorname{rad}
\end{equation}
is a \(6\)-dimensional symplectic space over \(\mathbb F_2\). The images of the \(63\) root lines are precisely the \(63\) nonzero vectors of \(\overline V\) \cite{CerchiaiVanGeemen,Manivel}; identify \(\overline V\) with \(V\). For distinct roots \(r,s\), one has \((r,s)\in\{-1,0,1\}\), while the mod-\(2\) symplectic pairing is \((r,s)\bmod 2\). Hence it vanishes exactly when \((r,s)=0\).
\end{proof}

For a generator \(P\) of \(W\), the seven root lines \(\phi^{-1}(P)\) are mutually orthogonal; their \(14\) roots form a copy \(X(P)\) of the \(7\)-dimensional cross polytope \(X_7\). Conversely, every \(X_7\subset R\) arises in this way: seven mutually orthogonal root lines map to seven pairwise orthogonal nonzero vectors, whose span is a maximal totally isotropic \(3\)-space.

A spread of \(W\) is a partition of its \(63\) points into nine generators. Therefore spreads of \(W\) are equivalent to unordered decompositions
\begin{equation}
R=X(P_1)\sqcup\cdots\sqcup X(P_9)
\end{equation}
into nine copies of \(X_7\).

\begin{proposition}[computer-assisted enumeration]
There are \(960\) unordered decompositions of the \(E_7\) root configuration into nine copies of \(X_7\).
\end{proposition}

\begin{proof}
It suffices to count spreads of \(W\). The finite exact-cover instance has universe the \(63\) points of \(W\) and blocks the \(135\) generators. The verification enumerates all exact covers by branching on an uncovered point and admitting only disjoint generators; each spread has a unique search path. The resulting count is \(960\).
\end{proof}

The dual description uses the spin embedding of \(DW(5,2)\). Its \(135\) points, namely the generators of \(W\), are realized as the \(135\) points of \(Q^+(7,2)\); under this realization, disjoint generators of \(W\) correspond to non-collinear points of \(Q^+(7,2)\). Hence the \(960\) spreads of \(W\) are the \(960\) ovoids of \(Q^+(7,2)\), where an ovoid is a \(9\)-point set with no two collinear points \cite{LuyckxThas}.

The same incidence appears in the \(56\)-point dual-root code. Let \(\mathcal F\) be the Fano plane on \(\Omega=\{1,\ldots,7\}\), and define
\begin{equation}
C_{56}=
\left\{
\frac1{\sqrt3}\sum_{i\in L}\varepsilon_i e_i:
L\in\mathcal F_{\mathrm{lines}},\ \varepsilon_i\in\{\pm1\}
\right\}
\subset S^6 .
\end{equation}
This is the standard \(56\)-point \(E_7^*\) minimal-vector code [citation needed: identification of this Fano-line sign model with \(E_7^*\)]. It has maximum inner product \(1/3\).

A cube in \(C_{56}\) is an antipodal \(8\)-point subset of real rank \(3\). For such a cube \(Z\), let \(D(Z)\) be its three unoriented edge directions. A cube decomposition
\begin{equation}
C_{56}=Z_1\sqcup\cdots\sqcup Z_7
\end{equation}
is Fano-coordinate if
\begin{equation}
\left|D(Z_1)\cup\cdots\cup D(Z_7)\right|=7 .
\end{equation}
Finite enumeration gives \(315\) cubes, \(63\) cube-edge directions, \(135\) Fano-coordinate cube decompositions, and \(960\) partitions of the \(63\) directions into nine Fano-coordinate \(7\)-sets.

Thus the working dictionary is
\[
\begin{array}{c|c|c|c|c}
\# 
& W(5,2) 
& E_7 
& C_{56} 
& Q^+(7,2) 
\\ \hline
28
& \text{---}
& \text{---}
& \text{dual-root lines}
& \text{---}
\\
63
& \text{points}
& \text{root lines}
& \text{cube-edge directions}
& \text{---}
\\
315
& \text{lines}
& 3\text{ mutually orthogonal root lines}
& 3\text{-cubes}
& \text{---}
\\
135
& \text{generators}
& X_7\text{ subconfigurations}
& \text{Fano-coord. cube dec.}
& \text{points}
\\
960
& \text{spreads}
& 9X_7\text{ decompositions}
& 9\text{-fold direction spreads}
& \text{ovoids}
\end{array}
\]

\subsection{Construction of \texorpdfstring{\(\texttt{14d252k2}\)}{14d252k2}}

Let \(A=C_{56}\subset\mathbb R^7\) be the dual-root code above. Let
\begin{equation}
\widehat R=\{r/\sqrt2:r\in R(E_7)\}\subset S^6
\end{equation}
be the normalized \(E_7\) root system, and choose a spread decomposition
\begin{equation}
\widehat R=X_1\sqcup X_2\sqcup\cdots\sqcup X_9
\end{equation}
into nine copies of \(X_7\). We use seven blocks, indexed by the coordinates of the first copy of \(\mathbb R^7\). Define
\begin{equation}
C_0=\{(a,0):a\in A\}\subset\mathbb R^7\oplus\mathbb R^7
\end{equation}
and, for \(1\le i\le7\),
\begin{equation}
C_i=
\left\{
\left({\frac{s}{\sqrt{3}}}e_i,\sqrt{\frac{2}{3}}\,x\right):
s=\pm1,\ x\in X_i
\right\}.
\end{equation}
Set \(C=C_0\cup C_1\cup\cdots\cup C_7\).

\begin{proposition}\label{prop:cohn-14d252}
The set \(C\) is an antipodal \(\texttt{14d252k2}\) configuration.
\end{proposition}

\begin{proof}
The pieces are disjoint, so
\begin{equation}
|C|=56+7\cdot2\cdot14=252.
\end{equation}
They are antipodal, and every point has norm \(1\). Inside \(C_0\), the inner products are those of \(C_{56}\). If \((a,0)\in C_0\) and
\begin{equation}
\left({\frac{s}{\sqrt{3}}}e_i,\sqrt{\frac{2}{3}}\,x\right)\in C_i,
\end{equation}
then the inner product is \(sa_i/\sqrt3\in\{0,\pm1/3\}\). For two distinct points in the same \(C_i\),
\begin{equation}
{\frac{1}{3}}ss'+{\frac{2}{3}}\langle x,y\rangle\in\{-1,-1/3,1/3\},
\end{equation}
because \(X_i\) is a cross polytope. For \(i\ne j\), the first coordinates are orthogonal, and distinct spread blocks contain distinct \(E_7\) root lines; hence \(\langle x,y\rangle\in\{0,\pm1/2\}\), and the resulting inner product is \(0\) or \(\pm1/3\). Thus every off-diagonal inner product is at most \(1/3\).
\end{proof}


\textit{PackingStar} experiments motivate the following restricted completeness problem: classify antipodal \(\texttt{14d252k2}\) codes whose off-diagonal cosine set is contained in \(\{-1,0,\pm1/3\}\). The data suggest that the construction above may be exhaustive in this restricted class, but this is not proved and is not used below.

We now record the finite ambiguity in Cohn's construction. This is a classification of the finite construction domain, not a classification of arbitrary antipodal \(\texttt{14d252k2}\) codes. Fix a spread
\begin{equation}
\mathcal S=\{X_1,\ldots,X_9\}.
\end{equation}
The \(960\) spreads form one orbit under \(\operatorname{Aut}(W(5,2))=\operatorname{Sp}(6,2)\). The stabilizer of \(\mathcal S\) induces a group of order \(1512\) on the nine blocks; the action is sharply \(3\)-transitive up to the field automorphism of \(\mathbb F_8\), so we identify it with \(\mathrm P\Gamma\mathrm L(2,8)\). 

The induced coordinate-permutation group of \(A=C_{56}\) is
\begin{equation}
\operatorname{Aut}(\operatorname{PG}(2,2))\cong\operatorname{GL}(3,2),
\end{equation}
of order \(168\). A Cohn parameter is an injection
\begin{equation}
\iota:\Omega\hookrightarrow\mathcal S,
\end{equation}
where \(\Omega\) is the set of seven Fano coordinates; the coordinate \(p\) uses the block \(\iota(p)\). The two blocks outside \(\iota(\Omega)\) are unused, and
\begin{equation}
|\operatorname{Inj}(\Omega,\mathcal S)|=9\cdot8\cdot7\cdot6\cdot5\cdot4\cdot3=181440.
\end{equation}
The evident relabellings give the double quotient
\begin{equation}
\mathrm P\Gamma\mathrm L(2,8)\backslash
\operatorname{Inj}(\Omega,\mathcal S)/
\operatorname{GL}(3,2).
\end{equation}
The finite check also verifies that any seven blocks of the fixed spread determine the two unused blocks: the only generators disjoint from the seven used blocks are the two complementary spread blocks. Thus, within the Cohn construction domain, no additional symplectic relabelling is lost by fixing \(\mathcal S\) and taking the double quotient above.

Enumeration of this quotient gives three orbits. In the same order as the representative list below, the corresponding automorphism group orders are \(5376,768,768\), and the point-orbit counts are \(4,20,20\). The automorphism orders include the normal sign kernel of order \(2^8\): seven independent sign changes in the first coordinate factor and the central sign on the \(E_7\) factor. A compact summary is
\[
\begin{array}{c|c|c|c|c|c}
\text{type}
& \iota
& |\mathcal O_\iota|
& \text{full-ordering count}
& |\operatorname{Aut}(C_\iota)|
& \#\operatorname{Aut}(C_\iota)\text{-orbits on }C_\iota
\\ \hline
1&(0,1,2,3,5,6,7)&12096&16&5376&4\\
2&(0,1,2,3,4,5,6)&84672&112&768&20\\
3&(0,1,2,3,4,5,8)&84672&112&768&20
\end{array}
\]
Here \(|\mathcal O_\iota|\) is the orbit size in \(\operatorname{Inj}(\Omega,\mathcal S)\) under \(\mathrm P\Gamma\mathrm L(2,8)\times\operatorname{GL}(3,2)\). Equivalently, full orderings of the nine blocks have
\begin{equation}
\frac{9!}{|\mathrm P\Gamma\mathrm L(2,8)|}=240
\end{equation}
orbits under the spread stabilizer. The map from a full ordering to a Cohn parameter forgets the order of the two unused blocks. These \(240\) ordered-spread orbits lie over the three isometry types as \(16,112,112\).

Representative coordinates for the three non-isometric antipodal configurations are included in the repository file
\(\texttt{S03\_(3, 252, 14)\_three 14d252k2.npz}\).

\textit{PackingStar} also found non-antipodal 14d252k2 codes with cosine set \(\{-1, -\frac{2}{3}, \pm \frac{1}{3}, 0\}\). They do not come from this construction but still contain a clear \(\texttt{7d56k2}\) subconfiguration, we provided them in the repository but leave that out here.

\subsection{\texorpdfstring{The \(\texttt{13d1146k1}\) configuration}{The 13d1146k1 configuration}}
\label{sec:13d1146k1}

\textit{PackingStar} found a kissing number configuration $C\subset S^{12}, |C|=1146$, whose Gram matrix has cosine set $\{-1,\pm\tfrac14,\pm\tfrac12,0\}$, with nonzero Gram spectrum
\begin{equation}
\operatorname{Spec}_{+}(G_C)=\{84^7,93^6\}.
\end{equation}
The first statistic is
\begin{equation}\label{eq:first-stat-13d1146k1}
\mathsf{Stat}_1(C)=
\left\{\!\left\{
\begin{aligned}
&(0^{432},(\pm\tfrac14)^{256},(\pm\tfrac12)^{100},(\pm1)^1)^{84},\\
&(0^{420},(\pm\tfrac14)^{252},(\pm\tfrac12)^{110},(\pm1)^1)^{896},\\
&(0^{456},(\pm\tfrac14)^{228},(\pm\tfrac12)^{116},(\pm1)^1)^{112},\\
&(0^{420},(\pm\tfrac14)^{240},(\pm\tfrac12)^{122},(\pm1)^1)^{54}.
\end{aligned}
\right\}\!\right\}.
\end{equation}
Thus \(C\) does not improve the known size \(1152\) in dimension \(13\),
but it improves the previous construction with all rational cosines of size \(1130\). More
importantly, it has fundamentally different structure.

The first-statistic classes of sizes \(84\) and \(54\) are lower-dimensional
blocks of dimensions \(7\) and \(6\). Removing them gives a \(1008\)-point
configuration $C_0$ of rank \(13\), with $\operatorname{Spec}_{+}(G_{C_0})=\{72^7,84^6\}$
and single first statistic
\begin{equation}
\mathsf{Stat}_1(C_0)=
\left\{\!\left\{
(0^{378},(\pm\tfrac14)^{216},(\pm\tfrac12)^{98},(\pm1)^1)^{1008}
\right\}\!\right\}.
\end{equation}

After an orthogonal change of coordinates, \(C_0\) lies in
\(\mathbb R^6\oplus\mathbb R^7\) with equal squared norm \(1/2\) in the two
summands. Its \(6\)-dimensional projection is the normalized \(E_6\) root
configuration of size \(72\), each point repeated \(14\) times. Its
\(7\)-dimensional projection is the normalized \(E_7\) root configuration of
size \(126\), each point repeated \(8\) times.

By analyzing the bi-partite construction that merge two kissing number configurations of dimension $d$ and $d'$ into a kissing number configuration of dimension $d+d'$, it motivates the following construction.

\subsection{Fiber tensor product}
\label{sec:fiber-tensor-product}
Let
\begin{equation}
S=X_1\sqcup\cdots\sqcup X_k\subset S^{d-1},
\qquad
S'=X'_1\sqcup\cdots\sqcup X'_k\subset S^{d'-1}
\end{equation}
be kissing number configurations. Assume that every fiber block has internal cosine
at most \(0\):
\begin{equation}
\langle x,y\rangle\le 0
\quad(x\ne y,\ x,y\in X_i),
\qquad
\langle x',y'\rangle\le 0
\quad(x'\ne y',\ x',y'\in X'_i).
\end{equation}
Let \(T\subset S^{d-1}\) and \(T'\subset S^{d'-1}\) be kissing number configurations
such that
\begin{equation}\label{eq:fiber-cap-condition}
\langle s,t\rangle\le {\frac{1}{\sqrt{2}}}
\quad(s\in S,\ t\in T),
\qquad
\langle s',t'\rangle\le {\frac{1}{\sqrt{2}}}
\quad(s'\in S',\ t'\in T').
\end{equation}
Define
\begin{equation}\label{eq:fiber-tensor-product-definition}
(S,T)\otimes_{\mathrm{fib}}(S',T')
=
\bigcup_{i=1}^k
\left\{
\frac1{\sqrt2}(x,x'):
 x\in X_i,
 x'\in X'_i
\right\}
\cup \{(t,0):t\in T\}
\cup \{(0,t'):t'\in T'\}.
\end{equation}

\begin{proposition}\label{prop:fiber-tensor-product}
The set \((S,T)\otimes_{\mathrm{fib}}(S',T')\) is a kissing number configuration in dimension
\(d+d'\). Its size is
\begin{equation}
\sum_{i=1}^k |X_i|\,|X'_i|+|T|+|T'|.
\end{equation}
\end{proposition}

\begin{proof}
All points have norm \(1\). For two product points the inner product is
\begin{equation}
{\frac{\langle x,y\rangle+\langle x',y'\rangle}{2}}.
\end{equation}
If both coordinates change, this is at most
\((1/2+1/2)/2=1/2\), since \(S\) and \(S'\) are kissing number configurations. If
only one coordinate changes inside one fiber, then the other coordinate
contributes \(1\), while the changed fiber contributes at most \(0\); hence the
inner product is at most \((1+0)/2=1/2\). A product point and a cap point have
inner product at most
\begin{equation}
{\frac{1}{\sqrt{2}}}\cdot {\frac{1}{\sqrt{2}}}={\frac{1}{2}}
\end{equation}
by \eqref{eq:fiber-cap-condition}. Points from the two caps are orthogonal to
each other, and each cap is a kissing number configuration by assumption.
\end{proof}
The configuration \(\texttt{13d1146k1}\) is the case
\begin{equation}
S=\texttt{6d72k1}=9X_4,
\qquad
T=\texttt{6d54k1},
\qquad
S'=\texttt{7d126k1}=9X_7,
\qquad
T'=\texttt{7d84k1}.
\end{equation}
Thus
\begin{equation}
|(S,T)\otimes_{\mathrm{fib}}(S',T')|=9\cdot 8\cdot 14+54+84=1146.
\end{equation}
Here \(T\) is the dual \(6\)-dimensional configuration, and \(T'\) is the
\(D_7\) root configuration. The needed cap inequalities are
\begin{equation}
\langle S,T\rangle\le {\frac{1}{\sqrt{2}}},
\qquad
\langle S',T'\rangle\le {\frac{1}{\sqrt{2}}}.
\end{equation}

We did not found a larger $S'$ and $T'$ that are coherent with $S,T$, but they do have different choices of the same size, and could be non-antipodal, which lead to non-isomorphic \(\texttt{13d1146k1}\) that has no more than 138 non-antipodal spheres with cosine \(-3/4\).

\textit{PackingStar} also found a "greatly non-antipodal variant" with same cosine set and has no more than 138 antipodal spheres. In this construction one replaces the \(6\)-dimensional fiber
\(9X_4\) by
\begin{equation}
\texttt{6d72k1}=8Y_3,
\end{equation}
where \(Y\) is the equilateral triangle in dimension \(2\), and \(Y_3\cong Y\perp Y\perp Y\) has size \(9\) and internal cosine at most \(0\). On
the \(7\)-dimensional side one uses a subconfiguration
\begin{equation}
\texttt{7d112k1}=8X_7\subset \texttt{7d126k1}.
\end{equation}
The same caps \(T\) and \(T'\) give
\begin{equation}
8\cdot 9\cdot 14+54+84=1146.
\end{equation}
The \(X\)- and \(Y\)-decompositions of \(\texttt{6d72k1}\) are described in the
next subsection.

It is worth mentioning that the fiber tensor product is not complete for the bi-partite construction, that not all the non-isomorphic bi-partite constructions of the same size could be constructed via fiber tensor product.

\paragraph{Record lower-bound examples from the same product.}
The same naive fiber tensor product reproduces record lower-bound
configurations in dimensions \(6,8,12,14,16\):
\begin{equation}
S=\texttt{2d6k1}=\bigsqcup_{i=1}^{3}\{\pm\omega^i\},
\quad T=\emptyset,
\quad S'=\texttt{4d24k1}=3X_4,
\quad T'=\texttt{4d24k1}
\quad\leadsto\quad 3\cdot 2\cdot 8+24=72,
\end{equation}
\begin{equation}
S=S'=\texttt{4d24k1}=3X_4,
\quad T=T'=S
\quad\leadsto\quad 3\cdot 8\cdot 8+2\cdot24=240,
\end{equation}
\begin{equation}
S=S'=\texttt{6d60k1}=5X_6,
\quad T=T'=S
\quad\leadsto\quad 5\cdot12\cdot12+2\cdot60=840,
\end{equation}
\begin{equation}
S=S'=\texttt{7d126k1}=9X_7,
\quad T=T'=\texttt{7d84k1}
\quad\leadsto\quad 9\cdot14\cdot14+2\cdot84=1932,
\end{equation}
\begin{equation}
S=S'=\texttt{8d240k1}=15X_8,
\quad T=T'=S
\quad\leadsto\quad 15\cdot16\cdot16+2\cdot240=4320.
\end{equation}

\subsubsection{\texorpdfstring{Two decompositions of \(\texttt{6d72k1}\)}{Two decompositions of 6d72k1}}
\label{subsec:two-decompositions-6d72}

We use two decompositions of the normalized \(E_6\) root configuration.  The
first is forced by the \(W(5,2)\) dictionary; the second is an Eisenstein
model decomposition into eight copies of a three-triangle configuration.

\paragraph{The decomposition \(\texttt{6d72k1}=9X_4\).}
Let \(R=R(E_7)\), and identify the root lines \(R/\{\pm1\}\) with the points
of \(W(5,2)\), as in the preceding \(W(5,2)\) section.  Let
\(R_6\subset R\) be an \(E_6\) root subsystem.  Then the \(36\) root lines of
\(R_6\) are the anisotropic points of an elliptic quadratic refinement
\begin{equation}
q:V\longrightarrow \mathbb F_2,
\qquad
q(x+y)+q(x)+q(y)=b(x,y),
\end{equation}
where \(b\) is the symplectic form defining \(W(5,2)\).  For example, in the
standard eight-coordinate model for \(E_7\), the subsystem
\begin{equation}
R_6=\{r\in R(E_7): r_7+r_8=0\}
\end{equation}
has this form.  The general case follows by conjugacy of \(E_6\) subsystems
inside the Weyl group of \(E_7\).

Now let \(P\) be a generator of \(W(5,2)\), i.e. a three-dimensional totally
isotropic subspace of \(V\).  Since \(b|_P=0\), the restriction \(q|_P\) is a
linear functional on \(P\).  It is not the zero functional: if \(q\) vanished on
\(P\), then the elliptic six-dimensional quadratic space \((V,q)\) would have a
three-dimensional totally singular subspace, contradicting its Witt index
\(2\).  Hence \(q|_P\) is a nonzero linear functional, and so exactly four
vectors of \(P\) satisfy \(q=1\).

Therefore every \(X_7\)-block coming from a generator \(P\) meets \(R_6\) in
four mutually orthogonal root lines, i.e. in eight roots forming a copy of the
four-dimensional cross polytope \(X_4\).  Consequently every spread
\begin{equation}
R(E_7)=X(P_1)\sqcup\cdots\sqcup X(P_9)
\end{equation}
induces a decomposition
\begin{equation}
R(E_6)=\bigl(R_6\cap X(P_1)\bigr)\sqcup\cdots\sqcup
\bigl(R_6\cap X(P_9)\bigr)=9X_4.
\end{equation}
Thus the statement is correct for every \(E_6\) root subsystem of the fixed
\(E_7\) root system and every \(9X_7\) spread decomposition.

\paragraph{The decomposition \(\texttt{6d72k1}=8Y_3\).}
Let \(\omega=e^{2\pi i/3}\), let \(\theta=\omega-\omega^2=i\sqrt3\), and
identify \(\mathbb C^3\) with \(\mathbb R^6\).  We normalize all displayed
vectors by \(1/\sqrt3\).  For \(\varepsilon\in\{\pm1\}\), define
\begin{equation}
B_\infty^\varepsilon
=
\left\{
{\frac{\varepsilon\theta\omega^a}{\sqrt{3}}}e_j:
 a\in\mathbb F_3,
 j=1,2,3
\right\},
\end{equation}
and, for \(s\in\mathbb F_3\), define
\begin{equation}
B_s^\varepsilon
=
\left\{
{\frac{\varepsilon}{\sqrt{3}}}(\omega^a,\omega^b,\omega^c):
 a+b+c=s
\right\}.
\end{equation}
Then
\begin{equation}
R(E_6)=
B_\infty^+\sqcup B_\infty^-\sqcup
\bigl(B_0^+\sqcup B_0^-\sqcup B_1^+\sqcup B_1^-\sqcup B_2^+\sqcup B_2^-\bigr)
\end{equation}
is the standard Eisenstein realization of the normalized \(E_6\) roots, written
as a disjoint union of eight blocks of size \(9\).

Each block is isometric to
\begin{equation}
Y_3=Y\perp Y\perp Y,
\qquad
Y=\{1,\omega,\omega^2\}\subset S^1,
\end{equation}
where \(\perp\) denotes disjoint union in mutually orthogonal real two-planes.
For \(B_\infty^\varepsilon\), this is immediate from the three coordinate
complex lines.  For \(B_s^\varepsilon\), the affine plane
\(a+b+c=s\) in \(\mathbb F_3^3\) is partitioned by the diagonal subgroup
\(\langle(1,1,1)\rangle\) into three cosets.  Each coset gives an equilateral
triangle, while two distinct cosets are orthogonal because
\(1+\omega+\omega^2=0\).  Hence each block has internal cosines only
\(-1/2\) and \(0\).

A finite enumeration inside \(R(E_6)\) gives \(240\) equilateral triangles,
\(320\) subconfigurations isometric to \(Y_3\), and \(17920\) exact covers of
\(R(E_6)\) by eight \(Y_3\)-blocks.  The above Eisenstein construction is one
such cover.

\subsection{\texorpdfstring{The \(\texttt{17d578k2}\) configurations}{The 17d578k2 configurations}}
\label{sec:17d578k2}

Let \(\mathcal N\subset \mathbb F_2^{16}\) be the Nordstrom--Robinson code, with
\(|\mathcal N|=256\) and minimum Hamming distance \(6\).  We write
\(N\!R_{16}\subset S^{15}\) for its spherical realization
\begin{equation}
N\!R_{16}
=
\{\pm e_i:1\le i\le 16\}
\cup
\left\{
\frac{1}{4}\bigl((-1)^{c_1},\ldots,(-1)^{c_{16}}\bigr):
(c_1,\ldots,c_{16})\in\mathcal N
\right\}.
\end{equation}
Thus \(|N\!R_{16}|=288\), and its off-diagonal inner products are contained in
\(\{-1,0,\pm\tfrac14\}\).  This is the \(16\)-dimensional Kerdock, or
Nordstrom--Robinson, spherical code.  Cohn, de Laat, and Leijenhorst proved by
exact three-point semidefinite programming bounds that it is the universally optimal and unique (up to isometry) spherical code $\texttt{16d288k3}$
\cite{CohnDeLaatLeijenhorst2024}.  The underlying binary code was
introduced by Nordstrom and Robinson \cite{NordstromRobinson}.

Let \(A,B\subset S^{15}\).  Define the double lift
\begin{equation}\label{eq:17d578-double-lift}
\mathcal L(A,B)=
\{(0^{16},1),(0^{16},-1)\}
\cup
\left\{\left(\sqrt{\frac{8}{9}}\,a,{\frac{1}{3}}\right):a\in A\right\}
\cup
\left\{\left(\sqrt{\frac{8}{9}}\,b,-{\frac{1}{3}}\right):b\in B\right\}.
\end{equation}
For two points in the same layer the inner product is
\begin{equation}
{\frac{8}{9}}\langle a,a'\rangle+{\frac{1}{9}},
\end{equation}
and for points in opposite layers it is
\begin{equation}
{\frac{8}{9}}\langle a,b\rangle-{\frac{1}{9}}.
\end{equation}
Consequently \(\mathcal L(A,B)\) is a \(\texttt{17d578k2}\) configuration whenever
\(A\) and \(B\) are copies of \(N\!R_{16}\) and
\begin{equation}\label{eq:two-NR-kissing-condition}
\langle a,b\rangle\le {\frac{1}{2}}
\qquad(a\in A,
 b\in B).
\end{equation}
Equivalently, \(A\cup B\) is a \(16\)-dimensional kissing number configuration.  Conversely,
every configuration of the form \eqref{eq:17d578-double-lift} satisfying the
\(\texttt{k2}\) bound is obtained from two \(16\)-dimensional codes satisfying
\eqref{eq:two-NR-kissing-condition}.

\textit{PackingStar} found four non-isomorphic examples of this form.

In the repository file $\texttt{S06\_(4, 578, 17)\_17d578k2.npz}$ the coordinates are ordered as $288+288+2$
namely the two affine layers at heights \(\pm1/3\), followed by the two axis
points \((0^{16},\pm1)\).  We denote the four Gram matrices by
\(M_0,M_1,M_2,M_3\).  They have the same positive Gram spectrum,
\begin{equation}\label{eq:17d578-spectrum}
\operatorname{Spec}_{+}(M_i)=\{32^{16},66^1\},
\end{equation}
and the same first statistic:
\begin{equation}\label{eq:first-stat-17d578k2}
\mathsf{Stat}_1(M_i)=
\left\{\!\left\{
\begin{aligned}
&((-1)^1,(-\tfrac13)^{288},(\tfrac13)^{288},1^1)^2,\\
&((-\tfrac79)^1,(-\tfrac59)^{20},(-\tfrac13)^{65},(-\tfrac19)^{248},
(\tfrac19)^{94},(\tfrac13)^{149},1^1)^{576}.
\end{aligned}
\right\}\!\right\}.
\end{equation}
Here, as before, the superscript outside a row is the number of rows of that type.

The four examples are distinguished by their automorphism-orbit decompositions.
For an orbit \(\mathcal O\), let
\(\mathsf{Stat}_1(\mathcal O)\) denote the first statistic of the sub-Gram matrix
indexed by \(\mathcal O\), and let \(\operatorname{Spec}_+(\mathcal O)\) denote the
positive spectrum of that sub-Gram matrix.  The exact orbit data are as follows.

\[
\begin{array}{c|c|c|c}
\text{code} & |\mathcal O| & \mathsf{Stat}_1(\mathcal O) & \operatorname{Spec}_+(\mathcal O)\\
\hline
M_0
&576&
((-\tfrac79)^1,(-\tfrac59)^{20},(-\tfrac13)^{64},(-\tfrac19)^{248},(\tfrac19)^{94},(\tfrac13)^{148},1^1)^{576}
&\{32^{16},64^1\}\\
&2&((-1)^1,1^1)^2&\{2^1\}\\
\hline
M_1
&256&
((-\tfrac79)^1,(-\tfrac59)^8,(-\tfrac13)^{32},(-\tfrac19)^{96},(\tfrac19)^{62},(\tfrac13)^{56},1^1)^{256}
&\{(\tfrac{128}{9})^{16},(\tfrac{256}{9})^1\}\\
&256&
((-\tfrac79)^1,(-\tfrac59)^{12},(-\tfrac13)^{16},(-\tfrac19)^{120},(\tfrac19)^{46},(\tfrac13)^{60},1^1)^{256}
&\{(\tfrac{128}{9})^{16},(\tfrac{256}{9})^1\}\\
&64&
((-\tfrac79)^1,(-\tfrac59)^4,(-\tfrac19)^{24},(\tfrac19)^{30},(\tfrac13)^4,1^1)^{64}
&\{(\tfrac{32}{9})^{16},(\tfrac{64}{9})^1\}\\
&2&((-1)^1,1^1)^2&\{2^1\}\\
\hline
M_2
&384&
((-\tfrac79)^1,(-\tfrac59)^{12},(-\tfrac13)^{48},(-\tfrac19)^{152},(\tfrac19)^{78},(\tfrac13)^{92},1^1)^{384}
&\{(\tfrac{64}{3})^{16},(\tfrac{128}{3})^1\}\\
&96&
((-\tfrac79)^1,(-\tfrac59)^4,(-\tfrac13)^{16},(-\tfrac19)^{24},(\tfrac19)^{30},(\tfrac13)^{20},1^1)^{96}
&\{(\tfrac{64}{9})^{12},(\tfrac{32}{3})^1\}\\
&96&
((-\tfrac79)^1,(-\tfrac59)^4,(-\tfrac13)^{16},(-\tfrac19)^{24},(\tfrac19)^{30},(\tfrac13)^{20},1^1)^{96}
&\{(\tfrac{32}{9})^{12},(\tfrac{32}{3})^5\}\\
&2&((-1)^1,1^1)^2&\{2^1\}\\
\hline
M_3
&512&
((-\tfrac79)^1,(-\tfrac59)^{18},(-\tfrac13)^{56},(-\tfrac19)^{220},(\tfrac19)^{86},(\tfrac13)^{130},1^1)^{512}
&\{(\tfrac{256}{9})^{16},(\tfrac{512}{9})^1\}\\
&32&
((-\tfrac79)^1,(-\tfrac19)^{16},(\tfrac19)^{14},1^1)^{32}
&\{(\tfrac{16}{9})^{16},(\tfrac{32}{9})^1\}\\
&32&
((-\tfrac79)^1,(-\tfrac19)^{16},(\tfrac19)^{14},1^1)^{32}
&\{(\tfrac{16}{9})^{16},(\tfrac{32}{9})^1\}\\
&2&((-1)^1,1^1)^2&\{2^1\}
\end{array}
\]

Thus \(M_0\) has only two point-orbits.  We call this two-orbit example the
\emph{Star}.
Projecting the two non-axis layers to \(\mathbb R^{16}\) and normalizing gives
\begin{equation}
A_i\cup B_i\subset S^{15},
\qquad |A_i|=|B_i|=288,
\end{equation}
with all cross inner products at most \(1/2\).  Hence each projection is a
\(16\)-dimensional kissing number configuration of size \(576\).  For the Star, this
projected configuration embeds into the normalized Barnes--Wall
configuration \(BW:\texttt{16d4320k1}\), which is the minimal vectors of Barnes--Wall lattice $\Lambda_{16}$.  It is organized as two disjoint copies of
\(N\!R_{16}\) inside the \(BW\).  Conversely, any two disjoint copies
\(A,A'\cong N\!R_{16}\) whose union is a kissing number configuration give a
\(\texttt{17d578k2}\) by \eqref{eq:17d578-double-lift}.

The \(BW\) is one of the best known kissing number configuration in
\(16\) dimensions \cite{CohnKissingNumbers}, and it seems to be the most symmetrical one ("Star") of $\texttt{16d4320k1}$,  .  The two Stars therefore suggests the following
finite-geometric question:
can the \(4320\) Barnes--Wall minimal vectors be decomposed into fifteen disjoint
copies of \(N\!R_{16}\)?


\subsection{\texorpdfstring{No decomposition of \(BW\) into \(15\,N\!R_{16}\)}{No decomposition of B into 15 NR16}}
\label{sec:bw-no-15-NR}

The off-diagonal inner products of $B$ are:
\begin{equation}
\{-1,0,\pm\tfrac14,\pm\tfrac12\}.
\end{equation}

There is a canonical decomposition of \(BW\) into
\(135\) cross polytopes \(X_{16}\) \cite{CohnJiaoKumarTorquato2011}.  Namely, for a
minimal vector \(v\), the stabilizer of \(v\) has two orbits on the \(1710\) minimal
vectors orthogonal to \(v\), of sizes \(1680\) and \(30\).  The set consisting of
\(v\), \(-v\), and the orbit of size \(30\) is a copy of \(X_{16}\), and these
copies form an automorphism-invariant partition
\begin{equation}\label{eq:BW-canonical-X16}
B=X_1\sqcup\cdots\sqcup X_{135},
\qquad |X_i|=32.
\end{equation}

For two distinct canonical blocks \(X_i,X_j\), there are exactly two possible
relations:
\begin{equation}\label{eq:BW-two-X16-relations}
\langle X_i,X_j\rangle\subset \{\pm\tfrac14\},
\qquad\text{or}\qquad
\langle X_i,X_j\rangle\subset \{0,\pm\tfrac12\}.
\end{equation}
Let \(\Gamma\) be the graph on the \(135\) blocks in which the first relation in
\eqref{eq:BW-two-X16-relations} is adjacency.  The finite verification identifies
\(\Gamma\) with the non-collinearity graph of \(Q^+(7,2)\).  In particular,
\(\Gamma\) is strongly regular with parameters
\begin{equation}\label{eq:Qplus-noncol-srg}
(v,k,\lambda,\mu)=(135,64,28,32),
\end{equation}
and its maximum cliques have size \(9\).  These \(9\)-cliques are precisely the
\(960\) ovoids of \(Q^+(7,2)\), in the \(W(5,2)\) / \(Q^+(7,2)\) dictionary developed above.

\begin{lemma}\label{lem:BW-288-uses-canonical-X16}
Every subset \(C\subset B\) isometric to \(N\!R_{16}\) is a union of nine canonical
blocks from \eqref{eq:BW-canonical-X16}.
\end{lemma}

\begin{proof}
Since \(N\!R_{16}\) is the unique optimal \(288\)-point code in \(\mathbb R^{16}\),
any such \(C\) has the intrinsic Nordstrom--Robinson decomposition into nine
cross polytopes
\begin{equation}
C=Y_1\sqcup\cdots\sqcup Y_9,
\qquad |Y_i|=32,
\end{equation}
with cross inner products \(\{\pm\tfrac14\}\) between distinct \(Y_i\)'s.  Moreover,
for each \(y\in C\), the set \(\{z\in C:\langle y,z\rangle=0\}\cup\{y,-y\}\) is the
unique intrinsic \(X_{16}\)-block through \(y\).

It remains to exclude a non-canonical intrinsic block.  If one existed, then, by
transitivity of the Barnes--Wall automorphism group on minimal vectors, we may fix
an orthogonal pair \(v_0,w_0\in Y_1\) from the \(1680\)-orbit rather than the
canonical \(30\)-orbit.  All points of \(Y_2,\ldots,Y_9\) must then lie in the
\(\pm\tfrac14\)-neighborhood of both \(v_0\) and \(w_0\), which contains \(1024\)
minimal vectors.  The verifier then recursively fixes one point in each further
intrinsic block and applies the same \(\pm\tfrac14\)-neighborhood condition.  Up to
the stabilizer action, the candidate counts are
\[
8\text{ blocks in }1024
\longrightarrow
7\text{ blocks in }384
\longrightarrow
6\text{ blocks in }128
\longrightarrow
5\text{ blocks in }64
\longrightarrow
4\text{ blocks in }0.
\]
The last zero is a contradiction.  Hence every intrinsic \(Y_i\) is one of the
canonical Barnes--Wall blocks \(X_j\).
\end{proof}

By Lemma~\ref{lem:BW-288-uses-canonical-X16}, a copy of \(N\!R_{16}\) inside \(BW\)
is exactly a union of nine canonical \(X_{16}\)-blocks.  By
\eqref{eq:BW-two-X16-relations}, the nine blocks must be pairwise adjacent in
\(\Gamma\).  Therefore
\begin{equation}\label{eq:BW-NR-ovoid-correspondence}
\left\{N\!R_{16}\text{ subcodes of }B\right\}
\quad\longleftrightarrow\quad
\left\{\text{ovoids of }Q^+(7,2)\right\}.
\end{equation}
Under this correspondence, a decomposition of \(BW\) into fifteen copies of
\(N\!R_{16}\) would be a partition of the \(135\) points of \(Q^+(7,2)\) into
fifteen disjoint ovoids.

\begin{theorem}\label{thm:BW-no-15-NR}
The Barnes--Wall minimal-vector configuration \(BW\) cannot be decomposed into
fifteen copies of \(N\!R_{16}\).  The maximum number of pairwise disjoint copies of
\(N\!R_{16}\) contained in \(BW\) is \(12\).  Equivalently, the maximum number of
pairwise disjoint ovoids in \(Q^+(7,2)\) is \(12\).
\end{theorem}

\begin{proof}
The problem is now a finite set-packing problem on the \(960\) ovoids of
\(Q^+(7,2)\).  The verification uses the automorphism group to reduce the search.
The group is transitive on ovoids, so we fix one ovoid \(O_0\).  The stabilizer of
\(O_0\) has two orbits on the ovoids disjoint from \(O_0\), of sizes \(504\) and
\(56\).  Fix one representative \(O_1\) from each of these two orbits.  In both
cases, an exact branch-and-bound search proves that no \(11\) further ovoids can
be chosen disjointly from \(O_0\cup O_1\).  Thus no family of \(13\) disjoint
ovoids exists.

On the other hand, the same search finds families of \(12\) disjoint ovoids.  Hence
the maximum is \(12\).  Via \eqref{eq:BW-NR-ovoid-correspondence}, this proves the
corresponding statement for copies of \(N\!R_{16}\) in the Barnes--Wall shell.
\end{proof}

The residue of a maximum family is also rigid. It will be used in later section for a new method of construction of the universally optimal configuration $\texttt{22d891k3}$.

\begin{proposition}\label{prop:BW-residue-Schlafli}
Let \(O_1,\ldots,O_{12}\) be any twelve pairwise disjoint ovoids of \(Q^+(7,2)\),
and let
\begin{equation}
R=Q^+(7,2)\setminus (O_1\cup\cdots\cup O_{12}).
\end{equation}
Then \(|R|=27\), and the graph induced on \(R\) by non-collinearity is the
Schl\"afli graph, i.e. the strongly regular graph with parameters
\begin{equation}
(27,16,10,8).
\end{equation}
Consequently every maximal decomposition of the Barnes--Wall shell has the form
\begin{equation}
4320=12\cdot 288+27\cdot 32,
\end{equation}
where the residual \(27\) canonical \(X_{16}\)-blocks carry the Schl\"afli graph.
\end{proposition}

\begin{proof}
Again fix \(O_0\), and then fix the second ovoid in one of the two stabilizer-orbits
of sizes \(504\) and \(56\).  For each representative, the branch-and-bound verifier
enumerates all completions to twelve disjoint ovoids; there are \(1760\) completions
in each case.  In every completion, the induced graph on the \(27\)-point residue has
constant degree \(16\), every adjacent pair has \(10\) common neighbors in the
residue, and every non-adjacent pair has \(8\).  These parameters identify the
residue as the Schl\"afli graph.
\end{proof}

For reference, one maximum family in our enumeration is
\begin{equation}
\{0,66,177,294,351,429,538,632,650,788,877,958\}.
\end{equation}
It gives an explicit decomposition
\begin{equation}
B=\left(\bigsqcup_{i=1}^{12} N_i\right)\sqcup
\left(\bigsqcup_{j=1}^{27} X_j\right),
\qquad
N_i\cong N\!R_{16},
\quad X_j\cong X_{16}.
\end{equation}

Finally, the same dictionary gives a useful complementary object.  A generator of
\(Q^+(7,2)\) has \(15\) points.  These points are pairwise collinear, hence form a
coclique in \(\Gamma\).  The corresponding union of \(15\) canonical
\(X_{16}\)-blocks has size
\begin{equation}
15\cdot 32=480,
\end{equation}
and is the \(D_{16}\) root subsystem visible in the Barnes--Wall coordinate model.

\subsection{\texorpdfstring{Kissing numbers in dimensions \(25\)--\(31\)}{Kissing numbers in dimensions 25--31}}
\label{sec:kissing-25-31}

Let \(\Lambda\) be the Leech lattice, and let \(L\subset S^{23}\) be its normalized minimal shell, so \(|L|=196560\).  We use the following form of the Cohn--Jiao--Kumar--Torquato construction, later improved by Kallal--Kan--Wang \cite{CohnJiaoKumarTorquato2011,KallalKanWang2017}.

\begin{proposition}[Leech lifting]
\label{prop:leech-kissing-lift}
Let \(S_1,\ldots,S_m\) be pairwise disjoint subsets of \(L\), each satisfying \(\langle x,x'\rangle\le 1/4\) for distinct \(x,x'\in S_i\).  Let
\begin{equation}
K_d=T_1\sqcup\cdots\sqcup T_m\subset S^{d-1}
\end{equation}
be a kissing number configuration, and assume that every \(T_i\) has internal cosine at most \(-1/2\).  Then there is a kissing number configuration in dimension \(24+d\) of size
\begin{equation}
196560+\sum_{i=1}^m (|T_i|-1)|S_i|.
\end{equation}
\end{proposition}

\begin{proof}
Replace each \(x\in S_i\) by the points \((\sqrt{2/3}\,x,\sqrt{1/3}\,t)\), \(t\in T_i\), and keep every point of \(L\setminus\bigcup_i S_i\) as \((x,0)\).  If two lifted points have the same Leech coordinate, their inner product is at most \(\frac23+\frac13(-\frac12)=\frac12\).  If their Leech coordinates are distinct but lie in the same \(S_i\), then either the \(T_i\)-coordinate is also distinct, giving at most \(\frac23\cdot\frac14+\frac13\cdot\frac12=\frac13\), or it is the same, giving \(\frac23\cdot\frac14+\frac13=\frac12\).  If they lie in different parts, the inner product is at most \(\frac23\cdot\frac12+\frac13\cdot\frac12=\frac12\).  Finally, \((x,0)\) has inner product at most \(\sqrt{2/3}\cdot\frac12<\frac12\) with a lifted point.
\end{proof}

Kallal--Kan--Wang used Leech subsets of size \(488\) \cite{KallalKanWang2017}.  \textit{PackingStar} finds subsets of size \(496>488\).  With the same lifting, the resulting record kissing numbers are \cite{CohnKissingNumbers}
\[
\begin{array}{c|c}
D & \text{size}\\ \hline
25&197056=196560+1\cdot496\\
26&198550=196560+4\cdot496+6\\
27&200044=196560+7\cdot496+12\\
28&204520=196560+16\cdot496+24\\
29&209496=196560+26\cdot496+40\\
30&220440=196560+48\cdot496+72\\
31&238350=196560+84\cdot496+126.
\end{array}
\]
The last summand comes from an extra copy \(K'_d\subset S^{d-1}\).  After Proposition~\ref{prop:leech-kissing-lift}, one may add \((0,u)\), \(u\in K'_d\), provided \(\langle u,t\rangle\le \sqrt3/2\) for all \(t\in K_d\), since the new cross inner product is at most \(\frac1{\sqrt3}\cdot\frac{\sqrt3}{2}=\frac12\).  This is parallel to the cap condition in the fiber tensor product construction, where \(\frac1{\sqrt2}\cdot\frac1{\sqrt2}=\frac12\).

The other \textit{PackingStar} input is a more efficient decomposition of the auxiliary kissing number configurations.  Let \(Y\) be an equilateral triangle, so its internal cosine is \(-1/2\), and let \(X_1\) be an antipodal pair.  In dimension \(5\), the auxiliary \(40\)-point kissing number configuration is decomposed as \(12Y\sqcup2X_1\), giving \(12(3-1)+2(2-1)=26\).  In dimension \(7\), the \(E_7\) root kissing number configuration is decomposed as \(42Y\), giving \(42(3-1)=84\).  We return to the \(7\)-dimensional decomposition below.

\subsubsection{\texorpdfstring{The \(496\)-point Leech subcode}{The 496-point Leech subcode}}
\label{subsec:S08-496-subcode}

We use the standard coordinate model for the Leech minimal vectors, in which the possible coordinate shapes are
\begin{equation}
{\frac{1}{\sqrt{8}}}(4^2,0^{22}),\qquad
{\frac{1}{\sqrt{8}}}(2^8,0^{16}),\qquad
{\frac{1}{\sqrt{8}}}(3^1,1^{23}).
\end{equation}
The verifier stores the integer vectors \(v\) before the factor \(1/\sqrt8\).  Thus \(v\cdot v=32\), and the unit vector is \(v/\sqrt{32}\).  Hence the condition \(\langle v/\sqrt{32},w/\sqrt{32}\rangle\le 1/4\) is the raw dot-product condition \(v\cdot w\le 8\).

The file \texttt{S08\_Leech\_Si\_subcodes.npz} contains six \(496\)-point subsets of the Leech shell.  For each of them the verifier checks \(\max_{x\ne y}x\cdot y=8\), so each is a \(24\mathrm d496\mathrm k3\) configuration.  We separate the six examples by invariants of the positive-contact graph \(x\cdot y=8\), but we do not use a classification of such subcodes.

One example is recoordinated in \texttt{S08\_Leech\_rearranged\_Si.npz}.  In this coordinate system it has the visible decomposition
\begin{equation}
496=128+128+192+48=(8X_8)+(8X_8)+(12X_8)+(12X_2).
\end{equation}
The last \(48\) points form the duadized Conway--Curtis cross layer.  More precisely, let
\begin{equation}
F=\{\pm 2\alpha_i:i\in\Omega\},\qquad |\Omega|=24,
\end{equation}
where \((\alpha_i,\alpha_j)=2\delta_{ij}\).  Thus \(F\) is a Conway cross, or norm-\(8\) frame, in the Leech lattice.  A duadization is a partition \(\Omega=\bigsqcup_{a=1}^{12}\{i_a,j_a\}\).  It gives
\begin{equation}
F_{\rm duad}=\bigcup_{a=1}^{12}
\{\pm(\alpha_{i_a}+\alpha_{j_a}),\,\pm(\alpha_{i_a}-\alpha_{j_a})\},
\end{equation}
which is \(12X_2\).  We use the name Conway--Curtis cross because Conway uses crosses in the characterization of the Leech lattice, and Curtis studies their fixed-cross geometry under involutions of \(Co_0\) \cite[Ch.~12]{ConwaySloaneSPLAG}\cite{CurtisSubgroupsI,CurtisSubgroupsII}.  In our coordinates the duads are \((0,12),(1,13),\ldots,(11,23)\), up to ordering.

The three larger pieces lie in three coordinate \(16\)-spaces.  Their complementary octads are
\begin{equation}
O_A=\{0,\ldots,7\},\qquad
O_B=\{8,\ldots,15\},\qquad
O_C=\{16,\ldots,23\}.
\end{equation}
In the standard Golay-code/MOG terminology, an octad is a block of the Steiner system \(S(5,8,24)\), and a trio is a partition of the \(24\) coordinates into three disjoint octads.  The Miracle Octad Generator of Curtis is the usual device for working with these objects \cite{CurtisMOG,ConwaySloaneSPLAG}.  Thus \(O_A,O_B,O_C\) form a trio.  The \(48\)-point duadized layer splits as three \(16\)-point slices, four duads on each edge of this trio.

The relation with the preceding Barnes--Wall section is as follows.  For an octad \(O\), the Leech minimal vectors with zero coordinates on \(O\) form a \(16\)-dimensional Barnes--Wall shell of size \(4320\).  In the canonical Barnes--Wall decomposition, the \(135\) copies of \(X_{16}\) are the points of \(Q^+(7,2)\), and \(9\)-point ovoids give the \(16\mathrm d288\mathrm k3\) Nordstrom--Robinson subcodes.  Generators of \(Q^+(7,2)\) have \(15\) points; their unions are \(D_{16}\) root subsystems of size \(15\cdot32=480\).  We only note that the same number \(135\) also occurs in Curtis's fixed-cross geometry for \(B\)- and \(D\)-involutions; nothing below depends on identifying these two appearances.

The three \(16\)-dimensional pieces of the \(496\)-set have two local types.  The first two \(128\)-point pieces are isomorphic; we call them twisted pieces.  Each has maximum same-support completion size \(160\).  The \(192\)-point piece is different: its same-support compatible pool has size \(288\), is a \(16\mathrm d288\mathrm k3\) code, and hence is Nordstrom--Robinson type.  Thus the local structure is
\begin{equation}
128_{\rm tw}+128_{\rm tw}+192_{\rm NR\text{-}extendable}+48_{\rm cross}.
\end{equation}

Finally, the octad sign changes and the duadization have different roles.  For an octad \(O\), the sign changes \(\varepsilon_O^F\) and \(\varepsilon_{\Omega\setminus O}^F=-\varepsilon_O^F\) are the two signed lifts of the same projective involution; in one convention the \(16\)-dimensional Barnes--Wall space is the \(+1\)-space, and in the other it is the \(-1\)-space \cite{LamShimakura2008}.  The duadization has two distinguished complementary transversal dodecads compatible with the Leech/Golay structure.  The corresponding dodecad sign changes preserve the \(12X_2\) layer, but not the full \(496\)-point code.

The recoordinated \(496\)-point code has a visible diagonal sign-change symmetry group \(2^6\).  Its trio-octad subgroup has order \(2^3\) and preserves the full Leech shell; its image in \(Co_1\) has order \(4\).  The remaining sign changes are symmetries of this spherical code but do not preserve the whole Leech shell in these coordinates.

\subsubsection{\texorpdfstring{The decomposition \(7\mathrm d126\mathrm k1=42Y\)}{The decomposition 7d126k1=42Y}}
\label{subsec:S08-7d126-42Y}

Let \(Y\) be an equilateral triangle.  \textit{PackingStar} found a decomposition of the normalized \(E_7\) root configuration
\begin{equation}
R(E_7)=Y_1\sqcup\cdots\sqcup Y_{42}.
\end{equation}
We do not yet have a conceptual classification of such decompositions.  Here we record one obstruction showing that this decomposition is not explained by pairing triangles into antipodal hexagons.

We use the notation from the preceding \(W(5,2)\) section: the \(63\) antipodal root lines of \(E_7\) are identified with the points of \(W(5,2)\), with symplectic form \(b\).  A projective line in \(PG(5,2)\) has the form
\begin{equation}
\ell=\{x,y,x+y\}.
\end{equation}
If \(b(x,y)=0\), then the three corresponding root lines are mutually orthogonal, and the six roots form \(X_3\).  If \(b(x,y)=1\), then the six roots form an \(A_2\) hexagon, i.e. \(Y\sqcup(-Y)\).  Thus a decomposition of \(R(E_7)\) into \(21\) antipodal hexagons would be a spread of \(PG(5,2)\) consisting only of non-isotropic lines.

This is impossible.  Let
\begin{equation}
q(x)=x_1x_4+x_2x_5+x_3x_6,
\qquad
q(x+y)+q(x)+q(y)=b(x,y).
\end{equation}
This is the same quadratic-refinement viewpoint used above for \(6\mathrm d72\mathrm k1=9X_4\).  Among the \(63\) nonzero vectors, exactly \(28\) have \(q(x)=1\).  For \(\ell=\{x,y,x+y\}\), the parity of
\begin{equation}
\sum_{z\in\ell}q(z)
\end{equation}
is \(b(x,y)\).  Hence an isotropic line contains an even number of \(q=1\) points, while a non-isotropic line contains an odd number.  In any line spread of \(PG(5,2)\), the \(21\) lines partition the \(63\) points, so the number of non-isotropic lines has the same parity as \(28\), hence is even.  It cannot be \(21\).

The bound is sharp: the verifier gives a spread with \(20\) non-isotropic lines and one isotropic line.  Therefore the best antipodal-hexagon decomposition has type
\begin{equation}
20A_2+X_3,
\end{equation}
not \(21A_2\).

Consequently the \textit{PackingStar} decomposition \(126=42Y\) is genuinely non-antipodal.  In the example stored in the verifier, the \(42\) triangles project to \(42\) non-isotropic lines of \(PG(5,2)\), each point of \(PG(5,2)\) occurs on exactly two of these lines, and no selected triangle is paired with its antipodal triangle.  Thus the projected object is a \(2\)-cover by non-isotropic lines, together with an orientation/sign lift to the \(E_7\) roots.  We leave the geometric explanation of this \(42Y\) cover open.


\section{Regular Yet Asymmetric Constellations}
This section studies constellations, the prescribed-support optimal configuration spaces. We separate optimality from two additional regularity properties: S-LP tightness and \(m\)-cosine \(m\)-design structure. The examples show that these notions need not coincide, revealing constellations that are optimal yet asymmetric, regular yet not LP-tight, or LP-tight for reasons beyond design regularity.

\subsection{Constellations}
\label{app:constellations}

We recall the prescribed-support notation.  For a finite cosine set
\(\mathcal C\subset[-1,1)\), let
\begin{equation}
\Omega(n,\mathcal C)
=
\{X\subset  S^{n-1}:
\langle x,y\rangle\in\mathcal C, \forall x\neq y\in X\}.
\end{equation}
The prescribed-support optimum is
\begin{equation}
K_{\mathcal C}(n)^*
=
\max
\left\{
|X|:
X\in\Omega(n,\mathcal C)
\right\}.
\end{equation}
The corresponding optimal configuration space is
\begin{equation}
\mathfrak O(n,\mathcal C)
=
\{X\in\Omega(n,\mathcal C): |X|=K_{\mathcal C}(n)^*\}/\!\sim ,
\end{equation}
where \(\sim\) means orthogonal equivalency and relabeling.

We shall call this same object a constellation.

\begin{definition}[Constellation]
For a finite cosine set \(\mathcal C\subset[-1,1)\), the constellation with
prescribed support \(\mathcal C\) in \( S^{n-1}\) is just the optimal configuration space:
\begin{equation}
\operatorname{Const}(n,\mathcal C)
:=
\mathfrak O(n,\mathcal C).
\end{equation}
\end{definition}
The two names emphasize two complementary viewpoints on the same object.
The term optimal configuration space is algorithmic and variational: it is the
part of the extremal configuration space that a discrete optimization procedure
such as \textit{PackingStar} seeks to reach.  From this viewpoint,
\(\mathfrak O(n,\mathcal C)\) records the transition from local saturation in
\(\mathfrak E(n,\mathcal C,N)\) to global optimality at the prescribed-support
bound \(K_{\mathcal C}(n)^*\).

The term constellation is mathematical and geometric.  It treats the same object as a static finite pattern selected by a maximal principle. These spaces can be small, highly structured, or unexpectedly branched, sometimes contain a specially symmetrical element that we call it a \emph{Star}. They form the natural
moduli spaces of equality in a finite-distance extremal problem.

The word constellation is intended in three simultaneous senses.  First, it is
literal: classical constellations are finite patterns on the celestial spheres,
whereas the objects studied here are finite patterns on high-dimensional unit
spheres.  Second, it is phenomenological: extremal configuration spaces are
large galaxies of locally saturated patterns, and the constellation is the
globally optimal part selected by the cardinality principle
\(K_{\mathcal C}(n)^*\).  Third, it is historical.  Kepler's sphere-packing
problem and the Newton--Gregory kissing-number problem both emerged from the
same celestial tradition in which arrangements of spherical
bodies became central mathematical objects.

In many of the regular cases below, this analogy becomes mechanical as well.
Let \(X\subset S^{n-1}\) be finite.  As defined by Leech in ~\cite{Leech1957}, we say that \(X\) is balanced, if for every Newtonian force law depending only on distance, the tangential resultant force at every
point vanishes.  Equivalently, for every function
\begin{equation}
w:I(X)\to\mathbb R,
\end{equation}
and every \(x\in X\), one has
\begin{equation}
\sum_{\substack{y\in X\\ y\ne x}}
w(\langle x,y\rangle)
\bigl(y-\langle x,y\rangle x\bigr)
=
0
\qquad\text{in }T_x S^{n-1}.
\end{equation}
Thus every distance-dependent central interaction produces no tangential
motion on the sphere.  This balance condition is not part of the definition of
a constellation; rather, it is an additional regularity phenomenon.  In
particular, when $X$ has \(m\) distinct inner
products beside $\pm1$ and is a spherical \(m\)-design, it would satisfy this force-balance property by Theorem 2.1 of ~\cite{CohnElkiesKumarSchuermann2010}. And this condition is very close to the S-LP tight condition that we will discuss below. This is why many interesting constellations are not merely conditional optimal codes, they are finite-distance
extremizers with a Newtonian equilibrium interpretation.

\subsection{S-LP tight constellations}

We next record two additional properties that a constellation may carry.  They
are not part of the definition of a constellation.  Rather, they distinguish
the constellations whose optimality is certified by linear programming and
whose representatives satisfy a design-theoretic regularity condition.

Let \(Q_i^{(n)}\) denote the normalized Gegenbauer polynomials for
\( S^{n-1}\), with \(Q_i^{(n)}(1)=1\).  For a finite cosine set
\(\mathcal C\subset[-1,1)\), define the Delsarte--Goethals--Seidel
prescribed-support linear programming bound by
\begin{equation}
\operatorname{LP}_{\mathcal C}(n)
=
\inf \frac{f(1)}{f_0},
\end{equation}
where the infimum is over all real polynomials
\begin{equation}
f(u)=f_0+\sum_{i\ge1} f_i Q_i^{(n)}(u)
\end{equation}
satisfying
\begin{equation}
f_0>0,\qquad f_i\ge0\quad(i\ge1), \qquad 
f(a)\le0\quad(a\in\mathcal C).
\end{equation}
Then every \(X\in\Omega(n,\mathcal C)\) satisfies
\begin{equation}
|X|\le \operatorname{LP}_{\mathcal C}(n).
\end{equation}

\begin{definition}[S-LP tight constellation]
A constellation \(\operatorname{Const}(n,\mathcal C)\) is called
S-LP tight if
\begin{equation}
K_{\mathcal C}(n)^*
=
\operatorname{LP}_{\mathcal C}(n).
\end{equation}
\end{definition}

When \(|\mathcal C|=m\), there is another regularity condition that often
appears in the examples.

\begin{definition}[\(m\)-cosine \(m\)-design regularity]
Let \(|\mathcal C|=m\).  A spherical code is called \(m\)-cosine \(m\)-design
regular if $X$ is a spherical \(m\)-design and
\begin{equation}
I(X)=\mathcal C
\end{equation}
and \(X\) is a spherical \(m\)-design.  A component of
\(\operatorname{Const}(n,\mathcal C)\) is called \(m\)-cosine \(m\)-design regular if every representative in that component has $I(X) = \mathcal{C}$ and has this property.
\end{definition}

The regularity condition implies Leech-type force balance as showed in ~\cite{CohnElkiesKumarSchuermann2010}. 

We will often use the following annihilator polynomial:
\begin{equation}
\Phi_{\mathcal C}(u)
=
\prod_{a\in\mathcal C}\frac{u-a}{1-a}
=
\sum_{i=0}^{m}\phi_i Q_i^{(n)}(u),
\qquad m=|\mathcal C|.
\end{equation}
If \(X\) has \(I(X)=\mathcal C\) and is a spherical \(m\)-design, then
\begin{equation}
\phi_0=\frac1{|X|}.
\end{equation}
Indeed,
\begin{equation}
\sum_{x,y\in X}\Phi_{\mathcal C}(\langle x,y\rangle)=|X|,
\end{equation}
because all off-diagonal terms vanish and all diagonal terms equal \(1\);
on the other hand, since \(\deg \Phi_{\mathcal C}=m\) and \(X\) is an
\(m\)-design,
\begin{equation}
\sum_{x,y\in X}\Phi_{\mathcal C}(\langle x,y\rangle)
=
|X|^2\phi_0.
\end{equation}
Hence \(\phi_0=1/|X|\).

Thus, an \(m\)-cosine \(m\)-design is one positivity condition away
from an S-LP certificate: if
\begin{equation}
\phi_i\ge0\qquad(1\le i\le m),
\end{equation}
then \(\Phi_{\mathcal C}\) itself proves the S-LP bound.  Strict positivity is
not required for S-LP tightness.  It only determines which harmonic moments are
forced by equality.  If a tight certificate has \(f_i>0\), then the degree
\(i\) harmonic moment must vanish at equality; if \(f_i=0\), that moment is not
forced by the certificate, although it may vanish for independent structural
reasons.

We keep these two properties separate from the definition of a constellation.
The following examples illustrate the main boundary phenomena.

\subsection{Edge cases of Constellations}

\begin{center}
\scriptsize
\setlength{\tabcolsep}{3.2pt}
\renewcommand{\arraystretch}{1.12}
\resizebox{\textwidth}{!}{%
\begin{tabular}{@{}l c c c l@{}}
\toprule
Example & S-LP tight? & Regular? & In constellation? & Note \\
\midrule
\(\{e_1,\ldots,e_n\}\subset S^{n-1}\)
& yes & no & yes & repeated-root certificate \\

signed simplex, \(I(C)=\{\pm1/n\}\)
& yes & no & yes & support/design incompatibility \\

simplex edge-midpoints \(8d36\)
& no & yes & yes & fractional LP bound below \(37\) \\

\(22d336\), \(I(C)=\{-3/5,-1/5,1/5\}\)
& no & no & yes & SDP-only, rank-dropped optimum \\

\(32d146880\)
& no & yes & no & same support admits larger codes \\

Hall--Janko / \(J_2{:}2\), \(63d280\)
& yes & partial & yes & nonnegative boundary case \\
\bottomrule
\end{tabular}%
}
\par\vspace{0.5em}
\refstepcounter{table}\label{tab:constellation-edge-cases}
\smallskip
{\small\textbf{Table \thetable.}
Boundary cases for S-LP tightness and \(m\)-cosine \(m\)-design regularity.}
\end{center}

\paragraph{The orthonormal basis.}
For \(C=\{e_1,\ldots,e_n\}\), one has \(I(C)=\{0\}\) and
\(K_{\{0\}}(n)^*=n\).  The S-LP certificate is not the degree-one annihilator
\(u\), but the repeated-root polynomial
\begin{equation}
u^2=\frac1nQ_0^{(n)}(u)+\frac{n-1}{n}Q_2^{(n)}(u).
\end{equation}
Thus the constellation is S-LP tight.  However, \(C\) is not a spherical
\(1\)-design, since \(\sum_i e_i\ne0\).  This is the basic case where LP
tightness holds but regularity fails.
\paragraph{The signed simplex.}
Let \(v_1,\ldots,v_{n+1}\subset S^{n-1}\) be a regular simplex, and
consider signed representatives
\begin{equation}
C_\varepsilon=\{\varepsilon_i v_i:1\le i\le n+1\},
\qquad
\varepsilon_i\in\{\pm1\}.
\end{equation}
For the prescribed support
\begin{equation}
\mathcal C=\left\{-\frac1n,\frac1n\right\},
\end{equation}
the annihilator is
\begin{equation}
\Phi_{\mathcal C}(u)
=
\frac{u^2-\frac1{n^2}}{1-\frac1{n^2}}
=
\frac1{n+1}Q_0^{(n)}(u)
+
\frac{n}{n+1}Q_2^{(n)}(u).
\end{equation}
Thus \(K_{\mathcal C}(n)^*\le n+1\), and equality is attained by every signed
simplex.  Hence these configurations lie in the constellation
\(\operatorname{Const}(n,\mathcal C)\), and the constellation is S-LP tight.

However, this example is not \(2\)-cosine \(2\)-design regular for the full
support \(\mathcal C\).  If all signs are equal, the code is the regular
simplex and is a spherical \(2\)-design, but its intrinsic support is only
\begin{equation}
I(C_\varepsilon)=\left\{-\frac1n\right\}\subsetneq\mathcal C.
\end{equation}
If both signs occur, then \(I(C_\varepsilon)=\mathcal C\), but
\begin{equation}
\sum_i\varepsilon_i v_i\ne0,
\end{equation}
since the only linear dependence among the simplex vertices is
\(\sum_i v_i=0\).  Thus the full-support representatives are not even
spherical \(1\)-designs.  This is precisely why the exact-support condition in
the definition of \(m\)-cosine \(m\)-design regularity is needed.

\paragraph{The simplex edge-midpoints \(8d36\).}
Let \(C\subset S^7\) be the normalized edge-midpoints of a regular
\(8\)-simplex.  Then
\begin{equation}
|C|=36,\qquad I(C)=\left\{-\frac27,\frac5{14}\right\},
\end{equation}
and \(C\) is a two-cosine two-design.  Its annihilator has the expansion
\begin{equation}
\Phi_{I(C)}=\frac1{36}-\frac7{81}Q_1^{(8)}+\frac{343}{324}Q_2^{(8)},
\end{equation}
so the constant coefficient is correct but the \(Q_1\)-coefficient is negative.
The finite-support LP value is
\begin{equation}
\operatorname{LP}_{I(C)}(8)=\frac{34200}{929}\approx36.81378.
\end{equation}
Thus the code is not exactly S-LP tight.  Since the LP bound is still below
\(37\), integrality gives \(K_{I(C)}(8)^*=36\).  Hence \(8d36\) is a regular
constellation, but only near-LP-tight.

\paragraph{The rank-dropped \(22d336\) case.}
For
\begin{equation}
\mathcal C=\left\{-\frac35,-\frac15,\frac15\right\},
\end{equation}
the prescribed-support problem in \( S^{21}\) has optimum \(336\), but
this optimum is not certified by the Delsarte LP bound: the LP bound is \(371\).
Instead, the sharp upper bound \(336\) is obtained by SDP methods.  Moreover,
the equality case drops rank: the optimal configuration is the known \(21d336\) code.

\paragraph{The \(32d146880\) shell.}
For the normalized minimal shell of a \(32\)-dimensional extremal even
unimodular lattice,
\begin{equation}
|C|=146880,\qquad I(C)=\left\{-1,0,\pm\frac14,\pm\frac12\right\},
\end{equation}
and \(C\) is a spherical \(7\)-design.  Nevertheless its normalized
annihilator begins
\begin{equation}
\Phi_{I(C)}
=
\frac1{146880}
+\frac1{4590}Q_1^{(32)}
-\frac{403}{164160}Q_2^{(32)}
-\frac{31}{969}Q_3^{(32)}
+\cdots ,
\end{equation}
so the simple annihilator is not LP-feasible.  More decisively, the same
prescribed support admits larger actual codes: the Leech kissing number configuration
embedded in \(\mathbb R^{32}\) already gives \(196560\) points, and the
full-dimensional orthogonal union \(C_{\Lambda_{24}}\cup C_{E_8}\subset
\mathbb R^{24}\oplus\mathbb R^8\) gives \(196800\) points.  Thus \(32d146880\)
is highly regular, but it is not a member of the corresponding constellation.
\paragraph{The Hall--Janko configuration \(63d280\).}
The Hall--Janko configuration gives a \(280\)-point code
\begin{equation}
C\subset S^{62},
\qquad
I(C)=\left\{-\frac19,\frac19\right\}.
\end{equation}
With $Aut = J_2:2$, where $J_2$ is the Hall-Janko sporadic group. Its first stat:
\begin{equation}
\mathsf{Stat}_1(C)=
\left\{\!\left\{
\begin{aligned}
&((-\tfrac19)^{144},(\tfrac19)^{135},(\pm1)^1)^{280}
\end{aligned}
\right\}\!\right\}.
\end{equation}
Its annihilator is
\begin{equation}
\Phi(u)
=
\frac{u^2-\frac1{81}}{1-\frac1{81}}
=
\frac1{280}Q_0^{(63)}(u)
+
\frac{279}{280}Q_2^{(63)}(u).
\end{equation}
Thus the annihilator is Gegenbauer-nonnegative, but the \(Q_1\)-coefficient is
zero.  It follows that the configuration is S-LP tight, but the LP certificate
does not by itself force the degree \(1\) moment. And indeed, flip any vectors in the set would break the first moment.

This is the subtle difference from the signed simplex.  The Hall--Janko code
has full support \(I(C)=\{\pm1/9\}\), and it is a spherical \(2\)-design by its
underlying spectral construction.  Hence it is genuinely \(2\)-cosine
\(2\)-design regular as an object.  The word ``partial'' means only several codes in the constellation is 2-design regular.

These examples explain why S-LP tightness and \(m\)-cosine \(m\)-design
regularity are kept as properties rather than built into the definition of a
constellation.  The orthonormal basis and the signed simplex are S-LP tight but
not design-regular.  The simplex edge-midpoints are regular and optimal but not
exactly LP-tight.  The \(32d146880\) shell is highly regular but not optimal for
its prescribed support.  The Hall--Janko example shows that nonnegative, rather
than strictly positive, Gegenbauer coefficients are the correct LP condition:
zero coefficients may occur naturally, and design regularity may come from
additional structure rather than from LP complementarity alone.

\subsection{Constellations of \textit{PackingStar}'s records}

We have state the Optimality theorem of those codes in constellation \texttt{12d81k3}, \texttt{20d405k3}, \texttt{21d567k3}, \texttt{22d352k4}. The optimality of the first three would be proved in \ref{sec:Fi21-constellation}, and the last one have historically been found as an Equiangular Tight Frame (ETF for short, the definition already implys the S-LP tightness and regularity), though not know as a record $\tfrac{1}{5}$-code before the \textit{PackingStar}'s discovery. The Last two seems to be unique, the first two has different solution, and both have a star element. 

\begin{figure}[p]
    \centering
    \includegraphics[
        width=1.32\textwidth,
        height=0.89\textheight,
        keepaspectratio
    ]{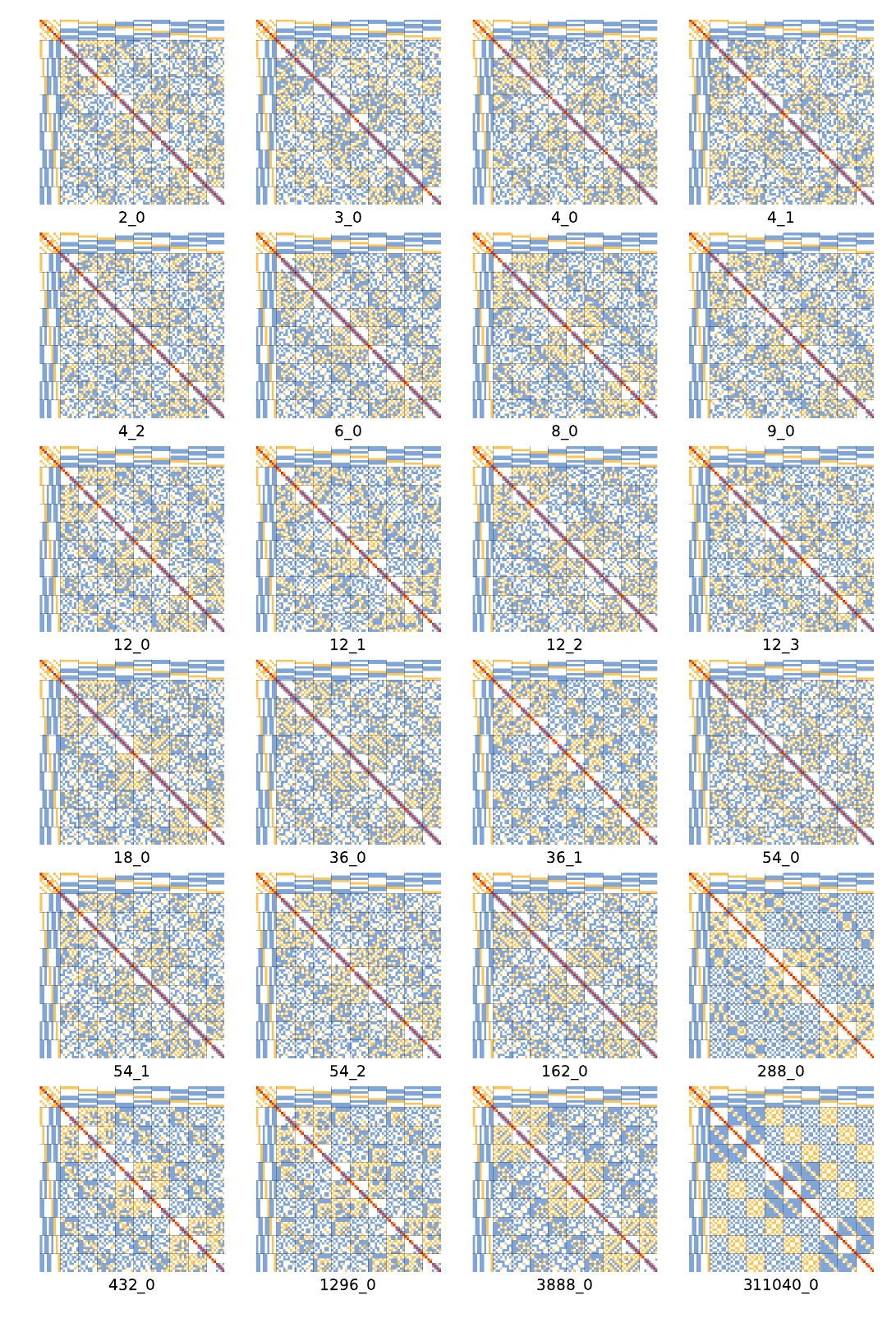}
    \vspace{-0.4em}
    \caption{\fontsize{9}{10}\selectfont
    \emph{Gram matrices in the $\texttt{12d81k3}$ constellation}: the 24 cosine Gram matrices of configurations in the $\texttt{12d81k3}$ constellation of size at least $31$. 
    The vectors are rearranged so that a fixed $Y\otimes Y$ code appears first as a generalized fiber tensor-product head, followed by nine aligned fibers of eight vectors each. 
    The label a\_b below each panel records the automorphism-group order $a=|\operatorname{Aut}(C)|$ and a local tag $b=0,1,2,\ldots$ distinguishing non-isomorphic configurations with the same value of $a$.
    The color scheme is the same as in the preceding figures: white for $1/4$, blue for $-1/8$, yellow for $-1/2$, and red for $1$ on diagonal.
    }
    \label{grams_12d81k3}
\end{figure}

The constellation of \texttt{12d81k3} has at least of size 31, for which the star could be defined as the kronecker tensor of the Schl\"afli code and the equilaterial triangle: $\texttt{6d27k3} \otimes_{kron} Y$, here in the Figure \ref{grams_12d81k3} we show 24 of them.

\subsection{Examples of S-LP-tight regular constellations}
\label{sec:examples-slp-regular-constellations}

We now record examples of constellations that carry both additional structures
introduced above: S-LP tightness and whole space \(m\)-cosine \(m\)-design regularity.
Thus, for a listed code \(C\), the prescribed-support optimum
\begin{equation}
K_{I(C)}(n)^*
\end{equation}
is certified by the Delsarte--Goethals--Seidel linear programming bound, and
\(C\) has intrinsic support \(I(C)\) with \(|I(C)|=m\) and is a spherical
\(m\)-design.

The tables are restricted to the nonclassical part of this phenomenon.  We do
not list universally optimal or sharp configurations, whose regularity is
governed by the stronger Cohn--Kumar theory; see \cite{CohnKumar2007}.  We also omit the standard two-distance sources, but with the following
convention.  A primitive strongly regular graph has two nontrivial spectral
embeddings, and each gives a two-cosine two-design.  However, the standard
degree-two S-LP certificate applies to a given embedding only when its two
inner products \(a,b\) satisfy \(a+b\le0.\)
Equivalently, for an \(\operatorname{srg}(v,k,\lambda,\mu)\) with nontrivial
eigenvalues \(r>s\), the \(r\)-embedding satisfies this condition when
\(v\le2(k-s)\), while the \(s\)-embedding satisfies it when
\(v\ge2(k-r)\).  Thus at least one of the two embeddings is covered by the
standard LP mechanism, but not necessarily both.

For equiangular-line systems, we likewise distinguish the projective line
system from its spherical realizations.  The antipodal lift of an equiangular
tight frame is a three-cosine three-design and is S-LP tight for the support
\(\{-1,\pm\alpha\}\).  A one-sided orientation is S-LP tight for the support
\(\{\pm\alpha\}\), but it is design-regular only when the chosen signing has
zero centroid. The rank-drop cases like \(22d336\) are also not included.

For entries identified from Cohn's spherical-code table, we cite the reference
displayed there; the archived table itself is \cite{cohn2024_spherical_codes}.
The following tables collect the remaining examples, in the range investigated
here, with \(|I(C)|>2\) and not universally optimal.

\begin{center}
\scriptsize
\setlength{\tabcolsep}{2.8pt}
\renewcommand{\arraystretch}{1.08}
\resizebox{\textwidth}{!}{%
\begin{tabular}{@{}l c c c l l c@{}}
\toprule
Code & $s$ & $\tau$ & $|I(C)|$ & $I(C)$ & Reference / construction & Status \\
\midrule
$6d32$        & $1/3$   & $3$ & $3$ & $\{-1,-1/3,1/3\}$ & \cite{van1991equilateral} & \\
$6d72$        & $1/2$   & $5$ & $4$ & $\{-1,-1/2,0,1/2\}$ & \cite{korkine1873formes,leech1969six} & \\
$7d126$       & $1/2$   & $5$ & $4$ & $\{-1,-1/2,0,1/2\}$ & \cite{korkine1873formes,leech1969six} & \\
$9d36$        & $1/5$   & $3$ & $3$ & $\{-3/5,-1/5,1/5\}$ & \cite{1995Spherical} & \\
$12d81$       & $1/4$   & $3$ & $3$ & $\{-1/2,-1/8,1/4\}$ & $\ast$; Fischer Tower & $s+$ \\
$14d64$       & $1/7$   & $3$ & $3$ & $\{-3/7,-1/7,1/7\}$ & \cite{1995Spherical} & \\
$15d128$      & $1/5$   & $3$ & $3$ & $\{-1/3,-1/15,1/5\}$ & \cite{1995Spherical} & \\
$15d135$      & $1/4$   & $3$ & $3$ & $\{-1/2,-1/8,1/4\}$ & $\ast$; Fischer Tower & $s+$ \\
$16d512$      & $1/3$   & $5$ & $4$ & $\{-1,-1/3,0,1/3\}$ & \cite{1995Spherical} & $s$ \\
$16d4320$     & $1/2$   & $7$ & $6$ & $\{-1,-1/2,-1/4,0,1/4,1/2\}$ & \cite{barnes1959some,BoyvalenkovCherkashinDragnev2025} & $s$ \\
$18d243$      & $1/4$   & $3$ & $3$ & $\{-1/2,-1/8,1/4\}$ & $\ast$; Fischer Tower & $s+$ \\
$18d1458$      & $2/5$  & $4$ & $4$ & $\{-1/2,-1/5,1/10,2/5\}$ & Genealogy $23d11178$ & $+$ \\
$20d256$      & $1/5$   & $3$ & $3$ & $\{-3/5,-1/5,1/5\}$ & \cite{1995Spherical} & \\
$20d405$      & $1/4$   & $3$ & $3$ & $\{-1/2,-1/8,1/4\}$ & $\ast$; Fischer Tower & $s$ \\
$21d252$      & $1/5$   & $3$ & $3$ & $\{-1,-1/5,1/5\}$ & best equiangular lines & $+$ \\
$21d336$      & $1/5$   & $3$ & $3$ & $\{-3/5,-1/5,1/5\}$ & \cite{BannaiSloane} & \\
$21d512$      & $5/21$  & $3$ & $3$ & $\{-11/21,-1/7,5/21\}$ & -1/8 section of 22d891k3 & $+$ \\
$21d567$      & $1/4$   & $3$ & $3$ & $\{-1/2,-1/8,1/4\}$ & $\ast$; Fischer Tower & \\
$21d672$      & $3/11$  & $3$ & $3$ & $\{-5/11,-1/11,3/11\}$ & Genealogy $22d1024$ & $+$ \\
\bottomrule
\end{tabular}%
}
\par\vspace{0.7em}
\refstepcounter{table}\label{tab:slp-tight-main-a}
\smallskip
{\small\textbf{Table \thetable.} S-LP-tight regular constellations, dimensions $6$--$21$. Here "*" means the codes and constellations discovered by \textit{PackingStar}. "+" in Status means the code is not optimal as spherical code of its cos value. "s" means we have confirmed its non-uniqueness. For Genealogy, see next section. SRG (Strongly Regular Graph), some of ETF (Equiangular Tight Frame) and Universal Optima are not included.}
\end{center}

\begin{center}
\scriptsize
\setlength{\tabcolsep}{2.8pt}
\renewcommand{\arraystretch}{1.08}
\resizebox{\textwidth}{!}{%
\begin{tabular}{@{}l c c c l l c@{}}
\toprule
Code & $s$ & $\tau$ & $|I(C)|$ & $I(C)$ & Reference / construction & Status \\
\midrule
$21d1296$     & $11/35$ & $3$ & $3$ & $\{-13/35,-1/35,11/35\}$ & $U_4(3):2$ & $+$ \\
$22d352$      & $1/5$   & $3$ & $3$ & $\{-1,-1/5,1/5\}$ & $\ast$; Higman--Sims & \\
$22d550$      & $1/4$   & $5$ & $5$ & $\{-1,-1/4,-1/6,1/6,1/4\}$ & Antipodal McL\cite{DelsarteGoethalsSeidel} & \\
$22d750$      & $2/7$   & $3$ & $3$ & $\{-3/7,-1/14,2/7\}$ & $U_3(5):2$ & $+$ \\
$22d1024$     & $3/11$  & $3$ & $3$ & $\{-5/11,-1/11,3/11\}$ & \cite{1995Spherical} & $+$ \\
$22d1100$     & $2/7$   & $3$ & $3$ & $\{-3/7,-1/14,2/7\}$ & HS$:2$ & $+$ \\
$22d2025$     & $7/22$  & $4$ & $3$ & $\{-4/11,-1/44,7/22\}$ & \cite{DelsarteGoethalsSeidel,BoyvalenkovCherkashinDragnev2025} &  $+$  \\
$22d2816$     & $1/3$   & $5$ & $4$ & $\{-1,-1/3,0,1/3\}$ & \cite{DelsarteGoethalsSeidel,BoyvalenkovCherkashinDragnev2025} & \\
$22d7128$     & $2/5$   & $4$ & $4$ & $\{-1/2,-1/5,1/10,2/5\}$ & McL$:2$ & $+$ \\
$22d20736$    & $5/11$  & $5$ & $5$ & $\{-7/11,-4/11,-1/11,2/11,5/11\}$ & $U_6(2):2$ & $+$ \\
$23d2048$     & $7/23$  & $3$ & $3$ & $\{-9/23,-1/23,7/23\}$ & \cite{DelsarteGoethalsSeidel,BoyvalenkovCherkashinDragnev2025} & \\
$23d11178$    & $2/5$   & $5$ & $4$ & $\{-1/2,-1/5,1/10,2/5\}$ & \cite{DelsarteGoethalsSeidel,BoyvalenkovCherkashinDragnev2025} & \\
$23d47104$    & $7/15$  & $7$ & $5$ & $\{-3/5,-1/3,-1/15,1/5,7/15\}$ & \cite{DelsarteGoethalsSeidel,BoyvalenkovCherkashinDragnev2025} & \\
$23d48600$    & $11/23$ & $5$ & $5$ & $\{-13/23,-7/23,-1/23,5/23,11/23\}$ & \cite{DelsarteGoethalsSeidel,BoyvalenkovCherkashinDragnev2025} & \\
$23d93150$    & $1/2$   & $7$ & $6$ & $\{-1,-1/2,-1/4,0,1/4,1/2\}$ & \cite{leech1967notes,BoyvalenkovCherkashinDragnev2025} & \\
$48d52416000$ & $1/2$   & $11$& $8$ & $\{-1,0,\pm1/2,\pm1/3,\pm1/6\}$ & \cite{BoyvalenkovCherkashin2025,BoyvalenkovCherkashinDragnev2025} & \\
\bottomrule
\end{tabular}%
}
\par\vspace{0.7em}
\refstepcounter{table}\label{tab:slp-tight-main-b}
\smallskip
{\small\textbf{Table \thetable.} S-LP-tight regular constellations, dimensions $21$--$48$. SRG, some of ETF and Universal Optima are not included.}
\end{center}

All the result listed here are easily calculated via DGS's theorem 4.3~\cite{DelsarteGoethalsSeidel}. Some of them have been covered in ~\cite{BoyvalenkovCherkashinDragnev2025} and ~\cite{BoyvalenkovCherkashin2025}. And there are also three cases we think are very interesting but cannot prove its supported optimality without adding extra conditions.

\begin{center}
\scriptsize
\setlength{\tabcolsep}{2.8pt}
\renewcommand{\arraystretch}{1.08}
\begin{tabular}{@{}l c c c l l@{}}
\toprule
Code & $s$ & $\tau$ & $|I(C)|$ & $I(C)$ & Extra condition / reference \\
\midrule
$72d6218175600$ & $1/2$ & $11$ & $10$ & $\{-1, 0, \pm1/8, \pm1/4, \pm3/8, \pm1/2\}$ & momentum $Q_6=0$ ; Extremal Even Unimodular lattice\cite{Nebe2010} \\
$78d142155$     &    $1/4$   & $5$  & $5$  & $\{-1/2,-5/16,-1/8,1/16,1/4\}$ & FTC geometry (see \ref{sec:FTC})\\
$86d114939$     &    $1/4$   & $5$  & $4$  & $\{-1/2,-1/8,1/16,1/4\}$ & FTC geometry \\
\bottomrule
\end{tabular}
\par\vspace{0.7em}
\refstepcounter{table}\label{tab:slp-tight-extra-geometry}
\smallskip
{\small\textbf{Table \thetable.} S-LP-tight cases using additional geometric condition.}
\end{center}

\section{Genealogy of Spherical codes}
This section studies spherical-code genealogy, the organization of regular descendants arising from meaningful linear subspaces of a larger code. The \(\tfrac14\)-record chain from \(\texttt{6d27k3}\) to \(\texttt{22d891k3}\) is governed by a common LP certificate and organized by Fischer geometry. This leads to the Fischer Tower of \(\texttt{22d891k3}\), which also provides structured benchmarks for spherical-code optimization.

\subsection{Chain of \texorpdfstring{\(\tfrac14\)}{1/4} record codes and Fischer Geometry}
\label{sec:Fi21-constellation}

Let
\begin{equation}
\mathcal C=\left\{-\frac12,-\frac18,\frac14\right\}.
\end{equation}
For this cosine set the same cubic polynomial
\begin{equation}
p(t)=\left(t-\frac14\right)\left(t+\frac18\right)\left(t+\frac12\right)
\end{equation}
gives the relevant supported-LP certificate.  If \(Q_i^{(n)}\) denotes the normalized Gegenbauer polynomial on \(S^{n-1}\), then
\begin{equation}
p(t)=
\frac{24-n}{64n}Q_0^{(n)}(t)
+\frac{3(30-n)}{32(n+2)}Q_1^{(n)}(t)
+\frac{3(n-1)}{8n}Q_2^{(n)}(t)
+\frac{n-1}{n+2}Q_3^{(n)}(t).
\end{equation}
For \(n<24\) these coefficients are positive, and \(p\) vanishes on \(\mathcal C\). Hence the Delsarte--Goethals--Seidel bound gives
\begin{equation}
|X|\le \frac{p(1)}{p_0}=\frac{81n}{24-n}.
\end{equation}
For \(n=6,12,15,18,20,21,22\), this gives \(27,\ 81,\ 135,\ 243,\ 405,\ 567,\ 891\) respectively.
\textit{PackingStar} attains these values.  Thus the constellations
\begin{equation}
\texttt{6d27k3},\ 
\texttt{12d81k3},\ 
\texttt{15d135k3},\ 
\texttt{18d243k3},\ 
\texttt{20d405k3},\ 
\texttt{21d567k3},\ 
\texttt{22d891k3}
\end{equation}
are governed by one LP certificate; the largest one is the universally optimal \(891\)-point code in dimension \(22\) \cite{DelsarteGoethalsSeidel,CohnKumar2007}.

We now work inside \(\texttt{22d891k3}\).  It contains exactly \(693\) Schläfli subcodes \(\texttt{6d27k3}\).  Each original point lies in \(21\) of them, and two distinct Schläfli subcodes meet in either \(0\) or \(3\) points.  The disjointness graph on these \(693\) subcodes has parameters \((693,512,376,384)\), the Fischer graph of \(U_6(2)=\mathrm{Fi}_{21}\) \cite{cuypers2012lie}.  Thus the Fischer points are the Schläfli subcodes.

(\emph{Note.}  Fixing one Schläfli subcode also gives a decomposition \(891=27+27\cdot32\).  This is exactly a generalized version of the fiber tensor product construction in ~\ref{sec:fiber-tensor-product}, between the Schläfli base and the \(27\cdot32\) residue coming from the maximal embedding of the universally optimal \(\texttt{16d288k3}\) code into the Barnes--Wall \(\texttt{16d4320k1}\) kissing structure discussed in Section~\ref{sec:bw-no-15-NR}.  This observation is not used below.)

A Fischer line is a triple of pairwise disjoint Schläfli subcodes whose union is
\begin{equation}
\texttt{12d81k3}\cong \texttt{6d27k3}\otimes Y,
\end{equation}
where \(Y=\texttt{2d3}\) the equilateral triangle. This is the star of its constellation. Each disjoint pair has a unique third member.  There are \(59136\) such lines, and each Schläfli subcode lies on \(256\) of them \cite{cuypers2012lie}.

A Fischer affine plane \(AG(2,3)\) has \(9\) Fischer points and \(12\) Fischer lines.  In \(\texttt{22d891k3}\), its \(9\) Schläfli subcodes are pairwise disjoint, so their union has \(9\cdot27=243\) points.  Its span has dimension \(18\), giving a copy of \(\texttt{18d243k3}\).  The map from Fischer affine planes to spherical-code supports is two-to-one: the \(197120\) Fischer affine planes give \(98560\) distinct \(243\)-point supports, each carrying exactly two affine-plane structures.

The dual affine planes give another intermediate code.  A dual affine plane of order \(2\) has six points \(p_{ij}\), \(1\le i<j\le4\), and four lines \(L_i=\{p_{ij}:j\ne i\}.\)
Thus \(p_{ij}\) and \(p_{kl}\) are collinear exactly when \(\{i,j\}\cap\{k,l\}\ne\emptyset\).  In the spherical code, collinear pairs correspond to disjoint Schläfli subcodes, while the three opposite pairs
\begin{equation}
(p_{12},p_{34}),\quad (p_{13},p_{24}),\quad (p_{14},p_{23})
\end{equation}
meet in \(3\) original points each.  Hence the union has \(6\cdot27-3\cdot3=153\) points and spans dimension \(16\).  This is the Fischer dual-affine-plane configuration \(\texttt{16d153k3}\).  Its total count is \(59136\cdot135/4=1995840\) \cite{cuypers2012lie}.

Two \(\texttt{18d243k3}\) supports sharing a Fischer line have union size \(243+243-81=405\) and span dimension \(20\), giving \(\texttt{20d405k3}\).  Each such \(\texttt{20d405k3}\) has three \(\texttt{21d567k3}\) completions inside \(\texttt{22d891k3}\); they have common intersection \(\texttt{20d405k3}\), and we have this "sunflower decomposition":
\begin{equation}
891=405\sqcup162\sqcup162\sqcup162.
\end{equation}
Where \(567=405\sqcup162\) forms a "petal". Thus we obtain the embedded inclusion sequence
\begin{equation}
\texttt{6d27}
\subset
\texttt{12d81}
\subset
\texttt{18d243}
\subset
\texttt{20d405}
\subset
\texttt{21d567}
\subset
\texttt{22d891}.
\end{equation}

\subsection{Fischer Tower and Genealogy}

The sequence above explains the \(\texttt{12d81k3}\), \(\texttt{18d243k3}\), \(\texttt{20d405k3}\), and \(\texttt{21d567k3}\) layers.  It misses \(\texttt{16d153k3}\), which lies between \(\texttt{12d81k3}\) and \(\texttt{20d405k3}\) but is not contained in \(\texttt{18d243k3}\).  It also does not produce \(\texttt{15d135k3}\).  Thus the Fischer geometry is only the first approximation to the subcode structure of \(\texttt{22d891k3}\).

The finer object is the family of meaningful linear-subspace descendants of \(\texttt{22d891k3}\).  A full enumeration of the subspace-generated matroid is quickly dominated by irregular intersections.  We therefore use a non-enumerative genetic search on subspaces, guided by number of points, first statistic, and SVD statistic. We call this method to study subcodes of a fixed spherical code the Genealogy method.

We compute the Genealogy of \(\texttt{22d891k3}\), and filtered the resulted space by keeps only the descendants with a single first-statistic type and a single SVD type, i.e. the distance-invariant tight-frame descendants, and then add the Fischer dual-affine-plane node \(\texttt{16d153k3}\). We call the resulting genealogical diagram the \emph{Fischer Tower} of \texttt{22d891k3};
the terminology is motivated by Fischer's theory of \(3\)-transposition groups
\cite{Fischer1971}.

\begin{figure}[!htb]
    \centering
    \includegraphics[angle=0, height=0.8\textheight]{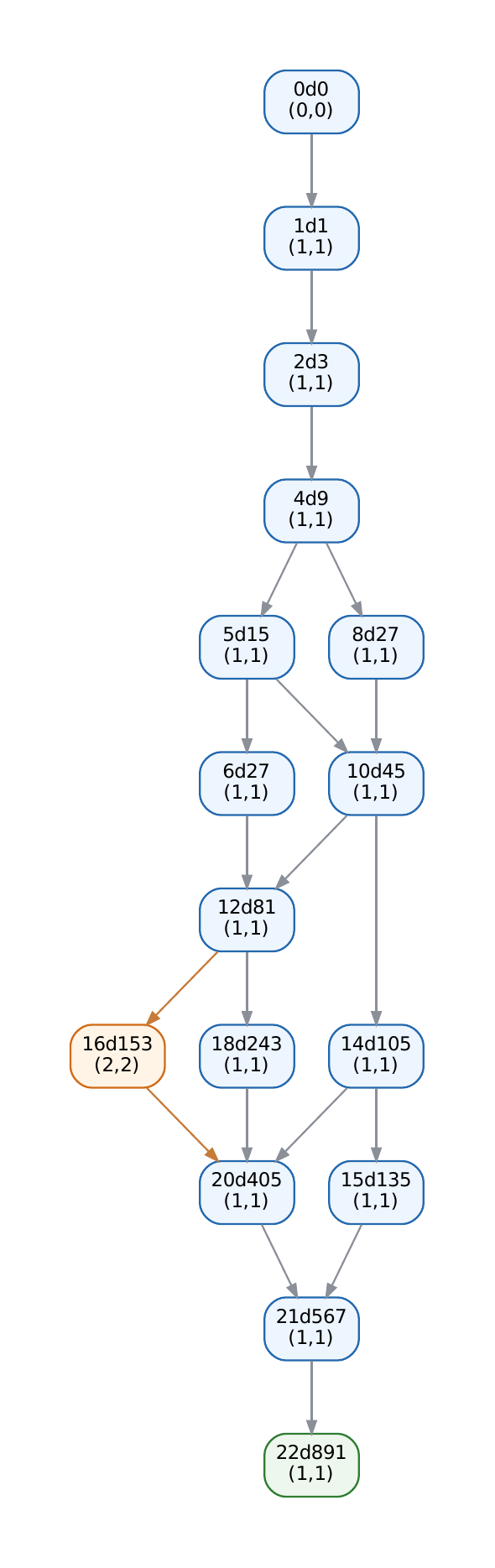}
    \vspace{-0.8em}
    \caption{\fontsize{11}{11}\selectfont
    \emph{Fischer Tower}: the main part of the genealogy of the linear-subspace spherical codes associated with $\texttt{22d891k3}$. 
    Each node represents a descendant spherical-code configuration obtained by restricting to a linear subspace, and each edge records a descent relation between configurations. 
    The diagram contains regular descendants of single SVD type and single first-statistic type, together with the two-orbit Fischer dual affine plane configuration $\texttt{16d153k3}$. 
    It reveals a structured family of highly regular codes organized around the $\texttt{22d891k3}$ configuration.
    }
    \label{genealogy_lr}
\end{figure}

\subsection{Genealogy as a benchmark for spherical-code optimization}

For \(-1\le \alpha<1\), let \(K(n,\alpha)\) denote the largest size of a spherical code \(X\subset S^{n-1}\) with \(\langle x,y\rangle\le \alpha\) for all distinct \(x,y\in X\).

The genealogy method is also useful as a benchmark generator for spherical-code optimization.  Besides \(\texttt{22d891k3}\), we computed genealogies for several affine subcodes of the Leech minimal-vector shell.  Some nodes in these genealogies, and some affine layers between nodes, give structured frame, completing holes of the frames gives new lower bounds in the table~\ref{tab:Genealogy_new_bounds}.  All inner products are verified exactly.

\begin{table}[!htb]
    \centering
    \footnotesize
    \setlength{\belowcaptionskip}{8pt}
    \caption{\fontsize{11}{11}\selectfont
    New lower bounds $K_{\mathrm{new}}(n,\alpha)$ discovered by the genealogy method, compared with the previous best known values $K_{\mathrm{prev}}(n,\alpha)$.}
    \renewcommand{\arraystretch}{1.15}
    \setlength{\tabcolsep}{14pt}
    \begin{tabular}{cccc}
        \hline
        $n$ &
        $\alpha$ &
        $\boldsymbol{K_{\mathrm{new}}(n,\alpha)}$ &
        $K_{\mathrm{prev}}(n,\alpha)$ \\
        \hline

        13 & $7/19$ &
        \textbf{240} &
        231 \cite{cohn2024_spherical_codes} \\

        15 & $\sqrt{3/40}$ &
        \textbf{416} &
        387 \cite{cohn2024_spherical_codes} \\

        15 & $5/11$ &
        \textbf{1215} &
        991 \cite{cohn2024_spherical_codes} \\

        16 & $3/8$ &
        \textbf{575} &
        567 \cite{cohn2024_spherical_codes} \\

        16 & $5/13$ &
        \textbf{642} &
        570 \cite{cohn2024_spherical_codes} \\

        16 & $29/73$ &
        \textbf{715} &
        685 \cite{cohn2024_spherical_codes} \\

        16 & $2/5$ &
        \textbf{771} &
        769 \cite{cohn2024_spherical_codes} \\

        16 & $3/7$ &
        \textbf{1182} &
        1056 \cite{1995Spherical} \\

        17 & $3/8$ &
        \textbf{745} &
        742 \cite{cohn2024_spherical_codes} \\

        17 & $5/13$ &
        \textbf{1026} &
        768 \cite{cohn2024_spherical_codes} \\

        17 & $2/5$ &
        \textbf{1104} &
        1088 \cite{cohn2024_spherical_codes} \\

        18 & $3/19$ &
        \textbf{100} &
        98 \cite{cohn2024_spherical_codes} \\

        18 & $13/69$ &
        \textbf{128} &
        123 \cite{cohn2024_spherical_codes} \\

        18 & $3/8$ &
        \textbf{1155} &
        869 \cite{cohn2024_spherical_codes} \\

        19 & $4/19$ &
        \textbf{202} &
        199 \cite{cohn2024_spherical_codes} \\

        19 & $3/11$ &
        \textbf{404} &
        360 \cite{cohn2024_spherical_codes} \\

        20 & $3/11$ &
        \textbf{524} &
        476 \cite{cohn2024_spherical_codes} \\

        20 & $\sqrt{3/40}$ &
        \textbf{567} &
        477 \cite{cohn2024_spherical_codes} \\

        20 & $3/10$ &
        \textbf{840} &
        641 \cite{cohn2024_spherical_codes} \\

        20 & $4/13$ &
        \textbf{1002} &
        774 \cite{cohn2024_spherical_codes} \\

        21 & $33/145$ &
        \textbf{373} &
        353 \cite{cohn2024_spherical_codes} \\

        21 & $4/15$ &
        \textbf{715} &
        684 \cite{cohn2024_spherical_codes} \\

        21 & $5/17$ &
        \textbf{974} &
        859 \cite{cohn2024_spherical_codes} \\

        21 & $3/10$ &
        \textbf{1092} &
        879 \cite{cohn2024_spherical_codes} \\

        22 & $3/11$ &
        \textbf{1068} &
        1024 \cite{1995Spherical} \\








        \hline
    \end{tabular}
    \label{tab:Genealogy_new_bounds}
\end{table}

\section{Euclidean representations and subgroup geometry}
\label{sec:euclidean-representations}


The present viewpoint starts from Conway's geometry of the Leech lattice.
The \(196560\) minimal vectors of the Leech lattice form a spherical code in
\(S^{23}\subset \mathbb R^{24}\). After identifying antipodal pairs, one obtains
\(98280\) projective lines. The full linear symmetry group of the lattice is
\(Co_0=\operatorname{Aut}(\Lambda)=2\cdot Co_1\), while the induced faithful
action on these \(98280\) lines is \(Co_1\). Thus the most natural
spherical-code realization of the largest simple Conway group is the action of
\(Co_1\) on the projectivized minimal vectors of the Leech lattice
\cite{ConwaySloaneSPLAG,WilsonFSGNotes}.

This is also the first model for the subgroup geometry considered here.
Conway's Leech-lattice methods already show that important sporadic groups
arise as stabilizers of small metric patterns in a highly symmetric Euclidean
configuration. The groups \(Co_2\) and \(Co_3\) appear as stabilizers of suitable
Leech-lattice vectors, while \(McL\) and \(HS\) appear from suitable
Leech-lattice triangles. Curtis developed this point of view further in his
work on subgroups of \(\cdot O\), especially through lattice stabilizers and
local structure; Wilson describes this as part of the ``lattice methods''
tradition in the study of maximal subgroups of sporadic groups
\cite{CurtisSubgroupsI,CurtisSubgroupsII,Wilson2017Sporadic}.

The Conway--Curtis cross introduced earlier in this paper belongs exactly to
this lineage. In our preceding construction it appears as the core frame inside
the \(496\)-point maximal \(1/4\)-subconfiguration of the Leech minimal-vector
code found by \textit{PackingStar}, and it is one of the basic ingredients in our new
kissing-number lower-bound constructions in dimensions \(25\) through \(31\).
From the subgroup point of view, its role is especially sharp: the stabilizer
of a cross in \(Co_0\) has shape \(2^{12}:M_{24}\), and its image in \(Co_1\)
is the maximal subgroup \(2^{11}:M_{24}<Co_1\). Its index is
\(8292375=N_8/48\), where \(N_8=398034000\) is the number of norm-eight
vectors in the third Leech shell; this is exactly the number of crosses
\cite{WilsonFSGNotes,Wilson1983Co1,Atlas1985}.

We use the term \emph{Euclidean representation} in this spirit. For a finite
group \(G\), a Euclidean representation in the present sense starts from a
finite \(G\)-invariant spherical code
\(\mathcal C\subset S^{n-1}\subset \mathbb R^n\), with \(G\leq O(n)\), together
with a chosen collection \(\mathfrak P(\mathcal C)\) of elementary metric
patterns inside \(\mathcal C\). The spherical formulation is natural for finite
groups: after translating to the fixed barycenter of an orbit, every finite
Euclidean isometry action becomes orthogonal, and each finite orbit lies on a
sphere.

We do not only ask for the action of \(G\) on the points of \(\mathcal C\). We
also study the action of \(G\) on elementary Euclidean substructures inside
\(\mathcal C\): points, antipodal pairs, regular triangles, hexagons, frames,
metric lines and planes, special partitions, and other small configurations
defined by inner products and incidence relations. If \(\Sigma\) is such a
pattern, then its stabilizer
\(G_\Sigma=\{g\in G:g\Sigma=\Sigma\}\) is a natural subgroup of \(G\), and the
orbit of \(\Sigma\) gives the coset action
\(G\curvearrowright\{g\Sigma:g\in G\}\cong G/G_\Sigma\). Thus a spherical code
does not merely realize a group as a symmetry group; it may also realize
permutation representations of that group through its internal metric geometry.

The terminology is compatible with Musin's study of Euclidean and spherical
representations of graphs as two-distance sets \cite{MusinTwoDistance}. If such
a graph has a large automorphism group, its spherical representation is at the
same time a Euclidean realization of that group. The terminology is particularly
natural in the present context of kissing number configurations; Musin also proved the
four-dimensional kissing number \(k(4)=24\) \cite{MusinKissing4}.

This viewpoint is compatible with, but different from, several classical
geometric traditions in group theory. Coxeter theory gives a Euclidean theory
of reflection groups, root systems, and regular polytopes
\cite{CoxeterRegularPolytopes,HumphreysCoxeter}. Tits buildings give incidence
geometries for groups of Lie type, while diagram and coset geometries organize
group actions through prescribed types of objects and incidences
\cite{TitsBuildings,Buekenhout1979,Giudici2012}. These theories usually begin
with reflection data, field geometry, or a chosen system of subgroups. Here the
starting point is a concrete spherical code
\(\mathcal C\subset S^{n-1}\subset \mathbb R^n\), and the question is which
subgroup data are forced by its elementary metric patterns.

The closest structural analogy is Aschbacher's theory of maximal subgroups of
finite classical groups. Aschbacher showed that maximal subgroups of finite
classical groups fall into eight geometric classes, together with an almost
simple residual class \cite{Aschbacher1984}. In that theory, the relevant
geometry is finite-field linear algebra. The present work seeks an analogous
role for Euclidean and spherical-code geometry: the basic objects are not
subspaces over a finite field, but metric patterns inside a finite spherical
code.

The guiding principle is therefore
\[
    \boxed{
    \text{elementary Euclidean pattern}
    \quad\longleftrightarrow\quad
    \text{subgroup type}.
    }
\]
A basic Euclidean substructure in a spherical code---a point, an antipodal pair,
a triangle, a hexagon, a frame, a metric line or plane, or a distinguished
decomposition---may be the geometric witness for a significant subgroup of
\(G\). When these elementary patterns account for all maximal subgroups of
\(G\), the spherical code becomes not merely a Euclidean realization of \(G\),
but a Euclidean map of its subgroup geometry.

More explicitly, relative to the chosen class \(\mathfrak P(\mathcal C)\) of
allowed metric patterns, we say that \(\mathcal C\) is a
\emph{maximal-subgroup-complete Euclidean representation} of \(G\) if every
conjugacy class of maximal subgroups of \(G\) occurs as \(G_\Sigma\) for some
\(\Sigma\in\mathfrak P(\mathcal C)\). In this sense, the study is a direct
continuation of the Conway--Curtis work and the ATLAS tradition: the ATLAS
supplies the subgroup data, Conway and Curtis provide the geometric prototype,
and the present work organizes these ideas around spherical codes arising from
kissing-number constructions.

\subsection{\texorpdfstring{The baby example: the icosahedral code and \(A_5\)}{The baby example: the icosahedral code and A5}}
\label{subsec:a5-icosahedral-code}

Let \(\mathcal I_{12}\subset S^2\subset \mathbb R^3\) be the \(12\) vertices of
a regular icosahedron. This is the classical three-dimensional kissing
configuration: the Newton--Gregory problem asked whether \(13\) equal spheres
could touch a central one, and the answer \(k(3)=12\) was proved by Schütte and
van der Waerden; Musin later gave another proof using an extension of the
Delsarte method \cite{SchutteWaerden1953,MusinKissing3}. The same \(12\)-point
configuration is also a sharp spherical code, hence universally optimal in the
sense of Cohn and Kumar \cite{CohnKumar2007}. Thus the regular icosahedron is
both the smallest nontrivial model for the present subgroup geometry and one of
the classical starting points of spherical-code geometry.

Let \(G=\operatorname{Rot}(\mathcal I_{12})\cong A_5\). The three conjugacy
classes of maximal subgroups of \(A_5\) are visible from the most elementary
metric patterns in \(\mathcal I_{12}\).

First, an antipodal pair of vertices \(\{p,-p\}\) is a \(5\)-fold axis. There
are \(6\) such pairs, and the stabilizer of one of them is \(D_5\) of order
\(10\). Thus the corresponding coset action is \(A_5/D_5\), of degree \(6\).

Second, a triangular face \(\Delta=\{p,q,r\}\) is a regular triangle of minimal
edge length inside \(\mathcal I_{12}\). The dual dodecahedron need not be added
as an external object: its \(20\) vertices are precisely the normalized
barycentres of these \(20\) face-triangle patterns. Hence an antipodal pair of
dodecahedron vertices is equivalently a pair of opposite face triangles in
\(\mathcal I_{12}\). There are \(10\) such pairs, and the stabilizer of one of
them is \(D_3\cong S_3\), of order \(6\). This gives the degree-\(10\) action
\(A_5/S_3\).

Third, consider a partition of \(\mathcal I_{12}\) into four mutually disjoint
face triangles. Equivalently, take an inscribed regular tetrahedron in the dual
dodecahedron. Such partitions form two chiral orbits of length \(5\) under
\(A_5\), and the stabilizer of a partition in either orbit is \(A_4\), the
rotation group of the tetrahedron. This gives the degree-\(5\) action
\(A_5/A_4\).

Thus the full maximal-subgroup structure of \(A_5\) is recovered from elementary
Euclidean substructures of the \(12\)-point icosahedral code:
\[
\begin{array}{c|c|c}
\text{Euclidean pattern in } \mathcal I_{12}
    & \text{number of copies}
    & \text{stabilizer in } A_5 \\ \hline
\text{antipodal vertex pair}
    & 6
    & D_5 \\
\text{opposite pair of face triangles}
    & 10
    & D_3\cong S_3 \\
\text{four-triangle tetrahedral partition}
    & 5 \text{ in each chiral orbit}
    & A_4 .
\end{array}
\]
Consequently
\begin{equation}
    \operatorname{MaxSub}(A_5)=\{D_5,\;S_3,\;A_4\}
\end{equation}
is realized by the basic Euclidean geometry of \(\mathcal I_{12}\). The
remaining visible axial structure, namely the \(15\) twofold axes coming from
antipodal edge-midpoints, has stabilizer \(V_4\); it is geometrically natural
but not maximal, since it lies inside a tetrahedral \(A_4\). This illustrates
the central point of the theory in the smallest possible nonabelian simple
group: a highly symmetric spherical code can serve not only as a representation
of the group, but as a Euclidean dictionary for its maximal subgroups
\cite{CoxeterRegularPolytopes,DixonMortimer1996}.

\subsection{Euclidean representation of \texorpdfstring{$U_6(2)$}{U6(2)} and \texorpdfstring{$\operatorname{Aut}(U_6(2))$}{Aut(U6(2))}}
\label{sec:u62-euclidean-representation}

We use the unique universally optimal 1/4 spherical code \texttt{22d891k3}, as a Euclidean model for \(U_6(2)\). Its Automorphism group is \(\widehat G=\operatorname{Aut}(\texttt{22d891k3})\cong\operatorname{Aut}(U_6(2))=U_6(2):S_3\); we write \(G_0=U_6(2)\) for the socle. The group acts simultaneously as an 891 points permutation and an orthogonal linear group.

The main point of this case is that the metric substructures of \texttt{22d891k3} give a maximal-subgroup-complete Euclidean representation for both \(G_0\) and \(\widehat G\). This is the first substantial complete family arising from the present method: the same code realizes not only the familiar Sch/Fischer geometry of \(U_6(2)\), but also the socle-only three-colour layers and the full-group-only special simplex layer. The Fischer-space background and for \(U_6(2)\) is discussed in \cite{cuypers2012lie}; here the objects are deliberately recorded in Euclidean spherical-code language.

\begin{table}[htbp]
\centering
\scriptsize
\setlength{\tabcolsep}{2.8pt}
\begin{tabularx}{\textwidth}{@{}>{\raggedright\arraybackslash}p{0.245\textwidth}>{\centering\arraybackslash}p{0.065\textwidth}>{\raggedright\arraybackslash}p{0.125\textwidth}>{\raggedright\arraybackslash}p{0.18\textwidth}>{\raggedright\arraybackslash}X@{}}
\toprule
Maximal subgroup of \(\widehat G=U_6(2):S_3\) & Index & Euclidean object & Reduced to \(G_0\) & Note \\
\midrule
\((U_5(2)\times3):2\) & 672 & isoclinic \(Y\)-spread & \(U_5(2)\) & Partition into 297 \(Y\)'s; perfect matching condition. \\
\(2^{1+8}:(U_4(2)\times3):2\) & 693 & \texttt{6d27k3} & \(2^{1+8}:U_4(2)\) & Sch object. \\
\(2^9:L_3(4).S_3\) & 891 & point & \(2^9:L_3(4)\) & Point stabilizer. \\
\(3^5:(S_6\times2)\) & 157696 & \texttt{20d405k3} & \(3^4:S_6\) & Not maximal in \(G_0\); central object in \(891=405+3\cdot162\). \\
\(2^{4+8}:(S_3\times A_5).S_3\) & 6237 & \(Y=\texttt{2d3}\) & \(2^{4+8}:(S_3\times A_5)\) & A regular \(-1/2\)-triangle. \\
\(L_2(8):3\times S_3\) & 6082560 & special 9-simplex & \(L_2(8):3\) & Not maximal in \(G_0\); distinguished \(-1/8\) 9-simplex. \\
\((U_4(2)\times S_3\times3).2\) & 59136 & \texttt{12d81k3} & \(U_4(2)\times S_3\) & Three disjoint \texttt{6d27k3}'s. \\
\(3^{1+4}:(Q_8\times Q_8):S_3.S_3\) & 98560 & \texttt{18d243k3} & \(3^{1+4}:(Q_8\times Q_8):S_3\) & Pure Euclidean 243-point support. \\
\(L_3(4):D_{12}\) & 228096 & \(-1/8\)-edge & \(L_3(4):2_1\) & Unordered edge of the \(-1/8\)-graph. \\
\bottomrule
\end{tabularx}
\caption{Euclidean maximal-subgroup dictionary for \(\operatorname{Aut}(U_6(2))\), omitting \(U_6(2):2\) and \(U_6(2):3\).}
\label{tab:aut-u62-maximal-euclidean}
\end{table}

\begin{table}[htbp]
\centering
\scriptsize
\setlength{\tabcolsep}{3pt}
\begin{tabularx}{\textwidth}{@{}>{\raggedright\arraybackslash}p{0.235\textwidth}>{\centering\arraybackslash}p{0.095\textwidth}>{\raggedright\arraybackslash}p{0.135\textwidth}>{\centering\arraybackslash}p{0.07\textwidth}>{\raggedright\arraybackslash}X@{}}
\toprule
Maximal subgroup of \(G_0=U_6(2)\) & Index & Euclidean object & Classes & Note \\
\midrule
\(U_5(2)\) & 672 & isoclinic \(Y\)-spread & 1 &  \\
\(2^{1+8}:U_4(2)\) & 693 & \texttt{6d27k3} & 1 &  \\
\(2^9:L_3(4)\) & 891 & point & 1 &  \\
\(U_4(3):2_2\) & 1408 per class & \texttt{21d567k3} & 3 & Oriented normals give three \texttt{22d2816k2} codes; a cross-colour pair with normal cosine \(1/2\) has hyperplane intersection \texttt{20d405k3}. \\
\(2^{4+8}:(S_3\times A_5)\) & 6237 & \(Y=\texttt{2d3}\) & 1 &  \\
\(S_6(2)\cong Sp_6(2)\) & 6336 per class & \texttt{15d135k3} & 3 & Finite-geometric note: the Lagrangian geometry of \(W(5,2)\). \\
\(M_{22}\) & 20736 per class & \(21d77\) facet & 3 & Three colours; defined below by a first-stat certificate. \\
\(U_4(2)\times S_3\) & 59136 & \texttt{12d81k3} & 1 &  \\
\(3^{1+4}:(Q_8\times Q_8):S_3\) & 98560 & \texttt{18d243k3} & 1 &  \\
\(L_3(4):2_1\) & 228096 & \(-1/8\)-edge & 1 &  \\
\bottomrule
\end{tabularx}
\caption{Euclidean maximal-subgroup dictionary for the socle \(U_6(2)\). Repetition are suppressed.}
\label{tab:u62-maximal-euclidean}
\end{table}

\subsubsection*{Isoclinic \texorpdfstring{$Y$}{Y}-spreads}

A \(Y\) is a \(\texttt{2d3}\) subcode \(\{a_1,a_2,a_3\}\) with all off-diagonal inner products \(-1/2\). An isoclinic \(Y\)-spread is a partition of \texttt{22d891k3} into 297 such triples such that, for any two triples, the \(-1/2\)-edges between them are either absent or form a perfect matching. To see the isoclinic property, write \(K=(\langle a_i,a_j\rangle)=\frac32 I-\frac12J\). If \(Z=\{b_1,b_2,b_3\}\) and \(B=(\langle a_i,b_j\rangle)\), then after relabelling either \(B=-\frac12K\) or \(B=\frac14K\). Since \(K\) restricts to \(\frac32I\) on \({\bf 1}^{\perp}\), the cross-map between the two 2-planes is scalar; the principal cosines are therefore \((1/2,1/2)\) or \((1/4,1/4)\). The 672 such spreads give the degree-672 action of \(U_6(2)\) and have stabilizer \(U_5(2)\).

\subsubsection*{Special 9-simplices}

A 9-simplex is a 9-subset \(S\subset\texttt{22d891k3}\) with \(\langle x,y\rangle=-1/8\) for all distinct \(x,y\in S\). It is called special if, for every triple \(T\subset S\), the six code points having inner product \(1/4\) with precisely the points of \(T\) and inner product \(-1/8\) with the other six points of \(S\) form two disjoint \(-1/8\)-triangles. The special 9-simplices form a single \(\widehat G\)-orbit of size 6082560 and it's the only "star" orbit of 9-simplices; their setwise stabilizer in \(\widehat G\) is \(L_2(8):3\times S_3\), and the reduced stabilizer in \(G_0\) is \(L_2(8):3\), which is not maximal. Hence the special 9-simplex is a full-group maximal object, not a socle maximal object.

\subsubsection*{The \texorpdfstring{$M_{22}$}{M22}-facet construction}
This is the only convex-geometric construction in the tables. 
Let \(F\subset\texttt{22d891k3}\) have size \(77\). 
We call \(F\) an \(M_{22}\)-facet if its internal first-stat is
\begin{equation}
\mathsf{Stat}_1(F)=
\left\{\!\left\{
((-\tfrac18)^{16},(\tfrac14)^{60},1^1)^{77}
\right\}\!\right\},
\end{equation}
and its centroid functional
\begin{equation}
\ell_F(r)=\sum_{q\in F}\langle r,q\rangle
\end{equation}
satisfies
\begin{equation}
\ell_F(r)<14
\qquad
(r\in\texttt{22d891k3}\setminus F).
\end{equation}
Indeed the displayed first-stat gives \(\ell_F(q)=14\) for every \(q\in F\), and also implies \(\dim\operatorname{aff}(F)=21\). Hence \(\operatorname{conv}(F)\) is an exposed facet of \(\operatorname{conv}(\texttt{22d891k3})\).

These are purely Euclidean conditions. If \(s_F=\sum_{q\in F}q\), then \(\langle s_F,p\rangle=1+60/4-16/8=14\) for \(p\in F\), and \(\|s_F\|^2=77\cdot14=1078\). Thus \(h_F=s_F/\sqrt{1078}\) is a unit normal with contact height \(14/\sqrt{1078}=2/\sqrt{22}\) on \(F\), while all other code points have smaller height. It follows that \(\operatorname{conv}(F)\) is an exposed facet of \(\operatorname{conv}(\texttt{22d891k3})\). The stabilizer of such a facet in \(G_0\) has order \(|U_6(2)|/20736=443520=|M_{22}|\); the group name is therefore assigned after the facet has been defined metrically. There are three \(G_0\)-classes of these facets, each of size 20736, and the outer \(S_3\) permutes the three colours. They are not all facets of \(\operatorname{conv}(\texttt{22d891k3})\): oriented \texttt{21d567k3} normals, for instance, give additional 162-vertex exposed facets.

\subsection{Euclidean representation of \texorpdfstring{$Sp_6(2)$}{Sp6(2)} from \texorpdfstring{\texttt{7d126k1}}{7d126k1}}
\label{subsec:sp62-e7-euclidean}

We use the ordinary kissing number configuration \texttt{7d126k1}, namely the normalized root system \(R=R(E_7)\subset S^6\), as a Euclidean model for \(Sp_6(2)\).  Let \(\mathcal L=R/\{\pm1\}\) be the set of its \(63\) antipodal root lines.  Then \(G=\operatorname{Aut}(\mathcal L)\cong W(E_7)/\{\pm1\}\cong Sp_6(2)\).  The point of this subsection is to record the ATLAS maximal subgroup classes of \(G\) \cite{Atlas1985} in the language of subcodes and decompositions of \texttt{7d126k1}.  The finite-geometric dictionary from the opening \(W(5,2)\)--\(E_7\) section is used only as a verification coordinate system; the objects below are stated as Euclidean patterns in the root code.

An \(X_m\) is a set of \(m\) mutually orthogonal antipodal root lines, equivalently a \(2m\)-point cross polytope in \(R\).  Thus an \(X_7\) is a maximal frame, and a \(9X_7\)-spread is a decomposition \(\texttt{7d126k1}=9X_7\).  An \(A_2\)-hexagon is an antipodal regular hexagon \(Y\sqcup(-Y)\) in a root plane.  We also use the \texttt{6d72k1} and \texttt{7d56k1} root subcodes already appearing above: the former is an \(E_6\) subsystem, as in the decomposition \(\texttt{6d72k1}=9X_4\) of Subsection~\ref{subsec:two-decompositions-6d72}, while the latter is an \(A_7\) subsystem.

\begin{table}[htbp]
\centering
\scriptsize
\setlength{\tabcolsep}{3pt}
\begin{tabularx}{\textwidth}{@{}>{\raggedright\arraybackslash}p{0.235\textwidth}>{\centering\arraybackslash}p{0.095\textwidth}>{\raggedright\arraybackslash}p{0.135\textwidth}>{\raggedright\arraybackslash}X@{}}
\toprule
Maximal subgroup of \(G=Sp_6(2)\) & Index & Euclidean object & Note \\
\midrule
\(U_4(2):2\) & 28 & \texttt{6d72k1} & An \(E_6\) root subsystem of \texttt{7d126k1}. \\
\(A_8:2\cong S_8\) & 36 & \texttt{7d56k1} & An \(A_7\) root subsystem of \texttt{7d126k1}. \\
\(2^5:A_6:2\) & 63 & root line \(X_1\) & One antipodal pair \(\{\pm r\}\subset R(E_7)\). \\
\(U_3(3):2\) & 120 & \(3/7\)-section & A \(63\)-subset of the \(135\) copies of \(X_7\), meeting every \(9X_7\)-spread in \(3\) or \(7\) blocks. \\
\(2^6:L_3(2)\) & 135 & \(X_7\) & A maximal cross polytope in \texttt{7d126k1}. \\
\(\bigl(2^{1+4}\times2^2\bigr):(S_3\times S_3)\) & 315 & closed \(X_3\) & Three mutually orthogonal root lines contained in exactly three \(X_7\)'s. \\
\(S_3\times A_6:2\) & 336 & \(A_2\)-hexagon & An antipodal hexagon \(Y\sqcup(-Y)\); this is the antipodal companion to the \(42Y\) phenomenon in Subsection~\ref{subsec:S08-7d126-42Y}. \\
\(L_2(8):3\) & 960 & \(9X_7\)-spread & A decomposition \(\texttt{7d126k1}=9X_7\), used throughout the fiber tensor product construction in Section~\ref{sec:fiber-tensor-product}. \\
\bottomrule
\end{tabularx}
\caption{Euclidean maximal-subgroup dictionary for \(Sp_6(2)\) from the projectivized \(E_7\) kissing code.}
\label{tab:sp62-e7-euclidean}
\end{table}

\subsubsection*{Closed \texorpdfstring{$X_3$}{X3}'s}
An \(X_3\) in \texttt{7d126k1} is a triple of mutually orthogonal antipodal root lines.  There are many such triples, so orthogonality alone is not the right Euclidean line object.  We call an \(X_3\) \emph{closed} if it is contained in exactly three maximal frames \(X_7\).  The closed \(X_3\)'s form a single \(G\)-orbit of size \(315\), and the stabilizer of one of them has shape \(\bigl(2^{1+4}\times2^2\bigr):(S_3\times S_3)\).  This definition uses only containment among cross polytopes of the kissing code; it is the Euclidean refinement needed to separate the \(315\)-row from arbitrary orthogonal triples.

\subsubsection*{\texorpdfstring{$3/7$}{3/7}-sections}
Let \(\mathcal X_7(R)\) be the set of the \(135\) copies of \(X_7\) in \texttt{7d126k1}, and let \(\mathcal D_9(R)\) be the set of the \(960\) decompositions \(\texttt{7d126k1}=9X_7\).  A \emph{\(3/7\)-section} is a subset \(\mathcal Q\subset\mathcal X_7(R)\) of size \(63\) such that \(|\mathcal Q\cap\mathcal S|\in\{3,7\}\) for every \(\mathcal S\in\mathcal D_9(R)\).  Equivalently, after fixing one spread \(\mathcal S_0=\{X_1,\ldots,X_9\}\), every choice of \(3\) or \(7\) blocks of \(\mathcal S_0\) has a unique \(3/7\)-closure; hence the number of sections is \(\binom{9}{3}+\binom{9}{7}=84+36=120\).  These sections form a single \(G\)-orbit, and the stabilizer of one section is \(U_3(3):2\).  The associated closed-\(X_3\) web is recovered afterwards by taking the closed \(X_3\)'s contained in three members of \(\mathcal Q\), so the table may be read using only the basic objects \(X_7\), \(9X_7\)-spreads, and this one intersection condition.

Thus \texttt{7d126k1} gives a maximal-subgroup-complete Euclidean representation of \(Sp_6(2)\).

\subsection{Euclidean representation of \texorpdfstring{$U_4(2)$}{U4(2)} from \texorpdfstring{\texttt{6d27k3}}{6d27k3}}
\label{subsec:u42-schlafli-euclidean}

We use the Schl\"afli code $C=\texttt{6d27k3}\subset S^5$, with off-diagonal cosine set $\{-\tfrac12,\tfrac14\}$, as a Euclidean model for $U_4(2)$.  Its full automorphism group is $\widehat G=\operatorname{Aut}(C)\cong W(E_6)\cong U_4(2):2$; we write $G=U_4(2)$ for the socle, the maximal subgroups of $U_4(2):2$ is simply the lift of those of $U_4(2)$ and $U_4(2)$ itself.  

The companion root code is not taken as an independent input: it is recovered from the $1/4$-edges of $C$ by $R(C)=\{\pm(x-y)/\|x-y\|:x,y\in C,\ \langle x,y\rangle=\tfrac14\}\cong\texttt{6d72k1}$.  Thus the $216$ positive edges fall into $36$ antipodal directions (unlike in the $U_6(2)$'s $-1/8$ edge case where they have no parallel edges), the root lines of the $E_6$ configuration.  The decompositions of this derived code, especially $\texttt{6d72k1}=9X_4$ and $\texttt{6d72k1}=8Y_3$, were described in Subsection~\ref{subsec:two-decompositions-6d72}.  The graph-theoretic labels below match Brouwer's graph page~\cite{BrouwerSchlaefli}; here they are translated into subcodes and spreads of $C$.

\begin{table}[htbp]
\centering
\scriptsize
\setlength{\tabcolsep}{3pt}
\begin{tabularx}{\textwidth}{@{}>{\raggedright\arraybackslash}p{0.235\textwidth}>{\centering\arraybackslash}p{0.095\textwidth}>{\raggedright\arraybackslash}p{0.135\textwidth}>{\raggedright\arraybackslash}X@{}}
\toprule
Maximal subgroup of $G=U_4(2)$ & Index & Euclidean object & Note \\
\midrule
$2^4:A_5$ & $27$ & point & One point of $\texttt{6d27k3}$. \\
$S_6$ & $36$ & root line & One antipodal root line of the derived $\texttt{6d72k1}$; equivalently one parallel class of $1/4$-edges of $C$. \\
$3^3:S_4$ & $40$ & $\texttt{4d9k3}$-spread & A decomposition $C=Q_1\sqcup Q_2\sqcup Q_3$ with $Q_i\cong\texttt{4d9k3}=Y\otimes Y$. \\
$3^{1+2}_+:2A_4$ & $40$ & Affine $Y$-spread & A partition $C=Y_1\sqcup\cdots\sqcup Y_9$ whose $\texttt{4d9k3}$-sections form the affine plane $AG(2,3)$. \\
$2.(A_4\times A_4).2$ & $45$ & $Y=\texttt{2d3}$ & A $-\tfrac12$-triangle in $C$, classically a tritangent triple. \\
\bottomrule
\end{tabularx}
\caption{Euclidean maximal-subgroup dictionary for $U_4(2)$ from the Schl\"afli code $\texttt{6d27k3}$.}
\label{tab:u42-schlafli-euclidean}
\end{table}

\subsubsection*{The two index-$40$ objects}
Let $Y=\texttt{2d3}$ be an equilateral triangle.  We call a subcode $Q\cong Y\otimes Y=\texttt{4d9k3}$ a Steiner square.  Explicitly, if $Y=\{u_0,u_1,u_2\}\subset S^1$, then $Y\otimes Y=\{u_i\otimes u_j:0\le i,j\le2\}\subset S^3$.  It has two transverse decompositions into three $Y$'s, given by rows and columns; two distinct points in the same row or column have inner product $-1/2$, while two points in different rows and different columns have inner product $1/4$.

Inside $C=\texttt{6d27k3}$ there are $120$ Steiner squares, so a single square is not an index-$40$ object.  The index-$40$ object for $3^3:S_4$ is the induced spread $C=Q_1\sqcup Q_2\sqcup Q_3$, with each $Q_i\cong\texttt{4d9k3}$.  The other index-$40$ object is an Affine $Y$-spread: a partition $C=Y_1\sqcup\cdots\sqcup Y_9$ into nine triangles such that the triples of blocks whose union is a Steiner square form the $12$ lines of an affine plane of order $3$ on $\{Y_1,\ldots,Y_9\}$. 

Thus $\texttt{6d27k3}$, together with its derived edge-direction code $\texttt{6d72k1}$, gives a maximal-subgroup-complete Euclidean representation of $U_4(2)$.

\subsection{From the Fischer tower to the \texorpdfstring{\(\mathrm{Fi}_{22}\)}{Fi22} code}
\label{subsec:fi22-orbit-code}

The Fischer tower suggests that \(22d891k3\) should be read not only as a terminal code, but also as the local residue for the next Fischer group. In the \(U_6(2)=\mathrm{Fi}_{21}\) model, the Schl\"afli \(6d27k3\) subcodes, the Fischer lines, and the affine-plane type substructures organize the configurations \(12d81k3\), \(18d243k3\), \(20d405k3\), and \(21d567k3\). It is therefore natural to ask whether \(\mathrm{Fi}_{22}\) carries a \(1/4\)-spherical code whose \(22d891k3\) residues have the same local geometry.

The degree was predicted by a finite-geometric count. Let \(\Gamma\) be the rank-three Fischer graph for \(\mathrm{Fi}_{22}\), with parameters \((3510,693,180,126)\), and regard its vertices as hypothetical \(891\)-blocks. The analogy with \(6d27k3\subset22d891k3\) says that two blocks should meet in a \(6d27k3\) precisely when the corresponding vertices of \(\Gamma\) are adjacent, and should otherwise be disjoint. Fix one block \(B\). Its \(693\) neighbours then give \(693\) intersections of size \(27\). If these are exactly the \(693\) Schl\"afli subcodes inside \(B\), then each point of \(B\) lies in \((693\cdot27)/891=21\) such intersections. Hence a global point lies in \(1+21=22\) blocks. Double-counting point--block incidences gives \(22N=3510\cdot891\), and therefore \(N=142155\).

The ATLAS data then identify this predicted degree with a canonical coset action \cite{Atlas1985}. Put \(G=\mathrm{Fi}_{22}\) and \(H=2^{10}:M_{22}\); then \([G:H]=142155\). Let \(X=G/H\). The permutation module \(\mathbb R^X\) has a \(78\)-dimensional real constituent \(V\), and the projection of the point vector \(\mathbf 1_H\) to \(V\) spans an \(H\)-fixed line. If \(v\) is a unit vector on this line, set \(C_{\mathrm{Fi}_{22}}=\{gv:g\in G\}\subset S^{77}\subset\mathbb R^{78}\). Its stabilizer is \(H\), so \(|C_{\mathrm{Fi}_{22}}|=142155\).

In the same model, the \(3510\) \(891\)-blocks are the cosets of \(K=2\cdot U_6(2)\). The \(G\)-orbit of one incident pair in \(G/H\times G/K\) has bidegrees \(22\) and \(891\), matching the count above. A double-coset calculation gives the off-diagonal cosine set \(\{-\tfrac12,-\tfrac5{16},-\tfrac18,\tfrac1{16},\tfrac14\}\). Thus \(C_{\mathrm{Fi}_{22}}\) is a \(\texttt{78d142155k3}\) configuration.

This is not an ordinary optimality claim for \(K(78,1/4)\). We also finds antipodal holes with covering cosines \(1/4\) and \(\sqrt3/8\), and therefore \(K(78,1/4)>142155\). The exceptional feature is instead local: every point has exactly \(154=2\cdot77\) neighbours at inner product \(-1/2\). This is the Euclidean property isolated below as the fully triangulated configuration condition.

\subsection{Fully triangulated configurations}\label{sec:FTC}

We write \(S^d\subset\mathbb R^{d+1}\), and write \(I(C)=\{x\cdot y:x,y\in C,\ x\ne y\}\). Let \(P_i^{(d)}\) denote the normalized Gegenbauer polynomial on \(S^d\), with \(P_i^{(d)}(1)=1\).

The preceding \(\mathrm{Fi}_{22}\) code is not characterized by the ordinary spherical linear-programming bound alone. Its additional geometric feature is that every point has exactly \(2\cdot77=154\) neighbours at inner product \(-1/2\). This motivates the following condition.

\begin{definition}
A finite set $K\subset S^d$ satisfies the \emph{FTC condition} if
$x\cdot y\le 1/4$ for all distinct $x,y\in K$, and every $x\in K$ has at
least $2d$ points $y\in K$ with $x\cdot y=-1/2$.
\end{definition}

The terminology is justified by the first result below: the triangulation is
not an extra assumption, but a consequence of the two metric hypotheses.

\begin{lemma}\label{lem:nonpositive}
Let $E=\{e_1,\ldots,e_m\}\subset S^{d-1}$ and suppose $e_i\cdot e_j\le 0$ for
$i\ne j$.  Then $m\le 2d$.  If $m=2d$, then $E=\{\pm f_1,\ldots,\pm f_d\}$ for
an orthonormal basis $f_1,\ldots,f_d$ of $\mathbb R^d$.
\end{lemma}

\begin{proof}
Let $G=(e_i\cdot e_j)$.  Then $G\succeq 0$, $\operatorname{rank}G\le d$, and
$\operatorname{tr}G=m$.  Since $0\le \|\sum_i e_i\|^2=m+2\sum_{i<j}G_{ij}$, we
have $\sum_{i<j}(-G_{ij})\le m/2$.  Also $G_{ij}\in[-1,0]$ implies
$G_{ij}^2\le -G_{ij}$, hence $\operatorname{tr}G^2=m+2\sum_{i<j}G_{ij}^2\le 2m$.
Thus $m^2=(\operatorname{tr}G)^2\le \operatorname{rank}(G)\operatorname{tr}G^2\le 2dm$,
so $m\le 2d$.  If equality holds, all inequalities above are equalities; in
particular $G_{ij}^2=-G_{ij}$ for $i\ne j$.  Hence $G_{ij}\in\{0,-1\}$, so the
vectors split into antipodal pairs, and different pairs are orthogonal.
\end{proof}

\begin{proposition}[local saturation]\label{prop:local-saturation}
Let $K\subset S^d$ satisfy the FTC condition.  For every $u\in K$, there are
exactly $2d$ points $v\in K$ with $u\cdot v=-1/2$.  Moreover, for a suitable
orthonormal basis $e_1,\ldots,e_d$ of $u^\perp$, these points are precisely
$-u/2\pm (\sqrt3/2)e_i$, $1\le i\le d$.  In particular, if $u\cdot v=-1/2$,
then $-u-v\in K$.
\end{proposition}

\begin{proof}
Put $N(u)=\{v\in K:u\cdot v=-1/2\}$.  For $v\in N(u)$ set
$e_v=(2/\sqrt3)(v+u/2)\in u^\perp$.  Then $\|e_v\|=1$, and for $v\ne w$,
$e_v\cdot e_w=(4/3)(v\cdot w-1/4)\le 0$.  Lemma~\ref{lem:nonpositive} gives
$|N(u)|\le 2d$, while the FTC condition gives $|N(u)|\ge 2d$.  Hence equality
holds and $\{e_v:v\in N(u)\}$ is a cross-polytope in $u^\perp$.  Thus with
$e_1,\ldots,e_d$ chosen from one vector in each antipodal pair, $N(u)=\{-u/2\pm
(\sqrt3/2)e_i:1\le i\le d\}$.  The opposite point to
$-u/2+(\sqrt3/2)e_i$ is $-u/2-(\sqrt3/2)e_i=-u-v$.
\end{proof}

The same local form gives a useful forbidden interval for inner products.

\begin{proposition}[the gap after $-1/2$]\label{prop:gap}
Let $K\subset S^d$ satisfy the FTC condition.  If $u,x\in K$, $u\ne x$, and
$u\cdot x\ne -1/2$, then $u\cdot x\ge \rho_d$, where
$\rho_d=(-d+3\sqrt{d+4})/(2(d+3))$.  Consequently
$\operatorname{Cos}(K)\subset \{-1/2\}\cup[\rho_d,1/4]$.  In particular,
$\rho_{77}=-5/16$.
\end{proposition}

\begin{proof}
Use the notation of Proposition~\ref{prop:local-saturation} and write
$x=tu+\sum_i\beta_i e_i$, where $t=u\cdot x$.  Since $x$ is not a $-1/2$
neighbour of $u$, the inequalities with all points $-u/2\pm(\sqrt3/2)e_i$
give $-t/2\pm(\sqrt3/2)\beta_i\le 1/4$.  Hence
$|\beta_i|\le (1+2t)/(2\sqrt3)$, and in particular $t\ge -1/2$.  Since
$1-t^2=\sum_i\beta_i^2$, we get $1-t^2\le d(1+2t)^2/12$, or equivalently
$(d+3)t^2+dt+(d-12)/4\ge 0$.  On $[-1/2,1/4]$ this is exactly $t\ge \rho_d$.
\end{proof}

\subsubsection{The \texorpdfstring{$Fi_{22}$}{Fi22} certificate}

\begin{theorem}\label{thm:fi22-bound}
Let $C\subset S^{77}$ satisfy
$|I(C)|\subset\{-1/2,-5/16,-1/8,1/16,1/4\}$.  If every point of
$C$ has at least $154$ neighbours at inner product $-1/2$, then
$|C|\le 142155$.  Equality is attained by the $Fi_{22}$ code.
\end{theorem}

\begin{proof}
Let $P_i=P_i^{(77)}$ be the normalized Gegenbauer polynomial on $S^{77}$.  Set
\begin{equation}
F(t)=1+\frac{85572}{7175}P_1(t)+\frac{1071312}{1075}P_3(t)+\frac{1294336}{41}P_5(t)+\frac{559153152}{1505}P_7(t).
\end{equation}
All displayed nonconstant coefficients are positive and all remaining coefficients are zero, so positive-definiteness gives
$\sum_{x,y\in C}F(x\cdot y)\ge |C|^2$.  Direct evaluation gives
$F(-5/16)=F(-1/8)=F(1/16)=F(1/4)=0$, while $F(1)=404109$ and $F(-1/2)=-1701$.
Thus, using at least $154$ contacts at $-1/2$ in every row,
$\sum_{x,y\in C}F(x\cdot y)\le |C|(404109-154\cdot 1701)=142155|C|$.  Hence
$|C|^2\le 142155|C|$.
\end{proof}

\subsubsection{Low-degree momentum}

Let $\Gamma$ be the graph on $K$ in which $u\sim v$ means $u\cdot v=-1/2$, and
let $A$ be its adjacency matrix.  By Proposition~\ref{prop:local-saturation},
$\Gamma$ is $2d$-regular, each edge lies in a unique triangle, the local graph is $dK_2$, and the local form at $u$ is $-u/2\pm(\sqrt3/2)e_i$.

\begin{proposition}\label{prop:spectral}
Let $K\subset S^d$ satisfy the FTC condition.  Then $A$ acts on the restrictions to $K$
of spherical harmonics of degrees $0,1,2,3$ by the respective scalars $2d$, $-d$, $(d-3)/2$, and $(9-d)/4$.  Consequently $d\equiv 1
\pmod 4$.  If $d>1$, then $K$ is a spherical $3$-design.
\end{proposition}

\begin{proof}
The claim is local.  For a linear form $\ell$, the neighbour sum is
$\sum_{v\sim u}\ell(v)=-d\ell(u)$.  For $h_Q(x)=x^TQx$ with $\operatorname{tr}Q=0$,
one has $\sum_{v\sim u}h_Q(v)=((d-3)/2)h_Q(u)$.  For $h_T(x)=T(x,x,x)$ with $T$ trace-free, the pair
$-u/2\pm(\sqrt3/2)e_i$ contributes
\begin{equation}
-h_T(u)/4-(9/4)T(u,e_i,e_i).
\end{equation}
Summing over $i$ and using $\sum_iT(u,e_i,e_i)=-T(u,u,u)$ gives the claimed cubic scalar.
The degree-three restriction space is nonzero, for instance by a zonal
harmonic, so $(9-d)/4$ is an eigenvalue of the integral symmetric matrix $A$;
as it is rational, it is an integer.  Hence $d\equiv 1\pmod4$.  Finally,
summing $Ah=\theta h$ over $K$ gives $(2d-\theta)\sum_{x\in K}h(x)=0$.  For
$d>1$ and degrees $1,2,3$, the displayed eigenvalues are not $2d$, so all
moments up to degree $3$ vanish.
\end{proof}

For later reference we record the resulting one-point residual identities.
Fix $u\in K$, put $R(u)=K\setminus(\{u\}\cup N(u))$, set $N=|K|$, and write
$t_x=u\cdot x$ for $x\in R(u)$.

\begin{proposition}[residual moments]\label{prop:residual}
If $d>1$, then
$\sum_{x\in R(u)}1=N-1-2d$, $\sum_{x\in R(u)}t_x=d-1$,
$\sum_{x\in R(u)}t_x^2=N/(d+1)-1-d/2$, and
$\sum_{x\in R(u)}t_x^3=d/4-1$.  Moreover $t_x\in[\rho_d,1/4]$, with
$\rho_d=(-d+3\sqrt{d+4})/(2(d+3))$.
\end{proposition}

\begin{proof}
The moment identities follow from the $3$-design property by subtracting the
contribution of $u$ and its $2d$ neighbours, whose inner product with $u$ is
$-1/2$.  The interval is Proposition~\ref{prop:gap}.
\end{proof}

The first consequences require only the elementary nonnegative polynomials
$(t-\rho_d)(1/4-t)$ and $(1/4-t)(t-\rho_d)^2$ on $[\rho_d,1/4]$.

\begin{proposition}\label{prop:small-residual}
Among $d=5,9,13,17,21$, the FTC condition is possible only for $d=5$ and
$d=21$.  In those two cases the residual moment inequalities force
$|K|=27$ and $|K|=891$, respectively.
\end{proposition}

\begin{proof}
Summing $(t-\rho_d)(1/4-t)\ge0$ and $(1/4-t)(t-\rho_d)^2\ge0$ over $R(u)$ and
using Proposition~\ref{prop:residual} gives one upper and one lower bound for
$N=|K|$.  Direct simplification gives
\[
\begin{array}{c|ccccc}
 d&5&9&13&17&21\\ \hline
 N\le&27&\frac{1345+160\sqrt{13}}{27}&\frac{4389+672\sqrt{17}}{47}&\frac{91449+15552\sqrt{21}}{499}&891\\
 N\ge&27&\frac{1225+760\sqrt{13}}{43}&\frac{350595+93744\sqrt{17}}{3001}&\frac{124659+27864\sqrt{21}}{457}&891
\end{array}
\]
For $d=9,13,17$ the lower bound is larger than the upper bound.  For $d=5$ and
$d=21$ both bounds coincide, giving $N=27$ and $N=891$.
\end{proof}

When $d=21$, equality in the preceding proof also fixes the local distribution:
$\rho_{21}=-1/8$, so every $x\in R(u)$ has $u\cdot x\in\{-1/8,1/4\}$.  Since
$|R(u)|=848$ and $\sum_{x\in R(u)}u\cdot x=20$, the two multiplicities are
$512$ at $-1/8$ and $336$ at $1/4$.

\subsubsection{Higher momentum obstructions}

We next exclude the remaining congruence classes up to $45$.  The argument is
still a momentum argument: if $F=\sum_i c_iP_i^{(d)}$ has $c_0=0$ and
$c_i\ge0$ for $i\ge4$, then, since $K$ is a $3$-design,
$\sum_{x,y\in K}F(x\cdot y)\ge0$.

\begin{proposition}\label{prop:higher-obstructions}
There is no FTC in dimensions $d=25,29,33,37,41,45$.
\end{proposition}

\begin{proof}
For $d=25,29,33,37,41$, put
$a_d=(-12+\sqrt{3(45-d)})/(d+3)$ and
$F_d(t)=(t-1/4)t^2(t-a_d)^2$.  Since $(d+3)a_d^2+24a_d+3=0$, the constant
Gegenbauer coefficient of $F_d$ is zero.  Its coefficients at $P_4^{(d)}$ and
$P_5^{(d)}$ are
$-d(d+2)(8a_d+1)/(4(d+3)(d+5))$ and $d(d+2)/((d+5)(d+7))$, respectively; both
are positive because $a_d<-1/8$.  Hence $\sum_{x,y}F_d(x\cdot y)\ge0$.  On the
other hand, $F_d(t)\le0$ for $t\le1/4$, so
$\sum_{x,y}F_d(x\cdot y)\le N(F_d(1)+2dF_d(-1/2))$, and
\[
\begin{array}{c|ccccc}
 d&25&29&33&37&41\\ \hline
 F_d(1)+2dF_d(-1/2)&\frac{645-195\sqrt{15}}{784}&\frac{198-153\sqrt3}{256}&-\frac{17}{48}&\frac{33-24\sqrt6}{64}&\frac{639-783\sqrt3}{1936}
\end{array}
\]
is negative in every column, a contradiction.

For $d=45$, set
$F(t)=(t-1/4)t^2(t+7/32)^2(t+197/20)$.  Its constant coefficient is zero and
its coefficients at $P_4^{(45)},P_5^{(45)},P_6^{(45)}$ are
$4468243/2457600$, $339669/41600$, and $2303/3120$.  Thus the double sum is
nonnegative.  But $F(t)\le0$ on $[-1/2,1/4]$ and
$F(1)+90F(-1/2)=-64503/163840<0$, again a contradiction.
\end{proof}

Combining Propositions~\ref{prop:spectral}, \ref{prop:small-residual}, and
\ref{prop:higher-obstructions}, we obtain:

\begin{corollary}\label{cor:up-to-45}
If $K\subset S^d$ satisfies the FTC condition and $d\le45$, then
$d\in\{1,5,21\}$.
\end{corollary}

\subsubsection{The Hermitian examples}

The infinite family predicted by the congruence is supplied by Hermitian dual
polar geometry.  Let $V$ be a $2k$-dimensional Hermitian space over
$\mathbb F_4$, and let $\Gamma_k$ be the graph whose vertices are the maximal
totally isotropic $k$-subspaces, two vertices being adjacent when they meet in
a hyperplane.  The number of vertices is
$N_k=\prod_{i=1}^k(2^{2i-1}+1)$.

\begin{proposition}\label{prop:hermitian-examples}
For every $k\ge1$, the standard first-eigenspace embedding of $\Gamma_k$ gives
an FTC in $S^{d_k}$, where $d_k=(4^k-1)/3$.
\end{proposition}

\begin{proof}
The graph $\Gamma_k$ has valency $2d_k$.  Indeed, a fixed generator has
$(4^k-1)/3=d_k$ hyperplanes, and each such hyperplane lies in exactly two other
generators.  The standard embedding associated with the eigenvalue $-d_k$ has
rank $d_k+1$.  Its cosine sequence is $\sigma_i=(-1/2)^i$, where $i$ is the distance in $\Gamma_k$: this follows from the intersection numbers
$b_i=2^{2i+1}(4^{k-i}-1)/3$, $c_i=(4^i-1)/3$, and the recurrence
$c_i\sigma_{i-1}+a_i\sigma_i+b_i\sigma_{i+1}=-d_k\sigma_i$ with
$\sigma_0=1$, $\sigma_1=-1/2$.  Therefore distinct embedded vertices have
inner product at most $1/4$, and the vertices at inner product $-1/2$ are
exactly the $2d_k$ neighbours.  This is the FTC condition.
\end{proof}

The first cases are
$(d_k,N_k)=(1,3),(5,27),(21,891),(85,114939),\ldots$.

\subsection{A Beautiful Euclidean Characterization Problem}

We end with what seems to be the most natural question suggested by the preceding argument.

Recall the definition of the FTC condition: Let \(K\subset S^d\) be a set of unit vectors such that \(x\cdot y\le 1/4\) for all distinct \(x,y\in K\), and suppose that every \(x\in K\) has at least \(2d\) neighbours \(y\in K\) with \(x\cdot y=-1/2\). We have shown that this local condition is automatically saturated: every point has exactly \(2d\) such neighbours, the \(-1/2\)-graph is locally a cross-polytope, every \(-1/2\)-edge is completed to an equilateral triangle, and \(d\equiv 1\pmod 4\). Moreover, for \(d\le 45\), the only possibilities are \(d=1,5,21\).

Thus the following elementary Euclidean question remains.

\[
\textbf{FTC Problem.}\qquad
\text{For which }d\text{ does an FTC in }S^d\text{ exist, and what are all such configurations?}
\]

The known examples are the Hermitian dual polar configurations, with
\begin{equation}
d=\frac{4^m-1}{3},
\end{equation}
and the sporadic \(Fi_{22}\) configuration in \(S^{77}\), with \(142155\) points. It is tempting to ask whether these examples are all.

The case \(d=77\) is especially striking. The \(Fi_{22}\) code has cosine set
\begin{equation}
\left\{-\frac12,-\frac5{16},-\frac18,\frac1{16},\frac14\right\}.
\end{equation}
The ordinary spherical LP bound does not prove its optimality. However, after imposing the FTC edge condition, the bound becomes sharp.
This raises a natural sequence of questions.

\[
\begin{array}{ll}
\text{(1)} & \text{Is the equality case in this FTC-assisted bound unique?} \\
\text{(2)} & \text{Could the optimality been proved without FTC, and thus the code is in the constellation?} \\
\text{(3)} & \text{Is the \(Fi_{22}\) code the unique FTC in }S^{77}? \\
\text{(4)} & \text{Are the known Hermitian examples and the \(Fi_{22}\) example all FTCs?}
\end{array}
\]

An affirmative answer to the third question would give a remarkably short Euclidean characterization of the \(Fi_{22}\) geometry. Indeed, the condition would mention no finite simple group, no rank-three action, no 3-transposition geometry, no buildings, and no incidence axioms. It would say only this:

\begin{equation}
x\cdot y\le \frac14,\qquad
|\{y:x\cdot y=-\frac{1}{2}\}|\ge154
\quad\text{for every }x\in K\subset S^{77}.
\end{equation}

From these two elementary metric conditions, the group \(Fi_{22}\) would emerge as:
\begin{equation}
Fi_{22}:2=\operatorname{Aut}(K).
\end{equation}
It is plausible that a full classification of FTCs will require tools beyond ordinary one-point linear programming. The local cross-polytope structure points toward Terwilliger algebras, two-point and three-point semidefinite programming, distance-regular graph theory, regular near polygons, and the theory of finite polar spaces. In this sense the FTC problem sits at a useful interface: it is stated in elementary Euclidean language, but it appears to remember deep incidence geometry.

If the classification problem has only the known solutions, then the \(Fi_{22}\) would admit one of the simplest possible Euclidean characterizations of a sporadic group.

\newpage

\bibliography{references} 

@misc{cohn2024_spherical_codes,
  author       = {Henry Cohn},
  title        = {Table of spherical codes},
  year         = {2024},
  howpublished = {https://dspace.mit.edu/handle/1721.1/153543},
}

@article{1995Spherical,
  title={Spherical codes generated by binary partitions of symmetric pointsets},
  author={ Ericson, T.  and  Zinoviev, V. },
  journal={IEEE Transactions on Information Theory},
  volume={41},
  number={1},
  pages={107-129},
  year={1995},
}

@article{ganzhinov2024spherical,
  title={Spherical codes with prescribed signed permutation automorphisms inside shells of low-dimensional integer lattices},
  author={Ganzhinov, Mikhail and {\"O}sterg{\aa}rd, Patric RJ},
  journal={IEEE Transactions on Information Theory},
  volume={70},
  number={12},
  pages={8669--8674},
  year={2024},
  publisher={IEEE}
}

@article{Nebe2010,
  title={An even unimodular 72-dimensional lattice of minimum 8},
  author={Nebe, Gabriele},
  journal={Journal f{\"u}r die reine und angewandte Mathematik (Crelles Journal)},
  volume={2012},
  number={673},
  pages={237--247},
  year={2012},
  publisher={De Gruyter}
}

@article{leech1969six,
  title={Six and seven dimensional non-lattice sphere packings},
  author={Leech, John},
  journal={Canadian Mathematical Bulletin},
  volume={12},
  number={2},
  pages={151--155},
  year={1969},
  publisher={Cambridge University Press}
}

@article{korkine1873formes,
  title={Sur les formes quadratiques},
  author={Korkine, Aleksandr and Zolotareff, G},
  journal={Mathematische Annalen},
  volume={6},
  number={3},
  pages={366--389},
  year={1873},
  publisher={Springer-Verlag Berlin/Heidelberg}
}

@article{barnes1959some,
  title={Some extreme forms defined in terms of Abelian groups},
  author={Barnes, Eric Stephen and Wall, Gordon Elliott},
  journal={Journal of the Australian Mathematical Society},
  volume={1},
  number={1},
  pages={47--63},
  year={1959},
  publisher={Cambridge University Press}
}

@article{BoyvalenkovCherkashin2025,
  title={The kissing number in 48 dimensions for codes with certain forbidden distances is 52 416 000},
  author={Boyvalenkov, Peter and Cherkashin, Danila},
  journal={Results in Mathematics},
  volume={80},
  number={1},
  pages={3},
  year={2025},
  publisher={Springer}
}

@article{BoyvalenkovCherkashinDragnev2025,
  title={Universal optimality of $ T $-avoiding spherical codes and designs},
  author={Boyvalenkov, Peter G and Cherkashin, Danila D and Dragnev, Peter D},
  journal={arXiv preprint arXiv:2501.13906},
  year={2025}
}

@book{ericson2001codes,
  title={Codes on Euclidean spheres},
  author={Ericson, Thomas and Zinoviev, Victor},
  volume={63},
  year={2001},
  publisher={Elsevier}
}

@article{cohn2011rigidity,
  title={Rigidity of spherical codes},
  author={Cohn, Henry and Jiao, Yang and Kumar, Abhinav and Torquato, Salvatore},
  journal={Geometry \& Topology},
  volume={15},
  number={4},
  pages={2235--2273},
  year={2011},
  publisher={Mathematical Sciences Publishers}
}

@article{leech1971sphere,
  title={Sphere packings and error-correcting codes},
  author={Leech, John and Sloane, NJA},
  journal={Canadian Journal of Mathematics},
  volume={23},
  number={4},
  pages={718--745},
  year={1971},
  publisher={Cambridge University Press}
}

@article{cohn2024table,
  title={Table of kissing number bounds},
  author={Cohn, Henry},
  journal={.},
  year={2024}
}

@book{gray2000hilbert,
  title={The Hilbert Challenge},
  author={Gray, Jeremy},
  year={2000},
  publisher={OUP Oxford}
}

@inproceedings{liberti2017mathematical,
  title={Mathematical programming bounds for kissing numbers},
  author={Liberti, Leo},
  booktitle={International Conference on Optimization and Decision Science},
  pages={213--222},
  year={2017},
  organization={Springer}
}

@article{liu2023kissing,
  title={Kissing number of codes: A survey},
  author={Liu, Yi and Cheng, Wei and Rioul, Olivier and Guilley, Sylvain and Sol{\'e}, Patrick},
  journal={Coding Theory and applications (tentative title)},
  year={2023},
  publisher={Springer}
}

@article{boyvalenkov2012survey,
  title={A SURVEY ON THE KISSING NUMBERS},
  author={Boyvalenkov, Peter and Dodunekov, Stefan and Musin, Oleg},
  journal={Serdica Math. J},
  volume={38},
  pages={507--522},
  year={2012}
}

@article{ganzhinov2025highly,
  title={Highly symmetric lines},
  author={Ganzhinov, Mikhail},
  journal={Linear Algebra and its Applications},
  volume={722},
  pages={12--37},
  year={2025},
  publisher={Elsevier}
}

@book{brass2005research,
  title={Research problems in discrete geometry},
  author={Brass, Peter and Moser, William OJ and Pach, J{\'a}nos},
  year={2005},
  publisher={Springer}
}

@article{cohn2024improved,
  title={Improved kissing numbers in seventeen through twenty-one dimensions},
  author={Cohn, Henry and Li, Anqi},
  journal={arXiv preprint arXiv:2411.04916},
  year={2024}
}

@article{cohn2007universally,
  title={Universally optimal distribution of points on spheres},
  author={Cohn, Henry and Kumar, Abhinav},
  journal={Journal of the American Mathematical Society},
  volume={20},
  number={1},
  pages={99--148},
  year={2007}
}

@article{cohn2024optimality,
  title={Optimality of spherical codes via exact semidefinite programming bounds},
  author={Cohn, Henry and de Laat, David and Leijenhorst, Nando},
  journal={arXiv preprint arXiv:2403.16874},
  year={2024}
}

@article{kallal2017improved,
  title={Improved lower bounds for kissing numbers in dimensions 25 through 31},
  author={Kallal, Kenz and Kan, Tomoka and Wang, Eric},
  journal={SIAM Journal on Discrete Mathematics},
  volume={31},
  number={3},
  pages={1895--1908},
  year={2017},
  publisher={SIAM}
}

@article{pfender2004kissing,
  title={Kissing numbers, sphere packings, and some unexpected proofs},
  author={Pfender, Florian and Ziegler, Gunter M},
  journal={Notices of the AMS},
  volume={51},
  number={8},
  pages={873--883},
  year={2004}
}

@book{conway2013sphere,
  title={Sphere packings, lattices and groups},
  author={Conway, John Horton and Sloane, Neil James Alexander},
  volume={290},
  year={2013},
  publisher={Springer Science \& Business Media}
}

@article{leech1964some,
  title={Some sphere packings in higher space},
  author={Leech, John},
  journal={Canadian Journal of Mathematics},
  volume={16},
  pages={657--682},
  year={1964},
  publisher={Cambridge University Press}
}

@article{leech1967notes,
  title={Notes on sphere packings},
  author={Leech, John},
  journal={Canadian Journal of Mathematics},
  volume={19},
  pages={251--267},
  year={1967},
  publisher={Cambridge University Press}
}

@incollection{van1991equilateral,
  title={Equilateral point sets in elliptic geometry},
  author={van Lint, Jacobus H and Seidel, Johan J},
  booktitle={Geometry and Combinatorics},
  pages={3--16},
  year={1991},
  publisher={Elsevier}
}

@misc{BrouwerSchlaefli,
   title = {The {Schl\"afli} graph},
   author = {A. E. Brouwer},
   howpublished = {\url{https://aeb.win.tue.nl/graphs/Schlaefli.html}},
 }

@article{cuypers2012lie,
  title={Lie algebras and 3-transpositions},
  author={Cuypers, H and Horn, M and Shpectorov, S and others},
  journal={Journal of Algebra},
  volume={368},
  pages={21--39},
  year={2012},
  publisher={Elsevier}
}

@article{Leech1957,
  author  = {Leech, John},
  title   = {Equilibrium of sets of particles on a sphere},
  journal = {The Mathematical Gazette},
  volume  = {41},
  number  = {336},
  pages   = {81--90},
  year    = {1957},
  doi     = {10.2307/3610579}
}

@article{CohnElkiesKumarSchuermann2010,
  author  = {Cohn, Henry and Elkies, Noam D. and Kumar, Abhinav and Sch{\"u}rmann, Achill},
  title   = {Point configurations that are asymmetric yet balanced},
  journal = {Proceedings of the American Mathematical Society},
  volume  = {138},
  number  = {8},
  pages   = {2863--2872},
  year    = {2010},
  doi     = {10.1090/S0002-9939-10-10284-6}
}

@article{musin2008kissing,
  title={The kissing number in four dimensions},
  author={Musin, Oleg R},
  journal={Annals of Mathematics},
  pages={1--32},
  year={2008},
  publisher={JSTOR}
}

@article{schutte1952problem,
  title={Das problem der dreizehn Kugeln},
  author={Sch{\"u}tte, Kurt and van der Waerden, Bartel Leendert},
  journal={Mathematische Annalen},
  volume={125},
  number={1},
  pages={325--334},
  year={1952},
  publisher={Springer}
}

@article{viazovska2017sphere,
  title={The sphere packing problem in dimension 8},
  author={Viazovska, Maryna S},
  journal={Annals of mathematics},
  pages={991--1015},
  year={2017},
  publisher={JSTOR}
}

@article{cohn2017sphere,
  title={The sphere packing problem in dimension 24},
  author={Cohn, Henry and Kumar, Abhinav and Miller, Stephen and Radchenko, Danylo and Viazovska, Maryna},
  journal={Annals of mathematics},
  volume={185},
  number={3},
  pages={1017--1033},
  year={2017},
  publisher={Department of Mathematics, Princeton University Princeton, New Jersey, USA}
}

@article{novikov2025alphaevolve,
  title={AlphaEvolve: A coding agent for scientific and algorithmic discovery},
  author={Novikov, Alexander and V{\~u}, Ng{\^a}n and Eisenberger, Marvin and Dupont, Emilien and Huang, Po-Sen and Wagner, Adam Zsolt and Shirobokov, Sergey and Kozlovskii, Borislav and Ruiz, Francisco JR and Mehrabian, Abbas and others},
  journal={arXiv preprint arXiv:2506.13131},
  year={2025}
}

@phdthesis{delsarte1973,
	author       = {Delsarte, Philippe},
	title        = {An Algebraic Approach to the Association Schemes of
	Coding Theory},
	school       = {Universit\'{e} Catholique de Louvain},
	year         = {1973},
	note         = {Philips Research Reports Supplements, No.\ 10}
}

@article{delsarte1977,
	author       = {Delsarte, Philippe and Goethals, Jean-Marie and
	Seidel, Johan Jacob},
	title        = {Spherical Codes and Designs},
	journal      = {Geometriae Dedicata},
	volume       = {6},
	number       = {3},
	pages        = {363--388},
	year         = {1977},
	doi          = {10.1007/BF03187604}
}

@article{schoenberg1942,
	author       = {Schoenberg, Isaac J.},
	title        = {Positive Definite Functions on Spheres},
	journal      = {Duke Mathematical Journal},
	volume       = {9},
	number       = {1},
	pages        = {96--108},
	year         = {1942},
	doi          = {10.1215/S0012-7094-42-00908-6}
}

@article{kabatyanskii1978,
	author       = {Kabaty\'{a}nski\u{\i}, Grigorii A. and
	Levenshtein, Vladimir I.},
	title        = {Bounds for Packings on a Sphere and in Space},
	journal      = {Problems of Information Transmission},
	volume       = {14},
	number       = {1},
	pages        = {1--17},
	year         = {1978}
}

@article{levenshtein1979,
	author       = {Levenshtein, Vladimir I.},
	title        = {On Bounds for Packings in $n$-Dimensional {E}uclidean
	Space},
	journal      = {Soviet Mathematics Doklady},
	volume       = {20},
	number       = {2},
	pages        = {417--421},
	year         = {1979}
}

@article{levenshtein1992,
	author       = {Levenshtein, Vladimir I.},
	title        = {Designs as Maximum Codes in Polynomial Metric Spaces},
	journal      = {Acta Applicandae Mathematicae},
	volume       = {29},
	number       = {1--2},
	pages        = {1--82},
	year         = {1992},
	doi          = {10.1007/BF00053379}
}

@article{bachoc2008,
	author       = {Bachoc, Christine and Vallentin, Frank},
	title        = {New Upper Bounds for Kissing Numbers from Semidefinite
	Programming},
	journal      = {Journal of the American Mathematical Society},
	volume       = {21},
	number       = {3},
	pages        = {909--924},
	year         = {2008},
	doi          = {10.1090/S0894-0347-07-00589-9}
}

@article{delaat2015,
	author       = {de Laat, David and Vallentin, Frank},
	title        = {A Semidefinite Programming Hierarchy for Packing Problems
	in Discrete Geometry},
	journal      = {Mathematical Programming, Series B},
	volume       = {151},
	number       = {2},
	pages        = {529--553},
	year         = {2015},
	doi          = {10.1007/s10107-015-0894-1}
}

@article{delaat2021,
	author       = {de Laat, David and Machado, Fabr\'{i}cio C. and
	Oliveira, Fernando M. de and Vallentin, Frank},
	title        = {{$k$}-Point Semidefinite Programming Bounds for
	Equiangular Lines},
	journal      = {Mathematical Programming},
	volume       = {194},
	number       = {1--2},
	pages        = {533--567},
	year         = {2022},
	doi          = {10.1007/s10107-021-01638-x},
	note         = {arXiv:1812.06045}
}

@article{cohn2024,
	author       = {Cohn, Henry and de Laat, David and Leijenhorst, Nando},
	title        = {Optimality of Spherical Codes via Exact Semidefinite
	Programming Bounds},
	journal      = {arXiv preprint arXiv:2403.16874},
	year         = {2024}
}

@article{polu2020generative,
  title={{Generative Language Modeling for Automated Theorem Proving}},
  author={Polu, Stanislas and Sutskever, Ilya},
  journal={arXiv preprint arXiv:2009.03393},
  year={2020},
  doi={10.48550/arXiv.2009.03393},
  url={https://arxiv.org/abs/2009.03393}
}

@inproceedings{polu2022formal,
  title={{Formal Mathematics Statement Curriculum Learning}},
  author={Polu, Stanislas and Han, Jesse Michael and Zheng, Kunhao and Baksys, Mantas and Babuschkin, Igor and Sutskever, Ilya},
  booktitle={International Conference on Learning Representations},
  year={2023},
  url={https://openreview.net/forum?id=-P7G-8dmSh4}
}

@inproceedings{zheng2021minif2f,
  title={{MiniF2F}: A Cross-System Benchmark for Formal Olympiad-Level Mathematics},
  author={Zheng, Kunhao and Han, Jesse Michael and Polu, Stanislas},
  booktitle={International Conference on Learning Representations},
  year={2022},
  url={https://openreview.net/forum?id=9ZPegFuFTFv}
}

@article{xin2024deepseekprover,
  title={{DeepSeek-Prover}: Advancing Theorem Proving in {LLMs} through Large-Scale Synthetic Data},
  author={Xin, Huajian and Guo, Daya and Shao, Zhihong and Ren, Zhizhou and Zhu, Qihao and Liu, Bo and Ruan, Chong and Li, Wenda and Liang, Xiaodan},
  journal={arXiv preprint arXiv:2405.14333},
  year={2024},
  doi={10.48550/arXiv.2405.14333},
  url={https://arxiv.org/abs/2405.14333}
}

@article{trinh2024alphageometry,
  title={Solving Olympiad Geometry without Human Demonstrations},
  author={Trinh, Trieu H. and Wu, Yuhuai and Le, Quoc V. and He, He and Luong, Thang},
  journal={Nature},
  volume={625},
  number={7995},
  pages={476--482},
  year={2024},
  doi={10.1038/s41586-023-06747-5},
  url={https://www.nature.com/articles/s41586-023-06747-5}
}

@article{hubert2026alphaproof,
  title={Olympiad-Level Formal Mathematical Reasoning with Reinforcement Learning},
  author={Hubert, Thomas and Mehta, Rishi and Sartran, Laurent and others},
  journal={Nature},
  volume={651},
  pages={607--613},
  year={2026},
  doi={10.1038/s41586-025-09833-y},
  url={https://www.nature.com/articles/s41586-025-09833-y}
}

@article{chervonyi2025alphageometry2,
  title={Gold-Medalist Performance in Solving Olympiad Geometry with {AlphaGeometry2}},
  author={Chervonyi, Yuri and Trinh, Trieu H. and Ol{\v{s}}{\'a}k, Miroslav and Yang, Xiaomeng and Nguyen, Hoang and Menegali, Marcelo and Jung, Junehyuk and Kim, Junsu and Verma, Vikas and Le, Quoc V. and Luong, Thang},
  journal={arXiv preprint arXiv:2502.03544},
  year={2025},
  doi={10.48550/arXiv.2502.03544},
  url={https://arxiv.org/abs/2502.03544}
}

@article{davies2021advancing,
  title={Advancing Mathematics by Guiding Human Intuition with {AI}},
  author={Davies, Alex and Veli{\v{c}}kovi{\'c}, Petar and Buesing, Lars and Blackwell, Sam and Zheng, Daniel and Toma{\v{s}}ev, Nenad and Tanburn, Richard and Battaglia, Peter and Blundell, Charles and Juh{\'a}sz, Andr{\'a}s and Lackenby, Marc and Williamson, Geordie and Hassabis, Demis and Kohli, Pushmeet},
  journal={Nature},
  volume={600},
  number={7887},
  pages={70--74},
  year={2021},
  doi={10.1038/s41586-021-04086-x},
  url={https://www.nature.com/articles/s41586-021-04086-x}
}

@article{fawzi2022alphatensor,
  title={Discovering Faster Matrix Multiplication Algorithms with Reinforcement Learning},
  author={Fawzi, Alhussein and Balog, Matej and Huang, Aja and Hubert, Thomas and Romera-Paredes, Bernardino and Barekatain, Mohammadamin and Novikov, Alexander and Ruiz, Francisco J. R. and Schrittwieser, Julian and Swirszcz, Grzegorz and Silver, David and Hassabis, Demis and Kohli, Pushmeet},
  journal={Nature},
  volume={610},
  number={7930},
  pages={47--53},
  year={2022},
  doi={10.1038/s41586-022-05172-4},
  url={https://www.nature.com/articles/s41586-022-05172-4}
}

@article{romera2024funsearch,
  title={Mathematical Discoveries from Program Search with Large Language Models},
  author={Romera-Paredes, Bernardino and Barekatain, Mohammadamin and Novikov, Alexander and Balog, Matej and Kumar, M. Pawan and Dupont, Emilien and Ruiz, Francisco J. R. and Ellenberg, Jordan S. and Wang, Pengming and Fawzi, Omar and Kohli, Pushmeet and Fawzi, Alhussein},
  journal={Nature},
  volume={625},
  number={7995},
  pages={468--475},
  year={2024},
  doi={10.1038/s41586-023-06924-6},
  url={https://www.nature.com/articles/s41586-023-06924-6}
}

@article{Aschbacher1984,
  author  = {Aschbacher, Michael},
  title   = {On the maximal subgroups of the finite classical groups},
  journal = {Inventiones Mathematicae},
  volume  = {76},
  pages   = {469--514},
  year    = {1984},
  doi     = {10.1007/BF01388470}
}

@book{Atlas1985,
  author    = {Conway, J. H. and Curtis, R. T. and Norton, S. P. and Parker, R. A. and Wilson, R. A.},
  title     = {{ATLAS} of Finite Groups: Maximal Subgroups and Ordinary Characters for Simple Groups},
  publisher = {Clarendon Press},
  address   = {Oxford},
  year      = {1985}
}

@article{BannaiSloane,
  author  = {Bannai, Eiichi and Sloane, N. J. A.},
  title   = {Uniqueness of certain spherical codes},
  journal = {Canadian Journal of Mathematics},
  volume  = {33},
  number  = {2},
  pages   = {437--449},
  year    = {1981},
  doi     = {10.4153/CJM-1981-038-7}
}

@article{Buekenhout1979,
  author  = {Buekenhout, Francis},
  title   = {Diagrams for geometries and groups},
  journal = {Journal of Combinatorial Theory, Series A},
  volume  = {27},
  number  = {2},
  pages   = {121--151},
  year    = {1979},
  doi     = {10.1016/0097-3165(79)90041-4}
}

@article{CerchiaiVanGeemen,
  author        = {Cerchiai, Bianca Letizia and van Geemen, Bert},
  title         = {From qubits to {$E_7$}},
  journal       = {Journal of Mathematical Physics},
  volume        = {51},
  number        = {12},
  pages         = {122203},
  year          = {2010},
  doi           = {10.1063/1.3519379},
  eprint        = {1003.4255},
  archivePrefix = {arXiv},
  primaryClass  = {math.AG}
}

@article{Fischer1971,
  author  = {Fischer, Bernd},
  title   = {Finite groups generated by {3}-transpositions. {I}.},
  journal = {Inventiones mathematicae},
  volume  = {13},
  pages   = {232--246},
  year    = {1971},
  doi     = {10.1007/BF01404633}
}

@article{CohnDeLaatLeijenhorst2024,
  author        = {Cohn, Henry and de Laat, David and Leijenhorst, Nando},
  title         = {Optimality of spherical codes via exact semidefinite programming bounds},
  journal       = {arXiv preprint},
  year          = {2024},
  eprint        = {2403.16874},
  archivePrefix = {arXiv},
  primaryClass  = {math.MG}
}

@article{CohnJiaoKumarTorquato2011,
  author        = {Cohn, Henry and Jiao, Yang and Kumar, Abhinav and Torquato, Salvatore},
  title         = {Rigidity of spherical codes},
  journal       = {Geometry \& Topology},
  volume        = {15},
  number        = {4},
  pages         = {2235--2273},
  year          = {2011},
  doi           = {10.2140/gt.2011.15.2235},
  eprint        = {1102.5060},
  archivePrefix = {arXiv},
  primaryClass  = {math.MG}
}

@misc{CohnKissingNumbers,
  author       = {Cohn, Henry},
  title        = {Kissing numbers},
  howpublished = {\url{https://cohn.mit.edu/kissing-numbers/}},
  note         = {Accessed 2026-05-24},
  year         = {2026}
}

@article{CohnKumar2007,
  author        = {Cohn, Henry and Kumar, Abhinav},
  title         = {Universally optimal distribution of points on spheres},
  journal       = {Journal of the American Mathematical Society},
  volume        = {20},
  number        = {1},
  pages         = {99--148},
  year          = {2007},
  doi           = {10.1090/S0894-0347-06-00546-7},
  eprint        = {math/0607446},
  archivePrefix = {arXiv},
  primaryClass  = {math.MG}
}

@book{ConwaySloaneSPLAG,
  author    = {Conway, J. H. and Sloane, N. J. A.},
  title     = {Sphere Packings, Lattices and Groups},
  edition   = {3},
  series    = {Grundlehren der mathematischen Wissenschaften},
  volume    = {290},
  publisher = {Springer},
  address   = {New York},
  year      = {1999},
  doi       = {10.1007/978-1-4757-6568-7}
}

@book{CoxeterRegularPolytopes,
  author    = {Coxeter, H. S. M.},
  title     = {Regular Polytopes},
  edition   = {3},
  publisher = {Dover Publications},
  address   = {New York},
  year      = {1973}
}

@article{CurtisMOG,
  author  = {Curtis, R. T.},
  title   = {A new combinatorial approach to {$M_{24}$}},
  journal = {Mathematical Proceedings of the Cambridge Philosophical Society},
  volume  = {79},
  number  = {1},
  pages   = {25--42},
  year    = {1976},
  doi     = {10.1017/S0305004100052075}
}

@article{CurtisSubgroupsI,
  author  = {Curtis, R. T.},
  title   = {On subgroups of {${\cdot}O$}. {I}. Lattice stabilizers},
  journal = {Journal of Algebra},
  volume  = {27},
  pages   = {549--573},
  year    = {1973}
}

@article{CurtisSubgroupsII,
  author  = {Curtis, R. T.},
  title   = {On subgroups of {${\cdot}O$}. {II}. Local structure},
  journal = {Journal of Algebra},
  volume  = {63},
  pages   = {413--434},
  year    = {1980}
}

@article{DelsarteGoethalsSeidel,
  author  = {Delsarte, P. and Goethals, J. M. and Seidel, J. J.},
  title   = {Spherical codes and designs},
  journal = {Geometriae Dedicata},
  volume  = {6},
  number  = {3},
  pages   = {363--388},
  year    = {1977},
  doi     = {10.1007/BF03187604}
}

@book{DixonMortimer1996,
  author    = {Dixon, John D. and Mortimer, Brian},
  title     = {Permutation Groups},
  series    = {Graduate Texts in Mathematics},
  volume    = {163},
  publisher = {Springer},
  address   = {New York},
  year      = {1996},
  doi       = {10.1007/978-1-4612-0731-3}
}

@article{Giudici2012,
  author  = {Giudici, Michael and Praeger, Cheryl E. and Pearce, Geoffrey},
  title   = {Basic coset geometries},
  journal = {Journal of Algebraic Combinatorics},
  volume  = {36},
  number  = {4},
  pages   = {561--594},
  year    = {2012},
  doi     = {10.1007/s10801-012-0350-8}
}

@book{HumphreysCoxeter,
  author    = {Humphreys, James E.},
  title     = {Reflection Groups and Coxeter Groups},
  series    = {Cambridge Studies in Advanced Mathematics},
  volume    = {29},
  publisher = {Cambridge University Press},
  address   = {Cambridge},
  year      = {1990}
}

@article{KallalKanWang2017,
  author        = {Kallal, Kenz and Kan, Tomoka and Wang, Eric},
  title         = {Improved lower bounds for kissing numbers in dimensions 25 through 31},
  journal       = {SIAM Journal on Discrete Mathematics},
  volume        = {31},
  number        = {3},
  pages         = {1895--1908},
  year          = {2017},
  doi           = {10.1137/16M1095810},
  eprint        = {1608.07270},
  archivePrefix = {arXiv},
  primaryClass  = {math.MG}
}

@article{LamShimakura2008,
  author        = {Lam, Ching Hung and Shimakura, Hiroki},
  title         = {Quadratic spaces and holomorphic framed vertex operator algebras of central charge 24},
  journal       = {Proceedings of the London Mathematical Society},
  volume        = {104},
  number        = {3},
  pages         = {540--576},
  year          = {2012},
  doi           = {10.1112/plms/pdr041},
  eprint        = {1010.5303},
  archivePrefix = {arXiv},
  primaryClass  = {math.QA}
}

@article{LuyckxThas,
  author  = {Luyckx, D. and Thas, J. A.},
  title   = {Trialities and 1-systems of {$Q^+(7,q)$}},
  journal = {Designs, Codes and Cryptography},
  volume  = {35},
  number  = {3},
  pages   = {337--352},
  year    = {2005},
  doi     = {10.1007/s10623-003-6742-y}
}

@article{Manivel,
  author        = {Manivel, Laurent},
  title         = {Configurations of lines and models of Lie algebras},
  journal       = {Journal of Algebra},
  volume        = {304},
  number        = {1},
  pages         = {457--486},
  year          = {2006},
  doi           = {10.1016/j.jalgebra.2006.04.029},
  eprint        = {math/0507118},
  archivePrefix = {arXiv},
  primaryClass  = {math.AG}
}

@article{MusinKissing3,
  author        = {Musin, Oleg R.},
  title         = {The kissing problem in three dimensions},
  journal       = {Discrete \& Computational Geometry},
  volume        = {35},
  number        = {3},
  pages         = {375--384},
  year          = {2006},
  doi           = {10.1007/s00454-005-1201-3},
  eprint        = {math/0410324},
  archivePrefix = {arXiv},
  primaryClass  = {math.MG}
}

@article{MusinKissing4,
  author        = {Musin, Oleg R.},
  title         = {The kissing number in four dimensions},
  journal       = {Annals of Mathematics},
  volume        = {168},
  number        = {1},
  pages         = {1--32},
  year          = {2008},
  doi           = {10.4007/annals.2008.168.1},
  eprint        = {math/0309430},
  archivePrefix = {arXiv},
  primaryClass  = {math.MG}
}

@article{MusinTwoDistance,
  author        = {Musin, Oleg R.},
  title         = {Graphs and spherical two-distance sets},
  journal       = {European Journal of Combinatorics},
  volume        = {80},
  pages         = {311--325},
  year          = {2019},
  doi           = {10.1016/j.ejc.2018.07.013},
  eprint        = {1608.03392},
  archivePrefix = {arXiv},
  primaryClass  = {math.MG}
}

@article{NordstromRobinson,
  author  = {Nordstrom, A. W. and Robinson, J. P.},
  title   = {An optimum nonlinear code},
  journal = {Information and Control},
  volume  = {11},
  number  = {5--6},
  pages   = {613--616},
  year    = {1967},
  doi     = {10.1016/S0019-9958(67)90835-2}
}

@book{PayneThas,
  author    = {Payne, Stanley E. and Thas, Joseph A.},
  title     = {Finite Generalized Quadrangles},
  edition   = {2},
  series    = {EMS Series of Lectures in Mathematics},
  publisher = {European Mathematical Society},
  address   = {Z{\"u}rich},
  year      = {2009},
  doi       = {10.4171/066}
}

@article{SchutteWaerden1953,
  author  = {Sch{\"u}tte, K. and van der Waerden, B. L.},
  title   = {Das Problem der dreizehn Kugeln},
  journal = {Mathematische Annalen},
  volume  = {125},
  pages   = {325--334},
  year    = {1953},
  doi     = {10.1007/BF01343127}
}

@book{TitsBuildings,
  author    = {Tits, Jacques},
  title     = {Buildings of Spherical Type and Finite {$BN$}-Pairs},
  series    = {Lecture Notes in Mathematics},
  volume    = {386},
  publisher = {Springer},
  address   = {Berlin},
  year      = {1974},
  doi       = {10.1007/BFb0057437}
}

@article{Wilson1983Co1,
  author  = {Wilson, R. A.},
  title   = {The maximal subgroups of Conway's group {$Co_1$}},
  journal = {Journal of Algebra},
  volume  = {85},
  number  = {1},
  pages   = {144--165},
  year    = {1983},
  doi     = {10.1016/0021-8693(83)90122-9}
}

@misc{Wilson2017Sporadic,
  author        = {Wilson, R. A.},
  title         = {Maximal subgroups of sporadic groups},
  year          = {2017},
  eprint        = {1701.02095},
  archivePrefix = {arXiv},
  primaryClass  = {math.GR},
  note          = {arXiv preprint}
}

@misc{WilsonFSGNotes,
  author = {Wilson, R. A.},
  title  = {Lecture notes on finite simple groups: the Leech lattice and the Conway group},
  note   = {Lecture notes},
  year   = {2008}
}
\bibliographystyle{unsrt}  


\end{document}